\documentclass[11pt,twoside,a4paper]{report}
\usepackage{styles/ttstyle}
\usepackage{enumerate}
\usepackage{fancyhdr}

\newcommand{\clearemptydoublepage}{
  \newpage{\pagestyle{empty}
  \cleardoublepage}}

\usepackage[round,colon,authoryear]{natbib}
\renewcommand{\cite}{\citep}

\title{
Learning to Predict Combinatorial Structures
}

\author{
Shankar Vembu
}

\date{}

\begin{document}

\maketitle
\clearemptydoublepage
\pagenumbering{roman}
\clearemptydoublepage
%\input{src/abstract_de}
%\clearemptydoublepage
\chapter*{Abstract}
\addcontentsline{toc}{chapter}{Abstract}
The major challenge in designing a discriminative learning algorithm for predicting structured data is to address the computational issues arising from the exponential size of the output space. Existing algorithms make different assumptions to ensure efficient, polynomial time estimation of model parameters. For several combinatorial structures, including cycles, partially ordered sets, permutations and other graph classes, these assumptions do not hold. In this thesis, we address the problem of designing learning algorithms for predicting combinatorial structures by introducing two new assumptions:
\begin{enumerate}[(i)]
\item The first assumption is that a particular counting problem can be solved efficiently. The consequence is a generalisation of the classical ridge regression for structured prediction.
\item The second assumption is that a particular sampling problem can be solved efficiently. The consequence is a new technique for designing and analysing probabilistic structured prediction models.
\end{enumerate}
These results can be applied to solve several complex learning problems including but not limited to multi-label classification, multi-category hierarchical classification, and label ranking.

\clearemptydoublepage
\chapter*{Acknowledgements}
\addcontentsline{toc}{chapter}{Acknowledgments}

To Stefan Wrobel, for giving me the opportunity to pursue doctoral studies at Fraunhofer IAIS, and for reading drafts of this manuscript and suggesting improvements.\\\\
To Michael May, for giving me the opportunity to pursue doctoral studies at the Knowledge Discovery lab.\\\\
To Michael Clausen, Peter K\"opke, and Andreas Weber, for serving on my thesis committee.\\\\
To Thomas G\"artner, for continual advice, for introducing me to graph theory and kernel methods, for teaching me how to write technical papers, for numerous discussions on topics related to machine learning, and for carefully reading drafts of this manuscript and suggesting improvements.\\\\
To Tamas Horv{\'a}th and Kristian Kersting, for numerous discussions on research in general.\\\\
To Mario Boley, for discussions on topics related to theoretical computer science, and for being a wonderful colleague.\\\\
To J\"org Kindermann, for providing me with computing resources needed to run the experiments described in Chapter 5.\\\\
To Daniela B\"orner, Renate Henkeler, and Myriam Jourdan, for helping me with administrative issues.\\\\
To Jens Humrich and Katrin Ullrich, for being wonderful colleagues.\\\\
To all the members of the IAIS.KD lab, for creating a conducive atmosphere that allowed me to work towards a doctorate in machine learning.\\\\
To family and friends.\\\\
And lastly, to David Baldacci's novels, for inspiring me to acknowledge in this way!

\clearemptydoublepage

\pagestyle{fancyplain}
\tableofcontents
\clearemptydoublepage

%\listoffigures
%\clearemptydoublepage

%\listoftables
%\clearemptydoublepage

% hack
\lhead[\fancyplain{}{\small\sl\thepage}]%
      {\fancyplain{}{\small\sl}}
\rhead[\fancyplain{}{\small\sl NOTATIONAL CONVENTIONS}]%
      {\fancyplain{}{\small\sl\thepage}}

\chapter*{Notational Conventions}
\addcontentsline{toc}{chapter}{Notational Conventions}
We will follow the notational conventions given below wherever possible.
\begin{itemize}
\item Calligraphic letters ($\Cal{A, B} \ldots$) denote sets (or particular spaces):
\begin{itemize}
\item $\Xcal$ an instance space.
\item $\Ycal$ a label set.
%\item $\Vcal$ the vertex set of a graph,
%\item $\Ecal$ the edge set of a graph, and
\item $\Hcal$ a Hilbert space.
\end{itemize}
\item Capital letters ($A, B, \ldots $) denote matrices or subsets of some set.
%\begin{itemize}
%\item $E$ the adjacency matrix of a graph.
%\end{itemize}
\item Bold letters or numbers denote special matrices or vectors:
\begin{itemize}
\item $\iden$ the identity matrix, i.e., a diagonal matrix (of appropriate dimension)
with all components on the diagonal equal to $1$.
\item $\zero$ the zero element of a vector space or a matrix with all components equal to $0$. For the vector spaces $\R^n$ the zero element is the vector (of appropriate dimension) with all components equal to $0$.
\hide{In function spaces we will sometimes use $\zero(\cdot)$ to make explicit that $\zero$ is a function}
\item $\one$ the matrix (in $\R^{n\times m}$) or the vector (in $\R^n$) with all elements equal to $1$.
\end{itemize}
\item Lowercase letters ($a, b, \ldots$) denote vectors, numbers, elements of some set, or functions:
\begin{itemize}
\item $m$ the number of training instances.
\item $x$ a single instance.
\item $y$ a single label.
\end{itemize}
\item Lowercase Greek letters $\alpha, \beta, \ldots$ denote real numbers.
\item Bold lowercase Greek letters $\balpha, \bbeta, \ldots$ denote vectors of real numbers.
\item Symbols:
\begin{itemize}
\item $A \dann B$: if $A$ then $B$.
\item $A \wenn B$: $A$ if $B$.
\item $A \gdw B$: $A$ if and only if $B$.
\item $f:\Xcal\to\Ycal$ denotes a function from $\Xcal$ to $\Ycal$.
\item $f(\cdot)$: to clearly distinguish a function $f(x)$ from a function value $f(x)$, we use $f(\cdot)$ for the function and $f(x)$ only for the value of the function $f(\cdot)$ applied to $x$. This is somewhat clearer than using $f$ for the function as (out of context) $f$ could be read as a number. \hide{In $\lambda$-notation we could denote $f(\cdot)$ as $\lambda x.f(x)$.}
\hide{\item $\{x\in\Xcal:p(x)\}$ denotes the set of elements of $\Xcal$ for which the function
$p:\Xcal\to\Omega$ evaluates to true
\item $\{f(\cdot) \mid f:\Xcal\to\Ycal \}$: the set of functions from $\Xcal$ to $\Ycal$
which are defined pointwise, that is, $\forall x \in \Xcal : f(x) \in \Ycal$
Alternative notations are $\{ f:\Xcal\to\Ycal \}$ and $\Ycal^\Xcal$ but we prefer
$\{f(\cdot) \mid f:\Xcal\to\Ycal \}$ for clarity
\item $f : A \to B \to C$: As a shorthand for a function that maps every element
of $A$ to function from $B$ to $C$ that could be denoted by $f : A \to (B \to C)$
we use the notation $f : A \to B \to C$. If arguments $a\in A$ and $b\in B$ are
given at the same time, the notation $f' : A \times B \to C$ could denote the same
function
}
\item $A^\top$ denotes the transpose of the matrix $A$.
\item $|\cdot|$ denotes the function returning the $\ell_1$ norm of a vector.
\item $\|\cdot\|$ denotes the function returning the $\ell_2$ norm of a vector.
\item $\numset{n}$ denotes the set $\{1,\ldots,n\}$.
\end{itemize}
\item Other notational conventions and exceptions:
\begin{itemize}
\item $A_{ij}$ denotes the component in the $i$-th row and $j$-th column of matrix $A$.
\item $A_{i\cdot}$ denotes the $i$-th row vector of matrix $A$.
\item $A_{\cdot j}$ denotes the $j$-th column vector of matrix $A$.
\item $\lambda$ denotes the regularisation parameter.
%\item $\mu$: a measure.
\item $P$ a probability distribution.
\item $\R$ the set of all real numbers.
\item $\N$ the set of all natural numbers $1,2,3,\ldots$.
%\item $\Omega$: the Booleans $\Omega = \{\top,\bot\}$.
\end{itemize}
\end{itemize}

\clearemptydoublepage

\pagenumbering{arabic}

% chapters
% hack
\lhead[\fancyplain{}{\small\sl\thepage}]%
      {\fancyplain{}{\small\sl\rightmark}}
\rhead[\fancyplain{}{\small\sl\leftmark}]%
      {\fancyplain{}{\small\sl\thepage}}

\chapter{Introduction}\label{ch:intro}
\begin{figure}[tb]
\begin{center}
\epsfig{file=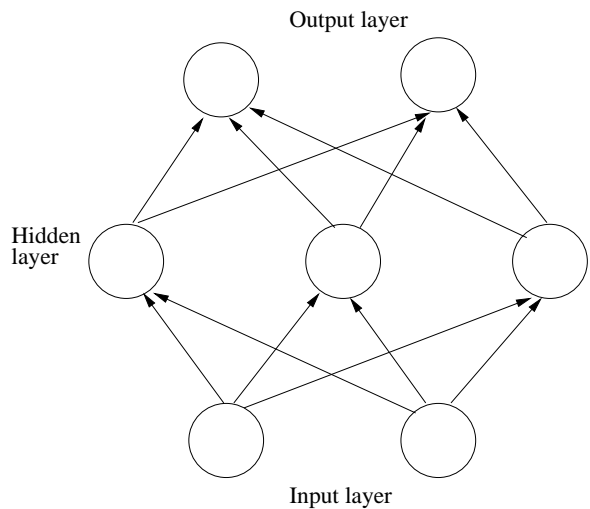,width=0.6\textwidth}
\end{center}
\caption{A multi-layered perceptron with input, output, and hidden layers.}
\label{fig:nn}
\end{figure}

The discipline of machine learning \cite{Mitchell06} has come a long way since Frank Rosenblatt invented the perceptron in 1957. The perceptron is a linear classifier: it takes a sequence of features as its input, computes a linear combination of the features and a sequence of weights, feeds the result into a step function (also known as activation function), and outputs a binary value ($0$ or $1$). A non-linear classifier can be designed using multiple layers of computational units (neurons) with non-linear activation function such as the sigmoid function resulting in what is called a multi-layered perceptron (MLP) or a feed-forward network (see Figure~\ref{fig:nn}). An MLP is a universal function approximator \cite{Cybenko89}, i.e, any feed-forward network with a single hidden layer comprising a finite number of units can be used to approximate any function. The weights of an MLP are \emph{learned} from a given set of training examples using a procedure called backpropagation \cite{Rumelhart/etal/86} which typically modifies the weights iteratively based on the error incurred on the current training example, with training examples being fed into the network in a sequential manner. We thus have a machinery that is able to learn from \emph{experience} and can be used for prediction thereby mimicking human behaviour (to a certain extent).

\begin{figure}[tb]
\begin{center}
\epsfig{file=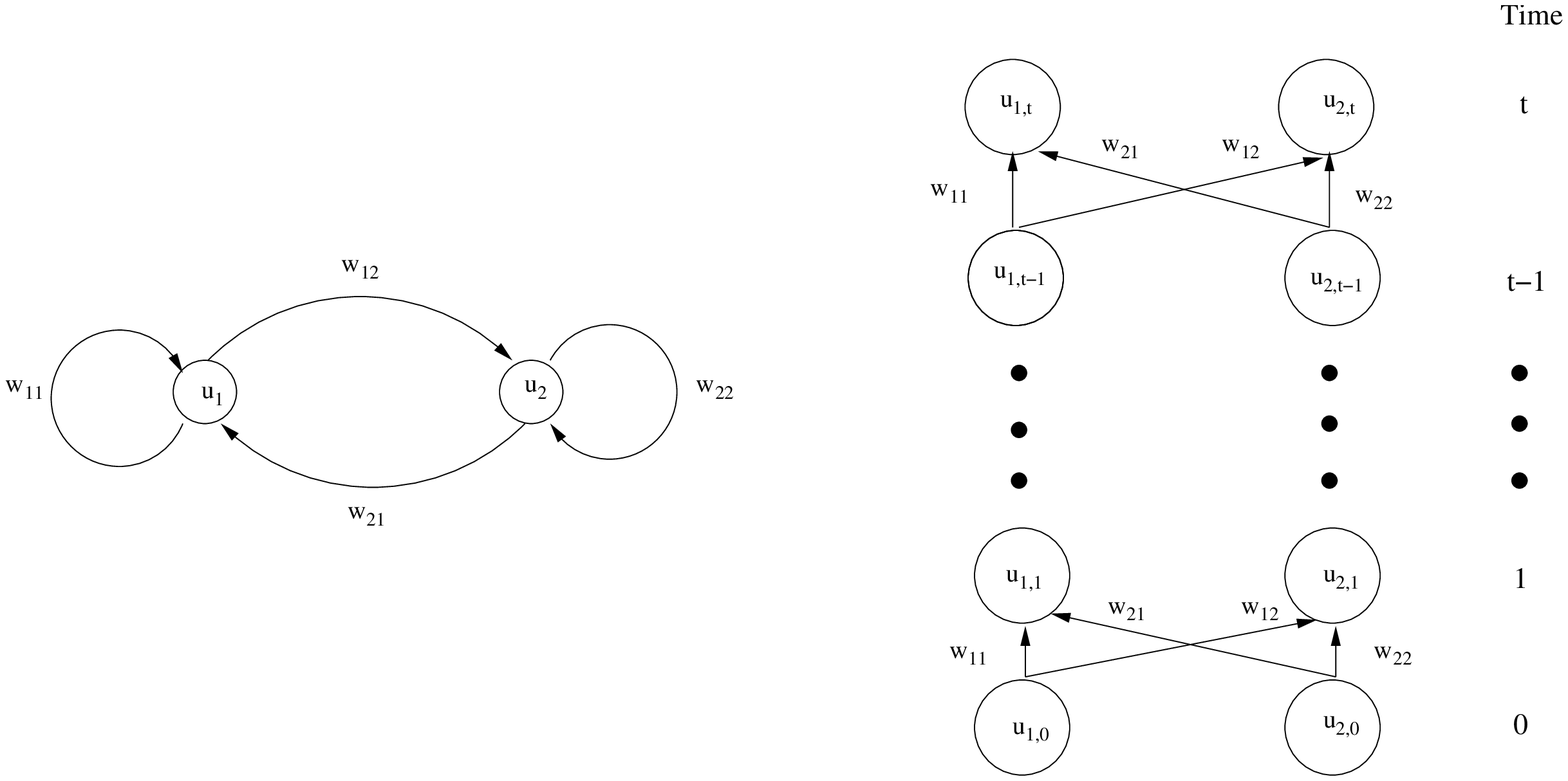,width=\textwidth}
\end{center}
\caption{A fully-connected recurrent neural network with two units, $u_1$ and $u_2$, and its equivalent feed-forward network \cite{Rumelhart/etal/87}. At every time step, a copy of the network is created. Thus the network unfolds in time and standard backpropagation can be used to train the network. As long as there is a constraint that the weights be the same for all the copies of the network over all the time steps, the behaviour of the feed-forward network will be equivalent to that of the recurrent network \cite{Rumelhart/etal/87}. The amount of contextual information used depends on the number of copies retained while training.}
\label{fig:rnn}
\end{figure}

This thesis is concerned with structured prediction --- \emph{the problem of predicting multiple outputs with complex internal structure and dependencies among them}. One of the classical structured prediction problems is temporal pattern recognition with applications in speech and handwriting recognition. The MLP as described above can be used to approximate a multi-valued function using multiple units in the output layer. The downside, however, is that it cannot model dependencies between the outputs since there are no connections between units within a layer. Recurrent neural networks or feedback networks \cite{Jordan86, Elman90, Hochreiter/Schmidhuber/97} address this problem by introducing feedback connections between units (see Figure~\ref{fig:rnn}). The feedback mechanism can be seen as introducing \emph{memory} into the network which makes the network particularly suitable to solve temporal pattern recognition problems\footnote{Temporal structure in data can also be modeled using a feed-forward network known as time delay neural network \cite{Waibal/etal/89}.}. Despite the fact that artificial neural networks can be used to handle temporal sequences, statistical models like hidden Markov models (HMM) \cite{Rabiner89} have enjoyed significant success in adoption (including commercial use) in speech recognition problems as they can be trained efficiently using procedures like the expectation-maximisation algorithm \cite{Dempster/etal/77}. An HMM is a dynamic Bayesian network that models the joint distribution of inputs and outputs by placing a Markovian assumption on the input sequence. \hide{The reader is referred to the works of \citet{Bilmes99,Sha07,Keshet07} that address some of the limitations of HMMs.}

As evidenced above, structured prediction and algorithms pertaining to it have existed since the mid-80s. Later, with the introduction of support vector machines (SVM) in the 90s \cite{Boser/etal/92,Cortes/Vapnik/95}, there has been a flurry of activity in formulating and solving every conceivable machine learning problem using tools from convex optimisation. Unsurprisingly, this has also had an effect on research in structured prediction resulting in several algorithmic developments including models/algorithms\footnote{Typically, a machine learning algorithm is a mechanism to estimate the parameters of an underlying model of the given data. We use the terms model and algorithm interchangeably.} like conditional random fields \cite{Lafferty/etal/01}, max-margin Markov networks \cite{Taskar/etal/03,Taskar/etal/05}, and structured SVMs \cite{Tsochantaridis/etal/05}. The contributions of this thesis follow this line of research in the following sense: 
\begin{quote}
we address some of the limitations of recent structured prediction algorithms when dealing with the specific problem of predicting combinatorial structures by proposing new techniques that will aid in the design and analysis of novel algorithms for structured prediction.
\end{quote}

\section{Structured Prediction}
We begin with a non-technical introduction to structured prediction focusing particularly on applications. A formal algorithmic treatment is deferred until the next chapter. 

We restrict ourselves to \emph{supervised learning} problems where training examples are provided in (input, output) pairs. This is in contrast to other machine learning problems like density estimation and dimensionality reduction that fall under the category of \emph{unsupervised learning}. More formally, let $\Xcal$ be the input space. For example, in OCR applications such as handwriting recognition, this space could represent the set of all possible digits and letters including transformations like scaling and rotation. Each element $x \in \Xcal$ is represented as a sequence of features (e.g., image pixels) in $\R^n$. Let $\Ycal$ be the discrete space\footnote{If the space is continuous, then it is a regression problem. We are only concerned with discrete output spaces.} of all possible outcomes. In handwriting recognition, this space could be the set of possible outcomes, i.e., digits `0'--`9' and letters `a'--`z'. The goal of a supervised learning algorithm is to learn a hypothesis (function) $f$ that maps all elements of the input space to all possible outcomes, i.e., $f : \Xcal \to \Ycal$. Typically, we fix the hypothesis space (decision trees, neural networks, SVMs), parameterise it, and \emph{learn} or estimate these parameters from a given set of training examples $(x_1,y_1), \ldots, (x_m,y_m) \in \Xcal \times \Ycal$, which are all drawn independently from an identical distribution (i.i.d.) $P$ over $\Xcal \times \Ycal$. The parameters are estimated by minimising a pre-defined loss function $\ell: \Ycal \times \Ycal \to \R$ --- such as the $0-1$ loss or the squared loss --- on the training examples with the hope of obtaining a small loss when predicting the outputs of unseen instances. Often, the hypothesis space is further restricted using a mechanism known as \emph{regularisation} so that the learned hypothesis performs well (w.r.t. the loss function) not only on training but also on unseen examples. This property of a learning algorithm is called \emph{generalisation}.

In structured prediction, the output space is \emph{complex} in the following sense: (i) there are dependencies between and internal structure among the outputs, and (ii) the size of the output space is exponential in the problem's input. As a simple example, consider multi-label classification where the goal is to predict, for a given input example, a subset of labels from among a set of pre-defined labels of size $d$. Clearly, the size of the output space $\Ycal$ is $2^d$. A reduction from multi-label to multi-class prediction may not yield good results as it does not take the correlations between labels into account \cite{McCallum99,Schapire/Singer/00}.

\subsubsection{Applications}
\begin{figure}[tb]
\begin{center}
\epsfig{file=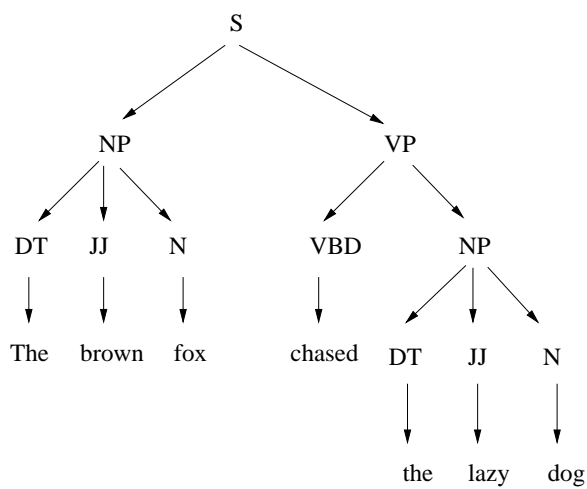,width=0.6\textwidth}
\end{center}
\caption{Illustration of parsing. The input is a sentence and the output is its parse tree.}
\label{fig:parse}
\end{figure}

Natural language processing (NLP) has always been a driving force behind research in structured prediction. Some of the early algorithmic developments in (discriminative) structured prediction \cite{Collins02} were motivated by NLP applications. Part-of-speech tagging is a classical example where the goal is to mark (tag) the words in a sentence with their corresponding parts-of-speech. An attempt to perform this tagging operation by treating the words independently would discard important contextual information. Linguistic parsing is the process of inferring the grammatical structure and syntax of a sentence. The output of this process is a parse tree, i.e, given a sentence of words, the goal is to output its most likely parse tree (see Figure~\ref{fig:parse}). 
Machine translation, which is the problem of translating text in one natural language into another, is another application where structured predition algorithms have been successfully applied \cite{Liang/etal/06}. The reader is referred to the works of \citet{Taskar/04} and \citet{Daume06} for more details on structured prediction applications in NLP.

Another group of applications is based on graph matching. A matching in a graph is a set of edges without common vertices. Finding a matching that contains the largest possible number of edges in bipartite graphs, also known as \emph{maximum bipartite matching}, is a fundamental problem in combinatorial optimisation with applications in computer vision. For example, finding a correspondence between two images is a graph matching problem and was recently cast as a structured prediction problem \cite{Caetano/etal/09}. Segmenting three-dimensional images obtained from robot range scanners into object categories is an important task for scene understanding, and was recently solved using a structured prediction algorithm \cite{Anguelov/etal/05}. Imitation learning is a learning paradigm where the learner tries to mimic the behaviour of an expert. In robotics, this type of learning is useful in planning and structured prediction has been successfully used to solve such problems \cite{Ratliff/etal/06}. The reader is referred to the works of \citet{Taskar/04} and \citet{Ratliff09} for more details on structured prediction applications in robotics. 

Further applications in bioinformatics and computational biology can be found in the works of \citet{Sonnenburg08} and \citet{Frasconi/Passerini/08}.

\section{Why Predict Combinatorial Structures?}\label{sc:whypredictcs}
We have already seen that some of the applications described in the previous section involve predicting combinatorial structures such as permutations in maximum bipartite matching and trees in linguistic parsing. Furthermore, several existing, well-studied machine learning problems can be formulated as predicting combinatorial structures.
\\\\
\noindent\textbf{Multi-label classification} \cite{Schapire/Singer/98,Schapire/Singer/00,Elisseeff/Weston/01,Fuernkranz/etal/08} is a generalisation of multi-class prediction where the goal is to predict a set of labels that are relevant for a given input. The combinatorial structure corresponding to this problem is the set of vertices of a hypercube.
\\\\
\noindent\textbf{Multi-category hierarchical classification} \cite{Cesa-Bianchi/etal/06,Rousu/etal/06} is the problem of classifying data in a given taxonomy when predictions associated with multiple and/or partial paths are allowed. A typical application is taxonomical document classification where document categories form a taxonomy. The combinatorial structure corresponding to this problem is the set of subtrees of a directed, rooted tree.
\\\\
\noindent\textbf{Label ranking} \cite{Dekel/etal/03} is an example of a complex prediction problem where the goal is to not only predict labels from among a finite set of pre-defined labels, but also to rank them according to the nature of the input. A motivating application is again document categorisation where categories are topics (e.g., sports, entertainment, politics) within a document collection (e.g., news articles). It is very likely that a document may belong to multiple topics, and the goal of the learning algorithm is to order (rank) the relevant topics above the irrelevant ones for the document in question. The combinatorial structure corresponding to this problem is the set of permutations.

\subsubsection{Real-world Applications}
In this thesis, we focus particularly on the problem of predicting combinatorial structures such as cycles, partially ordered sets, permutations, and other graph classes. There are several applications where predicting such structures is important. Consider route prediction for hybrid vehicles --- the more precise the prediction of routes, the better the optimisation of the charge/discharge schedule, resulting in significant reduction of energy consumption \cite{froehlich_route_2008}. Route prediction corresponds to prediction of cycles in a street network. The input space $\Xcal$ would be a set of properties of people, situations, etc.; the output space $\Ycal$ would be the set of cycles over the places of interest; and $y_i$ are the known tours of people $x_i$. Route prediction could also find interesting applications in the design of intelligent personal digital assistants that are smart enough to recommend alternative routes or additional places to visit.

As another application, consider de novo construction of (personalised) drugs from the huge space of synthesisable drugs (e.g., a fragment space) \cite{MauserH} --- better predictions lead to more efficient entry of new drugs into the market. The task here is to predict graphs (molecules) on a fixed set of vertices. State-of-the-art software systems to support drug design are\hide{in silico} virtual screening methods predicting the properties of database compounds. The set of molecules that can be synthesised is, however, orders of magnitude larger than what can be processed by these methods. In this application, $\Xcal$ would be some set of properties; $\Ycal$ would be a particular set of graphs over, say, functional groups; and $y_i$ are the compounds known to have properties $x_i$.

\hide{\todo{}graph cuts, partitions, maximum weight perfect matching, bipartite matchings, spanning trees \cite{Taskar/04}}

\section{Goals and Contributions}
The main goal of this thesis is to design and analyse machine learning algorithms for predicting combinatorial structures. This problem is not new and there exists algorithms \cite{Collins02,Taskar/etal/03,Taskar/04,Taskar/etal/05,Tsochantaridis/etal/05} to predict structures such as matchings, trees, and graph partitions. Therefore, as a starting point, we investigate the applicability of these algorithms for predicting combinatorial structures that are of interest to us. It turns out that the assumptions made by these algorithms to ensure efficient learning do not hold for the structures and applications we have in mind. We elucidate the limitations of existing structured prediction algorithms by presenting a complexity theoretic analysis of them. We then introduce two novel assumptions based on counting and sampling combinatorial structures, and show that these assumptions hold for several combinatorial structures and complex prediction problems in machine learning. The consequences of introducing these two assumptions occupy a major portion of this work and are briefly described below.

\subsection*{A New Algorithm for Structured Prediction}
We present an algorithm that can be trained by solving an unconstrained, polynomially-sized quadratic program. The resulting algorithmic framework is a generalisation of the classical regularised least squares regression, also known as ridge regression, for structured prediction. The framework can be instantiated to solve several machine learning problems, including multi-label classification, ordinal regression, hierarchical classification, and label ranking. We then design approximation algorithms for predicting combinatorial structures. We also present empirical results on multi-label classification, hierarchical classification and prediction of directed cycles. Finally, we address the scalability issues of our algorithm using online optimisation techniques such as stochastic gradient descent.

\subsection*{Analysis of Probabilistic Models for Structured Prediction}
Maximum a posteriori estimation with exponential family models is a cornerstone technique used in the design of discriminative probabilistic classifiers. One of the main difficulties in using this technique for structured prediction is the computation of the partition function. The difficulty again arises from the exponential size of the output space. We design an algorithm for approximating the partition function and the gradient of the log partition function with provable guarantees using classical results from Markov chain Monte Carlo theory \cite{Jerrum/etal/86,Sinclair/Jerrum/96,Randall/03}. We also design a Markov chain that can be used to sample combinatorial structures from exponential family distributions, and perform a non-asymptotic analysis of its mixing time. These results can be applied to solve several learning problems, including but not limited to multi-label classification, label ranking, and multi-category hierarchical classification.

\section{Thesis Outline}
The thesis is organised as follows:
\begin{description}
\item{Chapter 2} serves as an introduction to structured prediction. We begin with a description of generative and discriminative learning paradigms followed by a detailed exposition of several machine learning algorithms that exist in the literature for structured prediction.
\item{Chapter 3} is a survey, including original contributions, on algorithms for predicting a specific combinatorial structure --- permutations. \hide{This problem has recently attracted a lot of attention resulting in several algorithms.}
\item{Chapter 4} analyses existing discriminative structured prediction algorithms through the lens of computational complexity theory. We study the assumptions made by existing algorithms and identify their shortcomings in the context of predicting combinatorial structures. The study will consequently motivate the need to introduce two new assumptions --- the counting and the sampling assumption. We provide several examples of combinatorial structures where these assumptions hold and also describe their applications in machine learning.
\item{Chapter 5} proposes a new learning algorithm for predicting combinatorial structures using the counting assumption. The algorithm is a generalisation of the classical ridge regression for structured prediction.
\item{Chapter 6} analyses probabilistic discriminative models for structured prediction using the sampling assumption and some classical results from Markov chain Monte Carlo theory.
\item{Chapter 7} summarises the contributions of this thesis and points to directions for future research.
\end{description}

\section{Bibliographical Notes}
Parts of the work described in this thesis appear in the following publications:
\begin{itemize}
\item  Thomas G\"artner and Shankar Vembu. On Structured Output Training: Hard Cases and an Efficient Alternative.  \emph{Machine Learning Journal}, 76(2):227–242, 2009. Special Issue of the European Conference on Machine Learning and Principles and Practice of Knowledge Discovery in Databases 2009.
\item Shankar Vembu, Thomas G\"artner, and Mario Boley. Probabilistic Structured Predictors. In \emph{Proceedings of the 25th Conference on Uncertainty in Artificial Intelligence}, 2009.
\item Shankar Vembu and Thomas G\"artner. Label Ranking Algorithms: A Survey. To appear in a book chapter on \emph{Preference Learning}, Johannes Fürnkranz and Eyke Hüllermeier (Editors), Springer-Verlag, 2010.
\item Thomas G\"artner and Shankar Vembu. Learning to Predict Combinatorial Structures. In \emph{Proceedings of the Workshop on Structured Inputs and Structured Outputs at the 22nd Annual Conference on Neural Information Processing Systems}, 2008.
\item Shankar Vembu and Thomas G\"artner. Training Non-linear Structured Prediction Models with Stochastic Gradient Descent. In \emph{Proceedings of the 6th International Workshop on Mining and Learning with Graphs}, 2008.
\end{itemize}

\clearemptydoublepage
\chapter{Structured Prediction} \label{ch:prelims}
In supervised learning, the goal is to approximate an unknown target function $f : \Xcal \to \Ycal$ or to estimate the conditional distribution $p(y\mid x)$. \hide{We assume that there is an underlying joint distribution on $\Xcal \times \Ycal$ from which a set of $m$ iid training examples $S = \{(x_i,y_i)\}_{i=1}^m$ is drawn.}There are essentially two ways to define a learning model and estimate its parameters. In what is known as generative learning, a model of the joint distribution $p(x,y)$ of inputs and outputs is learned, and then Bayes rule is used to predict outputs by computing the mode of
\begin{equation*}
p(y\mid x) = \frac{p(x,y)}{\sum\limits_{z \in \Ycal} p(x,z)} \ .%= \frac{p(x \mid y)p(y)}{\sum\limits_{z \in \Ycal} p(x \mid z)p(z)} \ .
\end{equation*}
The above is also known as Bayes classifier. In discriminative learning, the goal is to directly estimate the parameters of the conditional distribution $p(y \mid x)$. According to \citet{Vapnik95}, one should always solve a problem directly instead of solving a more general problem as an intermediate step, and herein lies the common justification to model the conditional instead of the joint distribution. This has led to an upsurge of interest in the design of discriminative learning algorithms for structured prediction. Prominent examples include conditional random fields \cite{Lafferty/etal/01}, structured perceptron \cite{Collins02}, max-margin Markov networks \cite{Taskar/etal/03,Taskar/etal/05}, and support vector machines \cite{Tsochantaridis/etal/05}. It has now become folk wisdom to attack a learning problem using discriminative as opposed to generative methods. An interesting theoretical and empirical study was performed by \citet{Ng/Jordan/01} who showed that while discriminative methods have lower asymptotic error, generative methods approach their higher asymptotic error much more faster. This means that generative classifiers outperform their discriminative counterparts when labeled data is scarce.

A plethora of algorithms have been proposed in recent years for predicting structured data. The reader is referred to \citet{Bakir/etal/07} for an overview. In this chapter, we discuss several of these algorithms, in particular those that are relevant to and motivated the contributions of this thesis. The main goal of a learning algorithm is to minimise an appropriately chosen risk (loss) function depending on the problem or application. We describe several loss functions for binary classification and their extensions to structured prediction. The algorithmic contributions of this thesis fall under the category of discriminative learning. We therefore review classical approaches to discriminatively learning a classifier using perceptron, logistic regression and support vector machine, and show that many of the recently proposed structured prediction algorithms are natural extensions of them.

\hide{
We now highlight the differences between generative and discrminative classifiers using naive Bayes classifier and logistic regression as examples.
\subsubsection{Naive Bayes Classifier}
We follow \citet{Mitchell05}. For the sake of simplicity, consider boolean inputs and outputs, i.e., $\Xcal \subseteq \{0,1\}^d$ and $\Ycal = \{0,1\}$. Let $w_{ij} = P(X=x_i) \mid Y=y_j$, where $i$ is an index on the input dimension and $j$ is the number of outputs ($=2$), be the set of parameters that are to be estimated. Clearly, the number of parameters is exponential in the input dimension $d$ and therefore we need at least as many training examples to obtain reliable maximum-lilelihood parameter estimates. Therefore,  we need to make further assumptions on $P(X|Y)$ in order to use a Bayes classifier in practice, and the commonly used assumption is that of conditional independence of features which is to say that the distribution $P(X|Y)$ factories into $\prod_i P(X_i|Y)$. Consequently, the number of parameters to be estimated when modeling $P(X|Y) $has been reduced dramatically from $2(2^d-1)$ to $2d$. The two sets of parameters are given as
\begin{equation*}
w_{ijk} = P(X_i = x_{ij} | Y = y_k), \quad \text{and } \pi_k = P(Y=y_k) \ .
\end{equation*}
These parameters are estimated using maximum likelihood estimation (or by using maximum a posteriori estimation by defining a prior over the parameters). For count data, The MLE estimate of the parameters are given as 
\begin{equation*}
\hat{w}_{ijk} = \hat{P}(X_i = x_{ij} | Y = y_k) = \frac{\#D\{X_i = x_{ij} \wedge Y = y_k\}}{\#D\{Y=y_k\}} \ ,
\end{equation*}
and 
\begin{equation*}
\hat{\pi}_k = \hat{P}(Y=y_k) = \frac{\#D\{Y = y_k\}}{|D|} \ ,
\end{equation*}
where $\#D\{x\}$ returns the number of elements in $D$ that satisfy property $x$.

\subsubsection{Logistic Regression}
Logistic regression is a probabilistic binary classifier. The probability of class label $y$ given an input $x$ is modeled using exponential families
\begin{equation*} \label{eq:logreg}
p(y|x,w) = \frac{\exp(y\ip{\phi(x)}{w})} {\exp(y\ip{\phi(x)}{w}) + \exp(-y\ip{\phi(x)}{w})} \ .
\end{equation*}
Given a set of observationstraining sequence $X=(x_1,\cdots,x_m) \in \Xcal^m$, \mbox{$Y=(y_1,\cdots,y_m) \in \Ycal^m$}, the parameters $w$ can be estimated using the maximum (log) likelihood principle
\begin{equation*}
\begin{aligned}
\hat{w} & = \argmax\limits_{w} \left[ \ln p(Y|X,w)\right] \\
& = \argmax\limits_{w} \left[ \frac{1}{m} \sum\limits_{i=1}^m p(y_i | x_i, w) \right] \ .
\end{aligned}
\end{equation*}
}
\section{Loss Functions}
Given an input space $\Xcal$, an output space $\Ycal$, a probability distribution $P$ over $\Xcal \times \Ycal$, a loss function $\ell(\cdot,\cdot)$ maps pairs of outputs to a quantity that is a measure of ``discrepancy" between these pairs, i.e., $\ell: \Ycal \times \Ycal \to \R$. The goal of a machine learning algorithm is to minimise the true risk
\begin{equation*}
\Rcal_{\text{true}}(f)  = \int\limits_{\Xcal \times \Ycal} \ell(f(x),y) dP(x,y) \ .
\end{equation*}
Since we do not have any knowledge of the distribution $P$, it is not possible to minimise this risk. But given a set of training examples $\{(x_i,y_i)\}_{i=1}^m$ drawn independently from $\Xcal \times \Ycal$ according to $P$, we can minimise an approximation to the true risk known as the empirical risk
\begin{equation*}
\Rcal_{\text{emp}}(f)  = \sum\limits_{i=1}^m \ell(f(x_i),y_i) \ .
\end{equation*}

In structured prediction, a \emph{joint scoring function} on input-output pairs is considered, i.e., $f : \Xcal \times \Ycal \to \R$ (with an overload of notation), where the score is a measure of ``affinity" between inputs and outputs. Analogously, a joint feature representation of inputs and outputs $\phi: \Xcal \times \Ycal \to \R^n$ is considered. A linear scoring function parameterised by a weight vector $w \in \R^n$ is defined as 
\begin{equation*}
f(x,y) = \ip{w}{\phi(x,y)} \ .
\end{equation*}
The goal of a structured prediction algorithm is to learn the parameters $w$ by minimising an appropriate \emph{structured} loss function. Given a test example $x \in \Xcal$, the output is predicted as
\begin{equation} \label{eq:argmax}
\hat{y} = \argmax_{z \in \Ycal} {f(x,z)} = \argmax_{z \in \Ycal} \ip{w}{\phi(x,z)} \ .
\end{equation}
One of the major challenges in designing a structured prediction algorithm is to solve the above ``argmax problem". The difficulty arises in non-trivial structured prediction applications due to the exponential size of the output space. \hide{The solution to this problem depends very much on the loss function used and also on the structure of the output space.} 

In the following, we describe commonly used loss functions in classification and regression and extend them to structured prediction.

\subsubsection{Zero-One Loss}
The zero-one loss is defined as
\begin{equation*}
\ell_{\text{0-1}}(y,z) = 
\begin{cases}
0 & \text{if}~ y=z\\
1 &\text{otherwise}
\end{cases} \ .
\end{equation*}
It is non-convex, non-differentiable, and optimising it is a hard problem in general. Therefore, it is typical to consider approximations (surrogate losses) to it, for instance, by upper-bounding it with a convex loss such as the hinge loss\footnote{The hinge loss is non-differentiable, but learning algorithms like support vector machines introduce slack variables to mitigate this problem. Support vector machines will be described in the next section.}.

\subsubsection{Squared Loss}
The squared loss is commonly used in regression problems with $\Ycal \subseteq \R$ and is defined as
\begin{equation*}
\ell_{\text{square}}(y,z) = (y-z)^2 \ .
\end{equation*}
The extension of this loss function for structured prediction is non-trivial due to inter-dependencies among the multiple output variables. However, these dependencies can be removed by performing (kernel) principal component analysis \cite{Scholkopf/etal/97} on the output space and by subsequently learning separate regression models on each of the independent outputs. The final output can be predicted by solving a pre-image problem that maps the output in the transformed space back to the original output space. Thus, a structured prediciton problem can be reduced to regression. This technique is called kernel dependency estimation \cite{Weston/etal/02}.

\subsubsection{Hinge Loss}
The hinge loss became popular with the introduction of support vector machines \cite{Cortes/Vapnik/95}. For binary classification, it is defined as
\begin{equation*}
\ell_{\text{hinge}}(y,z) = \max(0, 1-yz) \ ,
\end{equation*}
where $y \in \{-1,+1\}$ and $z \in \R$ is the output of the classifier. The generalisation of hinge loss for structured prediction \cite{Taskar/etal/03,Tsochantaridis/etal/05} is defined with respect to a hypothesis $f$ and a training example $(x,y)$.  Let $\Delta:\Ycal\times\Ycal\to\R$ be a discrete (possibly non-convex) loss function, such as the Hamming loss or the zero-one loss, defined on the output space. Consider the loss function
\begin{equation*}
\ell^\Delta_{\max}(f,(x,y)) = \Delta (\argmax_{z \in \Ycal} f(x,z), y) \ .
\end{equation*}
To ensure convexity, the above loss function is upper-bounded by the hinge loss as 
\begin{equation} \label{eq:struct_hinge}
\ell^\Delta_\mathrm{hinge} (f,(x,y)) = \max_{z\in\Ycal} \left[ \Delta (z, y) +f(x,z) - f(x,y)\right] \ .
\end{equation}

\subsubsection{Logistic Loss}
The logistic loss is used in probabilistic models and is a measure of the negative conditional log-likelihood, i.e., $-\ln p(y\mid x)$. For binary classification (cf. logistic regression) it is defined as follows:
\begin{equation*}
\ell_{\log}(y,z) = \ln(1+ \exp(-yz)) \ .
\end{equation*}
For structured prediction, it is defined (again w.r.t. to a hypothesis $f$ and a training example $(x,y)$) as
\begin{equation*} 
\ell_{\log} (f,(x,y)) =  \ln \left[ \sum_{z \in \Ycal} \exp(f(x,z))\right] - f(x,y)  \ .
\end{equation*}

\subsubsection{Exponential Loss}
\begin{figure}[tb]
\begin{center}
\epsfig{file=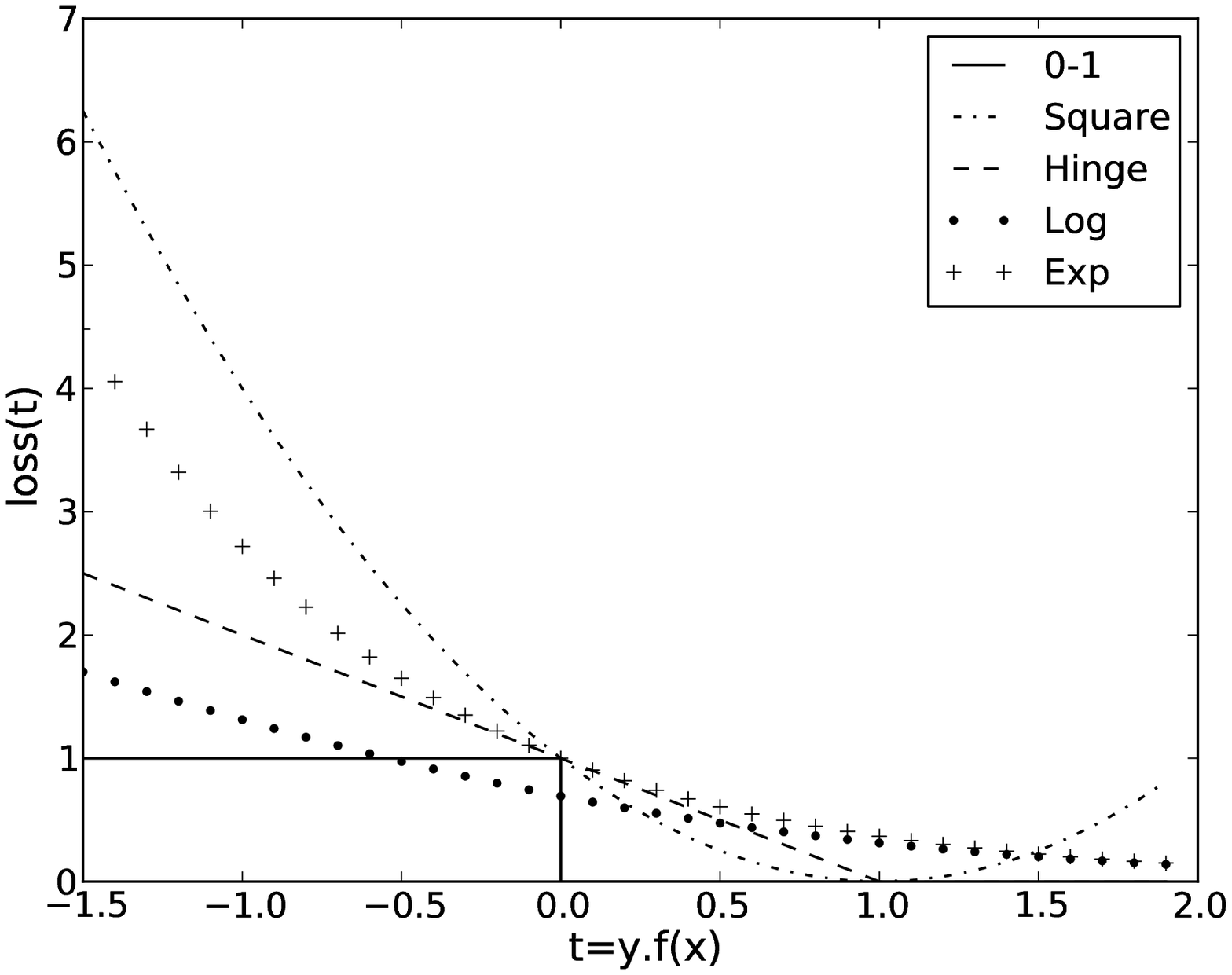,width=0.75\textwidth}
\end{center}
\caption{Various loss functions for binary classification.}
\label{fig:bloss}
\end{figure}

The exponential loss for binary classification is defined as
\begin{equation*}
\ell_{\exp}(y,z) = \exp(-yz) \ .
\end{equation*}
As shown in Figure~\ref{fig:bloss}, the exponential loss imposes a heavier penalty on incorrect predictions than the logistic loss. However, this also means that the exponential loss is sensitive to label noise. The exponential loss for structured prediction \cite{Altun/etal/03} is defined as
\begin{equation*}
\ell_{\exp} (f,(x,y)) = \sum_{z\in\Ycal} \exp \left[ f(x,z) - f(x,y)  \right] \ .
\end{equation*}
We will revisit this loss in Chapter~\ref{ch:csop}.

\section{Algorithms} \label{sc:algorithms}
\subsection*{Perceptron}
The perceptron {\cite{Rosenblatt58}, briefly described in the introductory chapter, is a simple online learning algorithm. It learns a linear classifier parameterised by a weight vector $w$ and makes predictions according to $\hat{y} = f(x) = \sgn(\ip{w}{x})$. The algorithm operates in rounds (iterations). In any given round, the learner makes a prediction $\hat{y}$ for the current instance $x$ using the current weight vector. If the prediction differs from the true label $y$ (revealed to the algorithm after it has made the prediction), then the weights are updated according to $w \gets w + yx$. The weights remain unchanged if the learner predicts correctly. If the data are linearly separable, then the perceptron makes a finite number of mistakes \cite{Block62,Novikoff62,Minsky/Papert/69} and therefore if the algorithm is presented with the training examples iteratively, it will eventually converge to the true solution, which is the weight vector that classifies all the training examples correctly.
\begin{theorem} \cite{Block62,Novikoff62}
Let $(x_1,y_1), \dots, (x_m,y_m)$ be a sequence of training examples with $\|x_i\| \leq R$ for all $i \in \numset{m}$. Suppose there exists a unit norm vector $u$ such that $y_i(\ip{u}{x_i}) \geq \gamma$ for all the examples. Then the number of mistakes made by the perceptron algorithm on this sequence is at most $(R /\gamma)^2$.
\end{theorem}

If the date is not separable, then we have the following result due to \citet{Freund/Schapire/99}.
\begin{theorem} \cite{Freund/Schapire/99}
Let $(x_1,y_1), \dots, (x_m,y_m)$ be a sequence of training examples with $\|x_i\| \leq R$ for all $i \in \numset{m}$. Let $u$ be any unit norm weight vector and let $\gamma > 0$. Define the deviation of each example as $d_i = \max(0, \gamma - y_i (\ip{u}{ x_i}))$ and define $D = \sqrt{\sum_{i=1}^m d_i^2}$. Then the number of mistakes made by the perceptron algorithm on this sequence of examples is at most 
\begin{equation*}
\frac{(R + D)^2}{\gamma^2} \ .
\end{equation*}
\end{theorem}

The perceptron can be extended to learn non-linear functions using the kernel trick \cite{Schoelkopf/Smola/02}. The resulting algorithm called kernel perceptron \cite{Freund/Schapire/99} learns functions of the form
\begin{equation*}
f(\cdot) = \sum\limits_{i=1}^m c_i k(x_i,\cdot) \ ,
\end{equation*}
where $k : \Xcal \times \Xcal \to \R$ is a reproducing kernel function\footnote{A reproducing kernel $k$ is a function with the following two properties: (i) for every $x \in \Xcal$, the function $k(x,\cdot)$ is an element of a Hilbert space $\Hcal$, (ii) for every $x \in \Xcal$ and every function $f(\cdot) \in \Hcal$, the reproducing property, $\ip{k(x,\cdot)}{f(\cdot)} = f(x)$, holds.} on the inputs and $\{c_i\}_{i \in \numset{m}}$ are the kernel expansion coefficients. For every reproducing kernel, there exists a function $\phi: \Xcal \to \Hcal$ (the high-dimensional, possibly infinite, feature space) such that $k(x,x') = \ip{\phi(x)}{\phi(x')}$. Thus any learning algorithm whose function can be represented as a linear combination of inner products can use the \emph{kernel trick} to avoid explicit computations of these inner products in the high-dimensional feature space. The kernel perceptron starts by setting all coefficients $c_i$ to $0$. It then operates in rounds, similar to the perceptron, by repeatedly cycling through the data. If the algorithm makes a mistake on a particular instance, then the corresponding coefficient is updated according to $c_i \gets c_i + y_i$. The algorithm stops when it classifies all training instances correctly. The convergence results of the perceptron can be extended to the kernelised case \cite{Gaertner/05}.
\begin{theorem} \cite{Gaertner/05}
Let $(x_1,y_1), (x_2,y_2), \dots, (x_m,y_m)$ be a sequence of training examples. Let $k : \Xcal \times \Xcal \to \R$ be a kernel function such that $k(x_i,x_i) \leq R$ for all $i \in \numset{m}$. Let \mbox{$f^*(\cdot) = \sum_{j=1}^m c_j k(x_j,\cdot)$} be a function that classifies all the training instances correctly. Suppose there exists a margin $\gamma$ such that $y_i\sum_{j=1}^m c_j k(x_j,x_i) > \gamma$ for all $i \in \numset{m}$. Then the number of mistakes made by the kernel perceptron algorithm on this sequence of examples is at most
\begin{equation*}
\frac{R \|f^*(\cdot)\|^2_{\Hcal}}{\gamma^2} \ .
\end{equation*}
\end{theorem}

The perceptron algorithm can be extended to predict structured data \cite{Collins02}. Consider linear scoring functions on input-output pairs $f(x,y) = \ip{w}{\phi(x,y)}$\footnote{Note the use of joint feature representaion of inputs and outputs (with a slight abuse of notation).}. In every iteration, the output structure of an instance $x$ is determined by solving the argmax problem, $\hat{y} = \argmax_{z \in \Ycal} f(x,z)$, using the current weight vector $w$. If this output is different from the true output, i.e., if $y \neq \hat{y}$, then the weight vector is updated as follows:
\begin{equation*}
w \gets w + \phi(x,y) - \phi(x,\hat{y}) \ .
\end{equation*}
The algorithm stops when all the training instances have been predicted with their correct output structures. Similar to the perceptron, convergence results can be established for structured prediction in the separable and inseparable cases \cite{Collins02}. Let $\text{GEN}(x)$ be a function which generates a set of candidate output structures for input $x$ and let $\bar{\text{GEN}}(x) = \text{GEN}(x) - \{y\}$ for a training example $(x,y)$. A training sequence $(x_1,y_1), \dots, (x_m,y_m)$ is said to be separable with a margin $\gamma$ if there exists a unit norm vector $v$ such that for all $i \in \numset{m}$ and for all $ z \in \bar{\text{GEN}}(x_i)$, the following condition holds: $\ip{v}{\phi(x_i,y_i)} - \ip{v}{\phi(x_i,z)} \geq \gamma$.
\begin{theorem}\cite{Collins02}\label{th:struct_perceptron}
Let $(x_1,y_1), (x_2,y_2), \dots, (x_m,y_m)$ be a sequence of training examples which is separable with margin $\gamma$. Let $R$ denote a constant that satisfies $\|\phi(x_i,y_i) - \phi(x_i,z)\| \leq R$, for all $i \in \numset{m}$, and for all $z \in \bar{\text{GEN}}(x_i)$. Then the number of mistakes made by the perceptron algorithm on this sequence of examples is at most  $R^2 / \gamma^2$.
\end{theorem}
For the inseparable case, we need a few more definitions. For an $(x_i,y_i)$ pair, define $m_i = \ip{v}{\phi(x_i,y_i)} - \max_{z \in \bar{\text{GEN}}(x_i) }\ip{v}{\phi(x_i,z)}$ and $\epsilon_i = \max\{0, \gamma-m_i\}$ and define $D_{v, \gamma} = \sqrt{\sum_{i=1}^m \epsilon_i^2}$. Then the number of mistakes made by the structured perceptron was shown by \citet{Collins02} to be at most
\begin{equation*}
\min\limits_{v, \gamma} \frac{(R+D_{v, \gamma})^2}{\gamma^2} \ .
\end{equation*}

\subsection*{Logistic Regression}
Logistic regression is a probabilistic binary classifier. The probability distribution of class label $y \in \{-1,+1\}$ for an input $x$ is modeled using exponential families
\begin{equation*} \label{eq:logreg}
p(y\mid x,w) = \frac{\exp(y\ip{\phi(x)}{w})} {\exp(y\ip{\phi(x)}{w}) + \exp(-y\ip{\phi(x)}{w})} \ .
\end{equation*}
Given a set of training examples $X=(x_1,\cdots,x_m) \in \Xcal^m$, \mbox{$Y=(y_1,\cdots,y_m) \in \Ycal^m$}, the parameters $w$ can be estimated using the maximum (log) likelihood principle by solving the following optimisation problem:
\begin{equation*}
\begin{aligned}
\hat{w} & = \argmax\limits_{w} \left[ \ln p(Y\mid X,w)\right] \\
& = \argmax\limits_{w} \left[ \frac{1}{m} \sum\limits_{i=1}^m p(y_i \mid  x_i, w) \right] \ .
\end{aligned}
\end{equation*}
Observe that the loss function minimised by this model is the logistic loss. Often, a Bayesian approach is taken to estimate the distribution of parameters 
\begin{equation*}
p(w \mid  Y,X) = \frac{p(w,Y\mid X)}{p(X)} = \frac{p(Y\mid X,w)p(w)}{p(X)}
\end{equation*}
and the mode of this distribution is used as a point estimate of the parameter vector. By imposing a Gaussian prior on $w$, $p(w) = \exp(-\lambda \|w\|^2)$, which acts as a regulariser with $\lambda$ being the regularisation parameter, a point estimate can be computed by maximising the joint likelihood in $w$ and $Y$:
\begin{equation*}
\begin{aligned}
\hat{w} & = \argmax\limits_{w} \left[ \ln p(w,Y\mid X)\right] \\
& = \argmax\limits_{w} \left[ \frac{1}{m} \sum\limits_{i=1}^m p(y_i \mid  x_i, w) - \lambda \|w\|^2\right] \ .
\end{aligned}
\end{equation*}
This technique of parameter estimation is known as maximum a posterior (MAP) estimation. The optimisation problem is convex in the parameters $w$ and differentiable, and therefore gradient descent techniques can be applied to find the global optimum solution.

Logistic regression can be extended for structured prediction with the class conditional distribution 
\begin{equation*}
p(y\mid x,w) = \frac{\exp(\ip{\phi(x,y)}{w})} {\sum\limits_{z \in \Ycal} \exp(\ip{\phi(x,z)}{w})} \ .
\end{equation*}
The denominator of the above expresion is known as the partition function $Z(w\mid x)= \sum_{z \in \Ycal} \exp(\ip{\phi(x,z)}{w})$.  Computation of this function is usually intractable for non-trivial structured output spaces, and depends very much on the features $\phi(x,y)$ and the structure of $\Ycal$. 

We now look at a particular structure --- sequences --- that motivated the development of one of the popular conditional probabilistic models for structured prediction --- conditional random fields (CRF) \cite{Lafferty/etal/01}. CRFs offer a viable alternative to HMMs for segmenting and labeleling sequences. Whereas HMMs model the joint distribution of input and outputs $p(x,y)$, CRFs model the conditional $p(y\mid x)$. A CRF is defined as follows \cite{Lafferty/etal/01}:
\begin{definition}
Let $X$ be a random variable over data sequences, of finite length $l$, to be labeled. Let $Y$ be a random variable over corresponding label sequences, where the components $Y_{i}, i \in \numset{l}$ can take values from a finite alphabet $\Sigma$. Let $G = (V, E)$ be a graph such that the vertex set $V$ indexes $Y$, i.e., $Y = (Y_{p})_{p \in V}$. Then $(X,Y)$ is a \emph{conditional random field} if, when conditioned on $X$, the random variables $Y_{p}$ satisfy the Markov property
\begin{equation*}
p(Y_{p} \mid  X, Y_{q}, q \neq p) = p(Y_{p} \mid  X, Y_{q}, q \sim p) \ ,
\end{equation*}
wher $q \sim p$ means that $q$ is a neighbour of $p$ in G.
\end{definition}
In the case of sequences, $G$ is a simple chain (see Figure~\ref{fig:hmm_crf}). The advantage of CRFs over HMMs is that the probability of transition between labels can depend on past and future observations (if available) and not only on the current observation.
\begin{figure}[tb]
\begin{center}
\epsfig{file=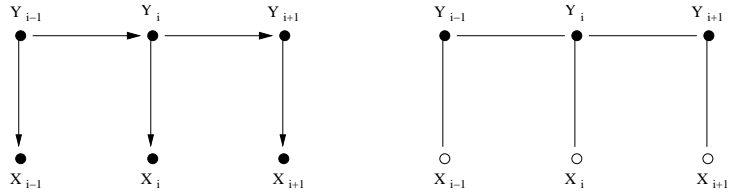,width=.75\textwidth}
\end{center}
\caption{Graphical models of HMMs (left) and CRFs. An open circle indicates that the variable is not generated by the model \cite{Lafferty/etal/01}.}
\label{fig:hmm_crf}
\end{figure}
The partition function for sequences is computationally tractable using dynamic programming techniques similar to the forward-backward algorithm of HMMs \cite{Lafferty/etal/01,Fei/Pereira/03}. Furthermore, CRFs are guaranteed to converge to the optimal solution due to the optimisation problem being convex, whereas HMMs can only guarantee a locally optimum solution using expectation-maximisation for parameter estimation.

\subsection*{Support Vector Machines}
A learning algorithm for binary classification that has attracted considerable interest during the past decade is the support vector machine \cite{Boser/etal/92,Cortes/Vapnik/95}. An SVM learns a hyperplane that separates positive and negative examples with a large margin thereby exhibiting good generalisation abilities \cite{Vapnik95}. It learns a linear function $f(x) = \ip{w}{\phi(x)}$ and minimises the hinge loss by optimising
\begin{equation*}
\min\limits_{w} \lambda \|w\|^2 + \frac{1}{m} \sum\limits_{i=1}^m \max\{0,1-y_i \langle w,\phi(x_i) \rangle\} \ .
\end{equation*}
The above optimisation problem can be rewritten as
\begin{equation} \label{eq:svm}
\begin{array}{ll}
\min\limits_{w, \mathbf{\xi}} & \lambda \|w\|^2 + \frac{1}{m} \sum\limits_{i=1}^m \xi_i \\
\mathrm{s.t.:} & y_i \langle w,\phi(x_i) \rangle \ge 1 - \xi_{i}, \forall i \in \numset{m}\\
& \xi_{i} \ge 0, \forall i \in \numset{m} \ ,
\end{array}
\end{equation}
where $\xi \in \R^m$ are the slack variables which correspond to the degree of misclassification of the training instances. Non-linear functions in the original feature space can be learned by mapping the features to a high-dimensional space and by using the kernel trick. It is computationally more convenient to optimise the Lagrangian dual rather than the primal optimisation problem \eq{eq:svm} due to the presence of box constraints as shown below:
\begin{equation}\label{eq:svm_dual}
\begin{array}{rll}
\min\limits_{c\in\R^m}
& \frac{1}{2}c^\top Y K Y c  -  \one^\top c
&
\medskip \\
\text{s.t.:}
& 0\leq c_i\leq\frac{1}{m\lambda},  \forall i \in \numset{m}
\end{array}
\end{equation}
where $K$ is the kernel matrix with entries $K_{ij}= k(x_i,x_j)$ and $Y$ is a diagonal matrix with entries $Y_{ii}=y_i$. The non-linear predictor can be expressed in the following form due to the representer theorem \cite{Schoelkopf/etal/01}:
\begin{equation*}
f(\cdot) = \sum\limits_{i=1}^m c_i k(x_i,\cdot) \ .
\end{equation*}

SVMs can be extended for structured prediction by minimising the structured hinge loss \eq{eq:struct_hinge} which was first proposed by \citet{Taskar/etal/03}. In structured SVMs \cite{Tsochantaridis/etal/05}, the following optimisation problem is considered:
\begin{equation}\label{eq:StructSVM1}
\begin{array}{ll}
\min\limits_{w,\mathbf{\xi}} & \lambda \|w\|^2 + \frac{1}{m} \sum\limits_{i=1}^m \xi_i \\
\mathrm{s.t.:} & \langle w,\phi(x_i,y_i) \rangle  - \langle w,\phi(x_i,z) \rangle \ge 1 - \frac{\xi_{i}}{\Delta(y_i,z)}, \forall z \in \Ycal \setminus y_i, \forall i \in \numset{m}\\
& \xi_{i} \ge 0, \forall i \in \numset{m} \ .
\end{array}
\end{equation}
Observe that the slack variables have been rescaled. In another formulation, proposed by \citet{Taskar/etal/03}, the margin is rescaled and the resulting optimisation problem is
\begin{equation}\label{eq:StructSVM2}
\begin{array}{ll}
\min\limits_{w,\mathbf{\xi}} & \lambda \|w\|^2 + \frac{1}{m} \sum\limits_{i=1}^m \xi_i \\
\mathrm{s.t.:} & \langle w,\phi(x_i,y_i) \rangle  - \langle w,\phi(x_i,z) \rangle \ge \Delta(y_i,z) - \xi_{i}, \forall z \in \Ycal \setminus y_i, \forall i \in \numset{m}\\
& \xi_{i} \ge 0, \forall i \in \numset{m} \ .
\end{array}
\end{equation}
While there are exponential number of constraints in the above optimisation problems, it is possible to employ the cutting plane method \cite{Tsochantaridis/etal/05} by designing an algorithm that returns the most violated constraint in polynomial time. The most violated constraint w.r.t. a training example $(x,y)$ can be computed by solving the following \emph{loss-augmented inference} (argmax) problems in the slack re-scaling \eq{eq:StructSVM1} and the margin re-scaling \eq{eq:StructSVM2} settings respectively:
\begin{equation}\label{eq:aug_inference1}
\hat{y} = \argmax\limits_{z \in \Ycal}  [1 - \ip{w}{\phi(x,y)-\phi(x,z)}]\Delta(z,y) \ ,
\end{equation}
and
\begin{equation} \label{eq:aug_inference2}
\hat{y} = \argmax\limits_{z \in \Ycal}  [\Delta(z,y) - \ip{w}{\phi(x,y)-\phi(x,z)}] \ .
\end{equation}
\citet{Tsochantaridis/etal/05} showed that a polynomial number of constraints suffices to solve the optimisation problem \eq{eq:StructSVM1} accurately to a desired precision $\epsilon$ assuming that the algorithm that returns the most violated constraint runs in polynomial time. Note that even if the argmax problem is tractable, solving the loss-augmented argmax problem requires further assumptions on the loss function such as it being decomposable over the output variables. An example of such a loss function on the outputs is the Hamming loss.

The optimisation problem \eq{eq:StructSVM2} can be rewritten using a single max constraint for each training example instead of the exponentially many in the following way:
\begin{equation}\label{eq:minmax}
\begin{array}{ll}
\min\limits_{w,\mathbf{\xi}} & \lambda \|w\|^2 + \frac{1}{m} \sum\limits_{i=1}^m \xi_i \\
\mathrm{s.t.:} & \ip{w}{\phi(x_i,y_i)} \ge \max\limits_{z \in \Ycal}  [\Delta(z,y) + \ip{w}{\phi(x,z)}]-\xi_i, \forall i \in \numset{m}\\
& \xi_{i} \ge 0, \forall i \in \numset{m} \ .
\end{array}
\end{equation}
The above formulation, known as min-max formulation, of the optimisation problem \eq{eq:StructSVM2} was proposed by \citet{Taskar/etal/05} who showed that if it is possible to reformulate the loss-augmented inference problem as a convex optimisation problem in a concise way, i.e., with a polynomial number of variables and constraints, then this would result in a joint and concise convex optimisation problem for the original problem \eq{eq:minmax}. In cases where it is not possible to express the inference problem as a concise convex program, \citet{Taskar/etal/05} showed that it suffices to find a concise \emph{certificate of optimality} that guarantees that $y = \argmax_{z \in \Ycal}  [\Delta(z,y) + \ip{w}{\phi(x,z)}]$. Intuitively, verifying that a given output is optimal can be easier than finding one.

Structured SVMs can be \emph{kernelised} by defining a joint kernel function on inputs and outputs $k[(x,y),(x',y')]$ and by considering the Lagrangian dual of the optimisation problem \eq{eq:StructSVM2}:
\begin{equation}\label{eq:NLStructSVM}
\begin{array}{ll}
\min\limits_{\balpha \in \R^{m|\Ycal|}} & \sum\limits_{i,j \in \numset{m},z,z' \in \Ycal} \balpha_{iz} \balpha_{jz'} k[(x_i,z),(x_j,z')] - \sum\limits_{i,z} \Delta(y_i,z)\balpha_{iz} \\
\mathrm{s.t.:} & \sum\limits_{z \in \Ycal} \balpha_{iz} \leq \lambda, ~ \forall i \in \numset{m} \\
& \balpha_{iz} \geq 0, ~ \forall i \in \numset{m}, ~\forall z \in \Ycal \ .
\end{array}
\end{equation}
The non-linear scoring function can be expressed as
\begin{equation*}
f(\cdot,\cdot) = \sum\limits_{i\in\numset{m}, z \in \Ycal} \balpha_{iz} k[(x_i,z),(\cdot,\cdot)] \ .
\end{equation*}
using the representer theorem \cite{Schoelkopf/etal/01}.

\section{Summary}
The main purpose of this chapter was to review classical discriminative machine learning algorithms, including perceptron, support vector machine and logistic regression, and describe how they can be extended to predict structured data. These extensions resulted in several recently proposed structured prediction algorithms such as conditional random fields \cite{Lafferty/etal/01}, max-margin Markov networks \cite{Taskar/etal/03,Taskar/etal/05}, and structured SVMs \cite{Tsochantaridis/etal/05}. In Chapter~\ref{ch:complexity}, we will discuss the assumptions made by these algorithms in order to ensure efficient learning, point to their limitations in the context of predicting combinatorial structures, and propose solutions to circumvent these problems.

\clearemptydoublepage

%%%%%%%%%%%% label ranking survey %%%%%%%%%%%%%%%%%
\chapter{Predicting Permutations}
Binary classification is a well-studied problem in supervised machine learning. Often, in real-world applications such as object recognition, document classification etc., we are faced with problems where there is a need to predict multiple labels. Label ranking is an example of such a complex prediction problem where the goal is to not only predict labels from among a finite set of predefined labels, but also to rank them according to the nature of the input. A motivating application is document categorisation where categories are topics (e.g.: sports, entertainment, politics) within a document collection (e.g.: news articles). It is very likely that a document may belong to multiple topics, and the goal of the learning algorithm is to order (rank) the relevant topics above the irrelevant ones for the document in question. 

Label ranking is also the problem of predicting a specific combinatorial structure --- permutations. It is an interesting problem as it subsumes several supervised learning problems such as multi-class, multi-label, and hierarchical classification \cite{Dekel/etal/03}. This chapter is a survey of label ranking algorithms.
\hide{
\begin{figure}[tb]\label{fig:loss}
\begin{center}
\epsfig{file=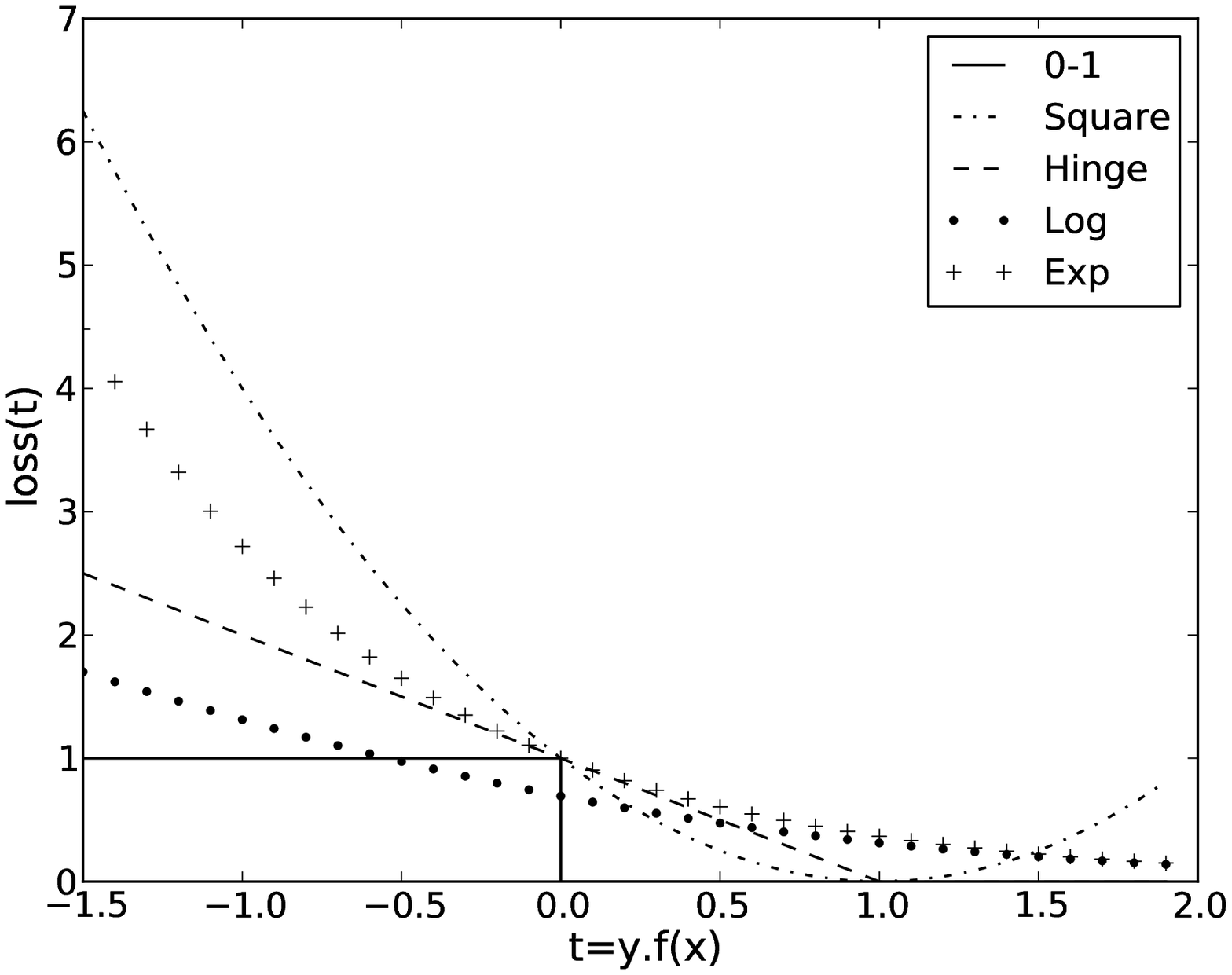,width=.75\textwidth}
\caption{Placeholder for figure explaining how label ranking subsumes several machine learning problems.}
\end{center}
\end{figure}
}
\section{Preliminaries}
We begin with some definitions from order theory, and describe distance metrics and kernels that will be used in this survey.

A binary relation $\succ$ on a (finite) set $\Sigma$ is a \emph{partial order} if $\succ$ is asymmetric \mbox{($a \succ b \dann \neg b \succ a$)} and transitive ($a \succ b \land b \succ c \dann a \succ c$). The pair $(\Sigma, \succ)$ is then called a \emph{partially ordered set} (or \emph{poset}).

We denote the set $\{(u,v) \in \Sigma \mid u \succ v\}$ by $p(\succ)$ and the set of all partial orders over $\Sigma$ by $\mathcal{P}_{\Sigma}$. Note that every partially ordered set $(\Sigma, \succ)$ defines a directed acyclic graph $G_{\succ}=(\Sigma, p(\succ))$. This graph is also called as \emph{preference graph} in the label ranking literature.

A partially ordered set $(\Sigma, \succ)$ such that $\forall u,v \in \Sigma : u \succ v \vee v \succ u$ is a \emph{totally ordered set} and $\succ$ is called a \emph{total order}, a \emph{linear order}, a \emph{strict ranking} (or simply \emph{ranking}), or a \emph{permutation}.\hide{We denote the set of all total orders over $\Sigma$ by $\mathcal{L}_{\Sigma}$.} A \emph{partial ranking} is a total order with ties.

A partial order $\succ'$ \emph{extends} a partial order $\succ$ on the same $\Sigma$ if \mbox{$u\succ v \dann u\succ' v$}. An extension $\succ'$ of a partial order $\succ$ is a \emph{linear extension} if it is totally ordered (i.e.,  a total order $\succ'$ is a \emph{linear extension} of a partial order $\succ$ if \mbox{$\forall u,v \in \Sigma$}, \mbox{$u\succ v \dann u\succ' v$}). A collection of linear orders $\succ_i$ \emph{realises} a partial order $\succ$ if $\forall u,v\in\Sigma, u\succ v \gdw \left(\forall i : u\succ_i v\right)$. We denote this set by $\ell(\succ)$. The \emph{dual} of a partial order $\succ$ is the partial order $\bar \succ$ with $\forall u,v \in \Sigma: u\bar \succ v \gdw v\succ u$.

\subsection*{Distance Metrics}
\emph{Spearman's rank correlation coefficient} ($\rho$) \cite{Spearman/04} is a non-parametric measure of correlation between two variables. For a pair of rankings $\pi$ and $\pi'$ of length $k$, it is defined as
\begin{equation*}
\rho = 1 - \frac{6D(\pi,\pi')}{k(k^2-1)} \ ,
\end{equation*}
where $D(\pi,\pi') = \sum_{i=1}^{k} (\pi(i)-\pi'(i))^2$ is the sum of squared rank distances. The sum of absolute differences $\sum_{i=1}^{k} |\pi(i)-\pi'(i)|$ defines the \emph{Spearman's footrule distance metric}.

\emph{Kendall tau correlation coefficient} ($\tau$) \cite{Kendall/38} is a non-parametric statistic used to measure the degree of correspondence between two rankings. For a pair of rankings $\pi$ and $\pi'$, it is defined as
\begin{equation*}
\tau = \frac{n_c - n_d}{\frac{1}{2}k(k-1)} \ ,
\end{equation*}
where $n_c$ is the number of concordant pairs, and $n_d$ is the number of discordant pairs in $\pi$ and $\pi'$. The number of discordant pairs defines the \emph{Kendall tau distance metric}.

\subsection*{Kernels} \label{ssc:kernels}
We now define kernels on partial orders and describe their properties.\\
 \hide{We now assume a fixed domain $\Sigma$. We have some obvious choises for kernel functions. All could be made more fancy by changing the measure but for simplicity we stick to the cardinality for now. For an application, consider the following example
\begin{example}
Suppose users rent movies and later compare them (instead of rating them). If we assume a consistent rating, we will observe a partial order for each user. To cluster the users (or to classify them according to, say, their favourite genre) we would like to have a kernel on the posets.
\end{example}}
\\ \noindent
\textbf{Position kernel:}
Define 
\begin{equation*}
k_\#:\Pcal\times\Pcal\to \R ~\mathrm{by}~ k_\#(\succ, \succ') = \sum_{u\in\Sigma} \kappa\left(|\{v\in \Sigma \mid v\succ u\}|, |\{v\in \Sigma \mid v\succ' u\}|\right)\ ,
\end{equation*}
where $\kappa$ is a kernel on natural numbers.
\begin{itemize}
\item This function is a kernel and can be computed in time polynomial in $\Sigma$.
%\item For $\kappa(a,b)=ab$ it is related to Spearman's footrule.
\item It is injective, in the sense that $k_\#(\succ, \cdot) = k_\#(\succ', \cdot) \gdw \succ = \succ'$, for linear orders but not for partial orders.
\end{itemize}

\noindent
\textbf{Edge Kernel:}
Define 
\begin{equation*}
k_p:\Pcal\times\Pcal\to \R ~\mathrm{by}~ k_p(\succ, \succ') = |p(\succ) \cap p(\succ')|\ .
\end{equation*}
\begin{itemize}
\item This function is a kernel and can be computed in time polynomial in $|p(\succ)|$. \hide{It would be interesting to see if we can compute it in time polynomial in $|m(\succ)|$.}
%\item It probably works reasonable on linear orders. It might not be all that useful on partial orders (but still worth a try).
\item This kernel is injective in the sense that $k_p(\succ, \cdot) = k_p(\succ', \cdot) \gdw \succ = \succ'$.
\end{itemize}
A downside of this kernel is that $a\succ b$ is as similar to $b\succ a$ as it is to $a\succ c$. However, we can overcome this problem easily. Let $\bar \succ$ be the dual partial order of $\succ$. Define 
\begin{equation*}
k_{\bar p}(\succ, \succ') = k_{p}(\succ, \succ') - k_{p}(\bar \succ, \succ')\ .
\end{equation*}
\begin{itemize}
\item This function is a kernel (the feature space has one feature for every pair of elements and the value of feature $uv$ is $+\sqrt{2}$ iff $u\succ v$, $-\sqrt{2}$ iff $v\succ u$, and $0$ otherwise).
\item It can be computed in time polynomial in $|p(\succ)|$.
\end{itemize}

\noindent
\textbf{Extension Kernel:}
Define 
\begin{equation*}
k_\ell:\Pcal\times\Pcal\to \R ~\mathrm{by}~ k_\ell(\succ, \succ') = |\ell(\succ) \cap \ell(\succ')|\ .
\end{equation*}
\begin{itemize}
\item This function is a kernel. \hide{It probably works reasonable on (sparse) partial orders.}
\item It is injective in the sense that $k_\ell(\succ, \cdot) = k_\ell(\succ', \cdot) \gdw \succ = \succ'$.
\item The kernel cannot be computed in polynomial time as counting linear extensions (or, equivalently, computing $k_\ell(\succ,\succ)$) is $\#$P-complete \cite{Brightwell/Winkler/92}. However, it can possibly be approximated as ($i$) the number of linear extensions can be approximated \cite{Huber06}, and ($ii$) the set of linear extensions can be enumerated almost uniformly.
\item We have $k_\ell(\succ, \succ')=0 \gdw \exists u,v \in \Sigma : u\succ v \wedge v\succ' u$. We call such partial orders \emph{contradicting}. 
\item For non-contradicting partial orders $\succ, \succ'$ define the partial order $\succ \cup \succ'$ such that $\forall u,v \in\Sigma: u (\succ \cup \succ') v \gdw u\succ v \vee u\succ' v$.
%\item Is $\ell(\succ \cup \succ') = \ell(\succ) \cap \ell(\succ')$ for non-contradicting partial orders? it looks obvious but needs a proof. If really so then we can immediatelly use the fully polynomial randomised approxiamtion scheme from counting linear extensions.
\end{itemize}
\hide{
\subsubsection{Chain Kernel}
\begin{itemize}
\item Consider the labelled graph $G^*_\succ$ obtained from a poset $(\Sigma,\succ)$ as $G^*_\succ=( \sigma, p(\succ), \iden(\cdot) )$
\item Apply the walk kernel (can be computed much faster in this case). 
\end{itemize}
}
\subsection*{Label Ranking --- Problem Definition}
Let $\mathcal{X} \subseteq \R^n$ be the input (instance) space, $\Sigma= \{1,\cdots,d\} = \numset{d}$ be a set of labels, and $\mathcal{Y}$ be the output space of all possible partial orders over $\Sigma$. Let $\train = \{(x_i,y_i)\}_{i \in \numset{m}} \in (\mathcal{X} \times \mathcal{Y})^m$ be a set of training examples. Let $G_i=(V_i,E_i)$ denote the preference graph corresponding to $y_i$, for all $i \in \numset{m}$. The goal of a label ranking algorithm is to learn a mapping $f: \mathcal{X} \to \mathcal{Y}$, where $f$ is chosen from a hypothesis class $\mathcal{F}$, such that a predefined loss function $\ell:\mathcal{F} \times \Ycal \times \Ycal \to \R$ is minimised. In general, the mapping $f$ is required to output a total order, but it is also possible to envisage settings where the desired output is a partial order. Let $k_{\Xcal}: \mathcal{X} \times \mathcal{X} \to \R$ and $k_\Ycal: \mathcal{Y} \times \mathcal{Y} \to \R$ denote positive definite kernels on $\mathcal{X}$ and $\mathcal{Y}$, respectively.
\hide{
\subsection{Related Problems}
We now describe some problems that are related to label ranking. A comprehensive survey of literature for these problems is beyond the scope of this chapter. Nevertheless, we refer to, what we believe, are important (and classical), and possibly also recent contributions.
\subsubsection{Multi-Label Classification}
Multi-label classification \cite{Schapire/Singer/98,Schapire/Singer/00,Elisseeff/Weston/01,Fuernkranz/etal/08} was introduced in Chapter \ref{ch:intro}. It is a generalisation of multi-class prediction where the goal is to predict a set of labels that are relevant for a given input. It is a special case of multi-label ranking \cite{Brinker/Huellermeier/07} where the preference graph is bipartite with directed edges between relevant and irrelevant labels.

\subsubsection{Object Ranking}
In this setting, the preference information is given on a subset of the input space and not on the labels. The goal is to learn a ranking function $f:\Xcal \to \R$ such that for any $a,b \in \Xcal$, $f(a) > f(b)$ iff $a \succ b$. Thus, a total order is induced on the input space. This setting has attracted a lot of attention recently, in particular, in information retrieval applications. Various approaches to solve this problem have been proposed in the literature \cite{Cohen/Schapire/Singer/99,Herbrich/et/al/00,Crammer/Singer/01,Joachims/02,Freund/et/al/03,Burges/et/al/05,Fung/etal/05,Burges/etal/06,Rudin/06}. Learning object ranking functions on structured inputs (graphs) was proposed recently in \cite{Agarwal/06,Agarwal/Chakrabarti/07,Vembu/etal/07}. A survey on object ranking algorithms also appears as a chapter in this book.

\subsubsection{Ordinal Regression}
Ordinal regression \cite{Herbrich/et/al/99,Herbrich/et/al/00,Chu/Keerthi/05,Chu/Ghahramani/05} is a form of multi-class prediction where the labels are defined on an ordinal scale and therefore cannot be treated independently. It is closely related to the object ranking problem where the preference information on (a subset of) the input space is a directed $k$-partite graph where $k$ is the number of ordinal values (labels).
}

\section{Learning Reductions}
Learning reductions are an efficient way to solve complex prediction problems using simple models like (binary) classifiers as primitives. Such techniques have been applied to solve problems like ranking \cite{Balcan/etal/08}, regression \cite{Langford/Zadrozny/05}, and structured prediction \cite{Daume06}, just to name a few.

Label ranking can be reduced to binary classification using Kesler's construction \cite{Nilsson65}. This approach was proposed by \citet{HarPeled/etal/02,HarPeled/etal/02b} under the name of constraint classification. The idea is to construct an expanded example sequence $\train'$ in which every example $(x,y) \in \R^n \times \Ycal$ with its corresponding preference graph $G=(V,E)$ is embedded in $\R^{dn} \times \{-1,1\}$, with each preference $(p,q) \in E$ contributing a single positive and a single negative example. The Kesler mapping $P$ is defined as follows:
\begin{equation*}
\begin{array}{l}
P_+(x,y) = \{(x \otimes \zero_p,1) \mid  (p,q) \in E\} \subseteq \R^{dn} \times \{1\} \\
P_-(x,y) = \{(-x \otimes \zero_q,-1) \mid  (p,q) \in E\} \subseteq \R^{dn} \times \{-1\} \ ,
\end{array}
\end{equation*}
where $\zero_j$ is a $d$-dimensional vector whose $j$th component is one and the rest are zeros. Let $P(x,y) = P_+(x,y) \cup P_-(x,y)$. The expanded set of examples is then given by
\begin{equation*}
\train'= P(\train) = \bigcup_{(x,y) \in \train} P(x,y) \subseteq \R^{dn} \times \{-1,1\} \ .
\end{equation*}
A binary classifier (linear separating hyperplane) trained on this expanded sequence can be viewed as a sorting function over $d$ linear functions, each in $\R^n$. The sorting function is given as $\argsort_{j \in \numset{d}} \ip{w_j}{x}$, where $w_j$ is the $j$-th chunk of the weight vector $w \in \R^{dn}$, i.e., $w_j = (w_{(j-1)n+1}, \cdots, w_{jn})$.

A reduction technique proposed by \citet{Fuernkranz/02} known as pairwise classification can be used to reduce the problem of multi-class prediction to learning binary classifiers. An extension of this technique known as ranking by pairwise comparison (RPC) was proposed in \cite{Fuernkranz/Huellermeier/03,Huellermeier/etal/08} to solve the label ranking problem. The central idea is to learn a binary classifier for each pair of labels in $\Sigma$ resulting in $d(d-1)/2$ models. Every individual model $\model_{pq}$ with $p,q \in \Sigma$ learns a mapping that outputs 1 if $p \succ_x q$ and 0 if $q \succ_x p$ for an example $x \in \Xcal$. Alternatively, one may also learn a model that maps into the unit interval $[0,1]$ instead of $\{0,1\}$. The resulting model assigns a valued preference relation $R_x$ to every example $x \in \Xcal$:
\begin{align*}
R_x(p, q) = & \begin{cases}
\model_{pq}(x) & \text{if}\ p < q \\
1-\model_{pq}(x) & \text{if}\ p > q \\
\end{cases}
\end{align*}
The final ranking is obtained by using a ranking procedure that basically tries to combine the results of these individual models to induce a total order on the set of labels. A simple ranking procedure is to assign a score $s_x(p) = \sum_{p \neq q} R_x(p,q)$ to each label $p$ and obtain a final ordering by sorting these scores. This strategy exhibits desirable properties like transitivity of pairwise preferences. Furthermore, the RPC algorithm minimises the sum of squared rank distances and an approximation of the Kendall tau distance metric under the condition that the binary models $\model_{pq}$ provide correct probability estimates, i.e., \mbox{$R_x(p,q) = \model_{pq}(x)=\Pr[p \succ_x q]$}.

\section{Boosting Methods} \label{sc:boosting}
A boosting \cite{Freund/Schapire/97} algorithm for label ranking was proposed by \citet{Dekel/etal/03}. A label ranking function $f: \Xcal \times \Sigma \to \R$ is learned such that for any given $x \in \Xcal$, a total order is induced on the label set by $p \succ_x q \iff f(x, p)>f(x,q)$.  The label ranking function is represented as a linear combination of a set of $L$ base ranking functions, i.e, $f(x,p)=\sum_{l=1}^L \lambda_l h_l(x,p)$, where $\{\lambda_l\}_{l \in \numset{L}}$ are parameters that are estimated by the boosting algorithm. We denote the label ranking induced by $f$ for $x$ by $f(x)$ (with a slight abuse of notation). A graph decomposition procedure $\mathcal{D}$, which takes a preference graph $G_i=(V_i,E_i)$ for any $x_i \in \Xcal$ as its input and outputs a set of $S_i$ subgraphs $\{G_{i,s}\}_{s \in \numset{S_i}}$, has to be specified as an input to the learning algorithm. A simple example of a graph decomposition procedure is to consider every edge $e \in E_i$ as a subgraph. Other examples include decomposing the graph into bipartite directed graph $G_{i,s}=(U_{i,s},V_{i,s},E_{i,s})$ such that \mbox{$|U_{i,s}|=1$} or $|V_{i,s}|=1$ (see Figure 2 in \citet{Dekel/etal/03} for an illustration). The generalised loss due to $f(x_i)$ w.r.t. $G_i$ is the fraction of subgraphs in $\mathcal{D}(G_i)$ with which $f(x_i)$ disagrees. The generalised loss over all the training instances is defined as
\begin{equation*}
\ell_{\text{gen}}(f,\train,\mathcal{D}) = \sum\limits_{i=1}^m \frac{1}{S_i} \sum\limits_{s=1}^{S_i} \delta(f(x_i),G_{i,s}) \ ,
\end{equation*}
where $\delta(\cdot,\cdot)$ is a loss function defined on the subgraphs such as the 0-1 loss or the ranking loss \cite{Schapire/Singer/00}. While minimising such a discrete, non-convex loss function is NP-hard, it is possible to minimise an upper bound given by
\begin{equation*}
\delta(f(x_i),G_{i,s}) \leq  \log_2 (1 + \sum\limits_{e \in E_{i,s}} \text{exp}( f(x_i,\text{term}(e)) - f(x_i,\text{init}(e)) )) \ .
\end{equation*}
where $\text{init}(e)$ (resp. $\text{term}(e)$) is the label corresponding to the initial (resp. terminal) vertex of any directed edge $e \in E_{i,s}$. To minimise this upper bound, \citet{Dekel/etal/03} proposed to use a boosting-style algorithm for exponential models \cite{Lebanon/Lafferty/01,Collins/etal/02} to estimate the model parameters $\lambda$ and also proved a bound on the decrease in loss in every iteration of the algorithm.

\section{Label Ranking SVM}
\citet{Elisseeff/Weston/01} proposed a kernel method for multi-label classification. A straightforward generalisation of this approach results in a label ranking algorithm. Define a scoring function for label $p$ and input $x$ as \mbox{$h_p(x)=\langle w_p, x\rangle$}, where $w_p$ is a weight vector corresponding to label $p$. These scoring functions together will define the mapping $f$ by a sorting operation, i.e., $f(x) = \argsort_{j \in \numset{d}} \ip{w_j}{x}$. The ranking loss \cite{Schapire/Singer/00} w.r.t. to a preference graph $G=(V,E)$ is defined as $\ell(f,x,y) = \frac{1}{|E|}|(p,q) \in E ~\mathrm{s.t.}~ h_p(x) \le h_q(x)|$. The following optimisation problem minimises the ranking loss:
\begin{equation*}\label{eq:SVMLRank1}
\begin{array}{ll}
\min\limits_{\{w_j\}_{j \in \numset{d}}} & \sum\limits_{j=1}^d \|w_j\|^2 + \lambda \sum\limits_{i=1}^m \frac{1}{|E_i|}\sum\limits_{(p,q) \in E_i} \xi_{ipq} \\
\mathrm{subject~to:} & \langle w_p - w_q, x_i \rangle \ge 1 - \xi_{ipq}, \forall (p,q) \in E_i, \forall i \in \numset{m}\\
& \xi_{ipq} \ge 0, \forall (p,q) \in E_i, \forall i \in \numset{m} \ ,
\end{array}
\end{equation*}
where $\lambda > 0$ is the regularisation parameter that trades-off the balance of the loss term against the regulariser.
\hide{
\begin{description}
\item{\textbf{Primal:}}
\begin{equation}\label{eq:SVMLRank1}
\begin{array}{ll}
\min\limits_{w_j,j=\numset{l}} & \sum\limits_{j=1}^l \|w_j\|^2 + C\sum\limits_{i=1}^m \frac{1}{|E_i|}\sum\limits_{(p,q) \in E_i} \xi_{ikl} \\
\mathrm{subject~to:} & \langle w_p - w_q, x_i \rangle \ge 1 - \xi_{ikl}, \forall (p,q) \in E_i, \forall i \\
& \xi_{ipq} \ge 0
\end{array}
\end{equation}
\item{\textbf{Dual:}} The dual formulation for multi-label classification is given in the technical report of \cite{Elisseeff/Weston/01} including an efficient technique to solve the dual optimisation problem.
\end{description}
The algorithm induces a total order on the labels using the individual scoring functions $\{r_i(x)\}_{i=1}^l$. Let us try to reformulate the optimisation problem using a different approach. Let $\phi(x,y)$ be a joint feature map on the input-output space. We assume that a label $y \in \mathcal{Y}$ can be decomposed into parts $r$ and that the joint feature map on these parts $\phi(x,r), \forall r \in y$ is well-defined. Consider a simple decomposition of a label $y$ into its individual edges (parts) $y_{pq}, \forall (p,q) \in E$ of the corresponding DAG. An alternative optimisation problem minimising the ranking loss could be formulated as follows:
\begin{description}
\item{\textbf{Primal:}}
\begin{equation}\label{eq:SVMLRank2}
\begin{array}{ll}
\min\limits_w &  \|w\|^2 + C\sum\limits_{i=1}^m \frac{1}{|E_i|}\sum\limits_{(p,q) \in E_i} \xi_{ipq} \\
\mathrm{subject~to:} & \langle w, \phi(x_i,y_{pq}) \rangle  - \langle w, \phi(x_i,y_{qp}) \rangle \ge 1 - \xi_{ipq}, \forall (p,q) \in E_i, \forall i \\
& \xi_{ipq} \ge 0
\end{array}
\end{equation}
\item{\textbf{Dual:}} todo
\end{description}

For any given instance $x$, label $p$ is ranked higher than label $q$ if $\langle w, \phi(x, y_{pq}) \rangle > \langle w, \phi(x, y_{qp}) \rangle$. A total order on the labels is thus obtained.

A shortcoming of this approaches is that it is not possible to train the algorithms using arbitrary loss functions or similarity measures (kernels) on the output space $\mathcal{Y}$. The algorithms also assume that the preferences (directed edges) are independent of each other thereby discarding any structural information of the label. The structural information could be nicely captured by defining an appropriate kernel on the label space. We then would like to design a label ranking algorithm that is able to directly operate on these kernels. Later sections will deal with algorithms of this kind.}

\citet{Shwartz/etal/06} considered the setting where the training labels take the form of a feedback vector $\gamma \in \R^d$. The interpretation is that label $p$ is ranked higher than label $q$ iff $\gamma_p > \gamma_q$. The difference $\gamma_p - \gamma_q$ encodes the importance of label $p$ over label $q$ and this information is also used in the optimisation problem. The loss function considered in this work is a generalisation of the hinge-loss for label ranking. For a pair of labels $(p,q) \in \Sigma$, the loss with respect to $f$ is defined as
\begin{equation*}
\ell_{p,q} (f(x),\gamma) = [(\gamma_p-\gamma_q) - (h_p(x)-h_q(x))]_+ \ ,
\end{equation*}
where $[a]_+ = \max(a,0)$. At the heart of the algorithm lies a decomposition framework, similar to the one mentioned in the previous section, that decomposes any given feedback vector into complete bipartite subgraphs, and losses are defined and aggregated over these subgraphs. This decomposition framework makes the approach very general, albeit at the cost of solving a complex optimisation problem. Interestingly, the quadratic programming formulation for multi-label classification as proposed by Elisseeff and Weston \cite{Elisseeff/Weston/01} can be recovered as a special case of this approach.

\hide{
Before proceeding to the next section, we would like to refer to the label ranking algorithm described in \cite{Shwartz/etal/06}. The authors formulate the task of ranking labels as a QP optimisation problem. In their approach, the training label takes the form of a \emph{feedback} vector $\gamma \in \R^n$ and the interpretation is that label $p$ is ranked higher than label $q$ iff $\gamma_p > \gamma_q$. The difference $\gamma_p - \gamma_q$ is interpreted as encoding the importance of label $p$ over label $q$ and this information is also used in the optimisation problem by defining a loss function which generalises the hinge-loss used in binary classification. The algorithm learns a mapping of the form $f: \mathcal{X} \to \R^n$ and outputs a total order on the label set $Y$. At the heart of the algorithm lies a decomposition framework that decomposes any given feedback vector into complete bipartite subgraphs, and losses are defined and aggregated over these subgraphs. This decomposition framework makes their approach very general, albeit at the cost of solving a complex optimisation problem, and indeed the QP formulation for multi-label classification as proposed in \cite{Elisseeff/Weston/01} can be recovered as a special case. What is unclear though is the construction of the feedback vector $\gamma$ given an arbitrary partial order or a weighted directed acyclic graph.
}

\section{Structured Prediction}\label{sc:lr_sp}
The motivation behind using a structured prediction framework to solve the label ranking problem stems from the added flexibility to use arbitrary loss functions and kernels, in principle, on the output space. In this section, we let $\Ycal$ to be the space of all total orders of the label set $\Sigma$.

Recall the optimisation problem of structured SVMs \cite{Tsochantaridis/etal/05} (cf. Chapter \ref{ch:prelims}):
\begin{equation}\label{eq:StructSVMLRank}
\begin{array}{ll}
\min\limits_{w,\mathbf{\xi}} & \lambda \|w\|^2 + \frac{1}{m} \sum\limits_{i=1}^m \xi_i \\
\mathrm{s.t.:} & \langle w,\phi(x_i,y_i) \rangle  - \langle w,\phi(x_i,z) \rangle \ge \Delta(y_i,z) - \xi_{i}, \forall z \in \Ycal \setminus y_i, \forall i \in \numset{m}\\
& \xi_{i} \ge 0, \forall i \in \numset{m} \ .
\end{array}
\end{equation}
Here, $\Delta:\Ycal \times \Ycal \to \R$ is a loss function on total orders. To handle the exponential number of constraints in the above optimisation problem, we need to design a separation oracle \cite{Tsochantaridis/etal/05} that returns the most violated constraint in polynomial time. The most violated constraint with respect to a training example $(x,y)$ can be computed using the following optimisation problem:
\begin{equation*}
\hat{y} = \argmax\limits_{z \in \Ycal} f(x,y)  + \Delta(z,y) \ .
\end{equation*}
The above loss-augmented inference problem and the decoding problem can be solved using techniques described by \citet{Le/Smola/07}. Here, the scoring function $f$ takes a slightly different form. Let $g(x,p;w_p) = \ip{\phi(x)}{w_p}$ ($\phi$ is feature map of inputs) denote the scoring function for an individual label $p \in \Sigma$ parameterised by weight vector $w_p$. Now define the scoring function $f$ for the pair $(x,y)$ as follows:
\begin{equation*}
f(x,y;\textbf{w}) = \sum\limits_{j=1}^d g(x,j)c(y)_j = \sum\limits_{j=1}^d \langle \phi(x), w_j \rangle c(y)_j \ ,
\end{equation*}
parameterised by the set $\mathbf{w} = \{w_j\}_{j \in \numset{d}}$ of weight vectors, where $c$ is a decreasing sequence of reals and $c(y)$ denotes the permutation of $c$ according to $y$, i.e., $c(y)_j = c_{y(j)}$ for all $j \in \numset{d}$. \hide{Note that the joint scoring function on input-output pairs can be written as $\Phi(x,y) = \sum_{j \in \numset{k}} \langle \phi(x), w_j \rangle c(y)_j$.}The final prediction $\hat{y} = \argmax_{y \in \Ycal} f(x,y)$ is obtained by sorting the scores $g(x,p)$ of the individual labels. This is possible due to the Polya-Littlewood-Hardy inequality \cite{Le/Smola/07}. The decoding problem is thus solved. We now turn our attention to designing a separation oracle. The goal is to find
\begin{equation} \label{eq:LinAs}
\begin{array}{ll}
\hat{y} & = \argmax\limits_{z \in \Ycal} f(x,y)  + \Delta(y,z) \\ 
& = \argmax\limits_{z \in \Ycal}  \sum\limits_{j=1}^d \langle \phi(x), w_j \rangle c(y)_j + \Delta(y, z) \ .
\end{array}
\end{equation}
For certain loss functions that are relevant in information retrieval applications, \citet{Le/Smola/07} showed that the above optimisation problem can be formulated as a linear assignment problem and can be solved using the Hungarian marriage method (Kuhn-Mungres algorithm) in $O(d^3)$ time. For arbitrary loss functions, it may not be feasible to solve the optimisation problem \eq{eq:LinAs} efficiently. Note that the term $\Delta(\cdot,\cdot)$ in the separation oracle and also in the constraint set of structured SVM specifies an output dependent margin. Replacing it with a fixed margin $\gamma ~(=1)$ would greatly simplify the design of separation oracle since it reduces to a sorting operation as in the decoding problem.\hide{The optimisation problem \eq{eq:StructSVMLRank} can be kernelised in its dual form. Let $\{\alpha_{iz}\}_{i \in \numset{n}, z \in \Ycal}$ denote the set of dual parameters. Let the joint scoring function on input-output pairs be an element of $\Hcal$ with $\Hcal=\Hcal_\Xcal\otimes\Hcal_\Ycal$ where $\Hcal_\Xcal, \Hcal_\Ycal$ are the RKHS of $k_\Xcal, k_\Ycal$ respectively and $\otimes$ denotes the tensor product. Note that the reproducing kernel of $\Hcal$ is then $k[(x,y),(x',y')]=k_\Xcal(x,x')k_\Ycal(y,y')$. The dual optimisation problem is then given as
\begin{equation*}
\begin{array}{ll}
\min\limits_{\alpha} & \sum\limits_{i,j \in \numset{n},z,z' \in \Ycal} \alpha_{iz} \alpha_{jz'} k_{\Xcal}(x_i,z)k_{\Ycal}(z,z') - \sum\limits_{i,z}\alpha_{iz} \\
\mathrm{subject~to:} & \sum\limits_{z \in \Ycal} \alpha_{iz} \leq \nu, ~ \forall i \in \numset{n} \\
& \alpha_{iz} \geq 0, ~ \forall i \in \numset{n}, ~\forall z \in \Ycal \ .
\end{array}
\end{equation*}
}
Since the optimisation problem \eq{eq:StructSVMLRank} can be kernelised in its dual form (cf. problem \eq{eq:NLStructSVM}), it allows us to use arbitrary kernel functions on the output space such as those described in Section \ref{ssc:kernels}.

\hide{
Let us consider a loss function that minimises the disagreement between permutations, i.e., the ranking loss defined in an earlier section. Let $L \in \R^{k \times k}$ be a matrix with elements $L_{ij} = 1 - \delta(\pi_i, j)$. Then, maximising (\ref{eq:LinAs}) amounts to solving the following linear assignment problem
\begin{equation*}
\argmax\limits_{\pi \in \Pi} \sum\limits_{i=1}^k C_{i,\pi_i} \mathrm{~where~} C_{ij} = c_jg_i - L_{ij} \ ,
\end{equation*}
and can be done using the Hungarian marriage method (Kuhn-Mungres algorithm) in $O(k^3)$ time.}

\section{Online Methods}
Online classification and regression algorithms like perceptron {\cite{Rosenblatt58} typically learn a linear model $f(x)=\ip{w}{x}$ parameterised by a weight vector $w \in \R^n$. The algorithms operate in rounds (iterations). In round $t$, \emph{nature} provides an instance to the learner; the learner makes a prediction using the current weight vector $w^t$; nature reveals the true label $y^t$ of $x^t$; learner incurs a loss $\ell(\ip{w^t}{x^t},y^t)$ and updates its weight vector accordingly. Central to any online algorithm is the update rule that is designed in such a way so as to minimise the cumulative loss over all the iterations. In label ranking scenarios, online algorithms \cite{Crammer/Singer/03a,Crammer/Singer/05,Shwartz/Singer/07} maintain a set of weight vectors $\{w_j\}_{j \in \numset{d}}$, one for every label in $\Sigma$, and the update rule is applied to each of these vectors.

Online algorithms for label ranking have been analysed using two different frameworks: passive-aggressive \cite{Crammer/etal/06} and  primal-dual \cite{Shalev-Shwartz/Singer/07}. Passive-aggressive algorithms for label ranking \cite{Crammer/Singer/05} are based on Bregman divergences and result in multiplicative and additive update rules \cite{Kivinen/Warmuth/97}. A Bregman divergence \cite{Bregman/97} is similar to a distance metric, but does not satisfy the triangle inequality and the symmetry properties. In every iteration $t$, the algorithm updates its weights in such a way that it stays close to the previous iteration's weight vector w.r.t. the Bregman divergence, and also minimises the loss on the current input-output $(x^t,y^t)$ pair. Let $W \in \R^{d \times n}$ denote the set of weight vectors in matrix form.  The following optimisation problem is considered:
\begin{equation*}
W^{t} = \argmin_W B_F(W\|W^{t-1}) + \lambda \ell(f(x^t;W),y^t) \ ,
\end{equation*}
where $B_F$ is the Bregman divergence defined via a strictly convex function $F$. The choice of the Bregman divergence and the loss function result in different update rules. Additive and multiplicative update rules can be derived respectively by considering the following optimisation problems \cite{Crammer/Singer/05}:
\begin{equation*}
W^{t} = \argmin_W \|W-W^{t-1}\|^2 + \lambda \ell(f(x^t;W),y^t) \ ,
\end{equation*}
and 
\begin{equation*}
W^{t} = \argmin_W D_{\text{KL}}(W\|W^{t-1}) + \lambda \ell(f(x^t;W^t),y^t) \ ,
\end{equation*}
where $D_{\text{KL}}$ is the Kullback-Liebler divergence. The loss functions considered by \citet{Crammer/Singer/05} is similar to the ones defined by \citet{Dekel/etal/03} (see also Section \ref{sc:boosting}), where a preference graph is decomposed into subgraphs using a graph decomposition procedure, and a loss function such as the 0-1 loss or the ranking loss is defined on every subgraph. The loss incurred by a ranker $W$ for a graph decomposition procedure $\mathcal{D}(G)$ is given as
\begin{equation*}
\ell(f(x;W),y) = \sum\limits_{g \in \mathcal{D}(G)} |\{(r,s) \in g : \ip{w_r}{x} \leq \ip{w_s}{x}\} \neq \emptyset | \ .
\end{equation*}

The primal-dual framework \cite{Shalev-Shwartz/Singer/07} was used by \citet{Shwartz/Singer/07} resuting in a unifying algorithmic approach for online label ranking. The loss function considered in this work is a generalisation of the hinge-loss for label ranking. The training labels are assumed to be a set of relevant and irrelevant labels (as in multi-label classification). For a given instance $x \in \Xcal$, let \mbox{$\Sigma_r \subseteq \Sigma$} denote the set of relevant labels. The hinge-loss for label ranking w.r.t. an example $(x^t,\Sigma_r^t)$ at iteration $t$ is defined as:
\begin{equation*}
\ell^{\gamma}(W^t;(x^t,y^t)) = \max\limits_{r \in \Sigma_r^t, s \notin \Sigma_r^t} [\gamma - (\ip{w_r^t}{x^t} - \ip{w_s^t}{x^t})]_+ \ .
\end{equation*}
The central idea behind the analysis is to cast online learning as an optimisation (minimisation) problem consisting of two terms: the complexity of the ranking function and the empirical label-ranking loss. The notion of duality in optimisation theory \cite{Boyd/Thomas/04} is used to obtain lower bounds on the optimisation problem, which in turn yields upper bounds on the number of prediction mistakes made by the algorithm. The reader is referred to \cite{Shwartz/Singer/07} that presents several update rules for label ranking, and these are also shown to generalise other update rules such as the ones defined by \citet{Crammer/Singer/03a}.

\section{Instance-based Learning}
In instance-based learning, the idea is to predict a label for a given instance based on local information, i.e., labels of neighbouring examples. In label ranking, these labels are rankings (partial orders, total orders, partial rankings) and one has to use aggregation algorithms \cite{Dwork/etal/01,Fagin/etal/04,Ailon/etal/05,Ailon07,Zuylen/Williamson/07} to combine rankings from neighbouring examples. Instance-based learning algoritms for label ranking were proposed recently by \citet{Brinker/Huellermeier/06,Brinker/Huellermeier/07,Cheng/Huellermeier/08,Cheng/etal/09}. Let $\{y_i\}_{i \in \numset{B}}$ denote a set of $B$ neighbouring rankings for any given instance $x \in \Xcal$. The goal is to compute a ranking $\hat{y}$ that is optimal w.r.t. a loss function $\ell:\Ycal \times \Ycal \to \R$ defined on pairs of rankings. More formally, the following optimisation problem needs to be solved:
\begin{equation}\label{eqn:rankaggr}
\hat{y} = \argmin\limits_{y \in \Ycal} \sum\limits_{i=1}^B \ell(y,y_i) \ .
\end{equation}
This is a very general statement of the problem. Various aggregation algorithms, which we survey in the sequel, can be used to solve this optimisation problem depending on the nature of the loss function and also on the inputs (of the optimisation problem).

\subsubsection{Aggregating Total Orders}
The problem of finding an optimal ranking when the inputs in the optimisation problem \eq{eqn:rankaggr} are total orders can be fomulated as a feedback arc set problem in digraphs (specifically in tournaments) \cite{Ailon/etal/05}. A tournament is a directed graph $G=(V,E)$ such that for each pair of vertices $p,q \in V$, either $(p,q) \in E$ or $(q,p) \in E$. The minimum feedback arc set (FAS) is the smallest set $E' \subseteq E$ such that $(V, E-E')$ is acyclic. The rank aggregation problem can be seen as special case of weighted FAS-tournaments; the weight $w_{pq}$ of an edge $(p,q)$ is the fraction of rankings that rank $p$ before $q$.

Optimising the Spearman footrule distance metric in the minimisation problem \eq{eqn:rankaggr} is equivalent to finding the minimum cost maximum matching in a bipartite graph with $d$ nodes \cite{Dwork/etal/01}. A $2$-factor approximation algorithm with time complexity $O(Bd+d\log d)$ was proposed by \citet{Fagin/etal/04}. Optimising the Kendall tau distance metric in \eq{eqn:rankaggr} is NP-hard \cite{Bartholdi/etal/89} and therefore one has to use approximation algorithms \cite{Ailon/etal/05,Zuylen/Williamson/07} to output a Kemeny optimal ranking. There exists a deterministic, combinatorial $8/5$-approximation algorithm for aggregating total orders \cite{Zuylen/Williamson/07}. The approximation ratio can be improved to $11/7$ by using a randomised algorithm \cite{Ailon/etal/05} and to $4/3$ by using a derterministic linear programming based algorithm \cite{Zuylen/Williamson/07}. A polynomial time approximation scheme was proposed by \citet{Mathieu/Schudy/07}. 

\hide{All of these algorithms assume that the weights satisfy the probability constraints ($w_{ij} + w_{ji} = 1$ for all pairs $i,j \in V$) and triangle inequality constraints ($w_{ij} + w_{jk} \geq w_{ik}$ for all triples $i,j,k \in V$). The reader is referred to \cite{Zuylen/Williamson/07} for an overview of approximation algorithms in cases where one of these assumptions fail.}

\subsubsection{Aggregating Partial Rankings}
A typical application of this setting is multi-label ranking \cite{Brinker/Huellermeier/07} where the preference graph is bipartite with directed edges between relevant and irrelevant labels. There exists a deterministic, combinatorial $8/5$-approximation algorithm for aggregating partial rankings \cite{Zuylen/Williamson/07}. The running time of this algorithm is $O(d^3)$. A slightly better approximation guarantee of $3/2$ can be obtained by using a deterministic, linear programming based algorithm \cite{Zuylen/Williamson/07}. These algorithms minimise the Kemeny distance between the desired output and the individual partial rankings. An exact method for aggregating partial rankings using (generalised) sum of squared rank distance metric was proposed by \citet{Brinker/Huellermeier/07}.

\subsubsection{Aggregating Partial Orders}
In this setting, we allow inupt labels to be partial orders and the desired output is a total order. To the best of our knowledge, there are no approximation algorithms to aggregate partial orders, but it is possible to reduce the problem to that of aggregating total orders as follows: given a partial order, sample a set (of some fixed cardinality) of linear extensions \cite{Huber06} and use existing approximation algorithms for aggregating total orders. If the desired output is a partial order and not a total order, one can  consider the following optimisation problem:
\begin{equation*}
\hat{y} = \argmax\limits_{z \in \mathcal{Y}} \sum\limits_{i=1}^m k_\Xcal(x_i, x)k_\Ycal(y_i, z) \ .
\end{equation*}
Under the assumption that $k_\Xcal(\cdot,\cdot) \geq 0$ and $k_\Ycal(\cdot,\cdot) \geq 0$, and if the edge kernel (cf. Section~\ref{ssc:kernels}) on partial orders is used, the above optimisation problem can be approximately solved using the maximum acyclic subgraph algorithm \cite{Hassin/Rubinstein/94,McDonald/etal/05}.

\section{Summary}
Label ranking is a specific example of the learning problem of predicting combinatorial structures. The problem has attracted a lot of interest in recent years as evidenced by the increasing number of algorithms attempting to solve it. The main purpose of this chapter was to give an overview of existing literature on label ranking algorithms. While most of these are specialised algorithms, we have seen in Section~\ref{sc:lr_sp} that the problem can also be solved within the structured prediction framework using structured SVMs. We will revisit the problem of predicting permutations --- as an example --- in the next chapter.

\clearemptydoublepage
\chapter{Complexity of Learning} \label{ch:complexity}
In Chapter~\ref{ch:prelims}, we discussed several discriminative structured prediction algorithms. We will now revisit some of these algorithms, try to get a deeper understanding of the assumptions they make to ensure efficient learning, and identify their shortcomings in the context of predicting combinatorial structures. We will then introduce two new assumptions and show that they hold for several combinatorial structures. These assumptions will be used in the design and analysis of structured prediction algorithms in subsequent chapters.

\section{Efficient Learning}\label{sc:eff_learn}
Recall that the structured loss with respect to a hypothesis $h$ and a training example $(x,y)$ is defined as $\ell^\Delta_{\max}(h,(x,y)) = \Delta (\argmax_{z \in \Ycal} h(x,z), y)$, where $\Delta:\Ycal\times\Ycal\to\R$ is a discrete, non-convex loss function defined on the output space. To ensure convexity, it is typical to upper-bound this loss by the structured hinge loss $\ell^\Delta_\mathrm{hinge} (h,(x,y)) = \argmax_{z\in\Ycal} \left[ \Delta (z, y) + h(x,z) - h(x,y)\right]$. Regularised risk minimisation based approaches \cite{Tsochantaridis/etal/05,Taskar/etal/03,Taskar/etal/05} aim at solving the optimisation problem $Q(\{(x_i,y_i) \mid i \in \numset{m}\}) = $
\begin{equation}\label{eq:cvx}
\begin{array}{rl}
\argmin\limits_{h\in\Hcal} & \lambda \Omega[h] + \sum\limits_{i\in\numset{m}} \xi_{i}\medskip\\
\text{subject to} &  h(x_i,y_i)-h(x_i,z) \geq \Delta (z, y_i) - \xi_{i}, \quad\forall i\in\numset{m}, \forall z\in\Ycal \setminus y_i \smallskip\\
& \xi_{i}\geq0, \quad \forall i\in\numset{m}\ ,
\end{array}
\end{equation}
where $\lambda > 0$ is a regularisation parameter, $\Hcal$ is a reproduding kernel Hilbert space with a corresponding kernel $k[(x,y),(x',y')]$, and $\Omega: \Hcal \to \R$ is a convex regularisation function such as the squared $\ell_2$ norm, $\|\cdot\|^2$.

The major issue in solving this optimisation problem is that the number of constraints grows proportional to $|\Ycal|$. If the set $\Ycal$ is parameterised by a finite alphabet $\Sigma$, then the number of constraints is usually exponential in $|\Sigma|$. To ensure polynomial time complexity different assumptions need to be made, and depending on the nature of $\Omega$ different methods are used that iteratively optimise and add violated constraints. We now describe these assumptions in decreasing order of strength.

\subsubsection{\textsc{Decoding}} The strongest assumption is the existence of a polynomial time algorithm for exact decoding \cite{Tsochantaridis/etal/05}. \emph{Decoding} refers to the problem of computing $\argmax_{z\in\Ycal} h(x,z)$ for a given $(h,x)\in\Hcal\times\Xcal$.

\subsubsection{\textsc{Separation}} A weaker assumption made by structured perceptron \cite{Collins02} is the existence of a polynomial time algorithm for separation\footnote{Algorithms for decoding and separation are also referred to as inference algorithms in the literature.}. \emph{Separation} refers to the problem of finding for a given scoring function $h$, $x\in\Xcal$ and $y\in\Ycal$, any $z\in\Ycal$ such that $h(x,z)>h(x,y)$ if one exists or prove that none exists otherwise. A polynomial time algorithm for exact decoding implies a polynomial time algorithm for separation.

\subsubsection{\textsc{Optimality}} An even weaker assumption is that optimality is in NP \cite{Taskar/etal/05}. \emph{Optimality} refers to the problem of deciding if for a given scoring function $h$, $x\in\Xcal$ and $y\in\Ycal$, it holds that $y \in \argmax_{z\in\Ycal} h(x,z)$. A polynomial time algorithm for separation implies a polynomial time algorithm for optimality.

For several combinatorial structures considered in this work, there exists a short certificate of non-optimality (i.e., non-optimality is in NP), but there is no short certificate of optimality unless coNP=NP (complexity classes are illustrated in Figure~\ref{fig:comp_class}).
\begin{figure}[tb]
\begin{center}
\epsfig{file=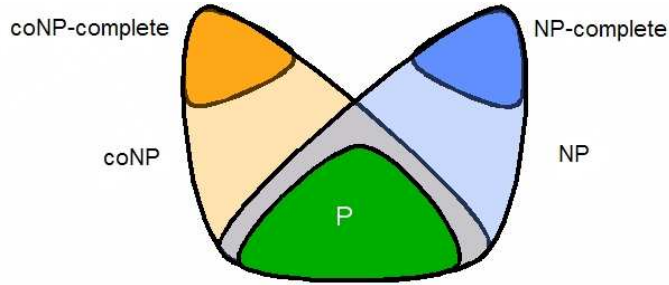,width=.75\textwidth}
\end{center}
\caption{Complexity class diagram.}
\label{fig:comp_class}
\end{figure}
This implies that polynomial time algorithms for exact decoding and separation do not exist. In other words, none of the existing structured prediction algorithms \cite{Collins02,Taskar/etal/03,Tsochantaridis/etal/05,Taskar/etal/05} can be trained efficiently to predict the combinatorial structures that are of interest to us.

Recently, there have been some attempts to use approximate inference algorithms for learning structured prediction models. \citet{Kulesza/Pereira/07} performed a theoretical analysis of the relationship between approximate inference and efficient learning. They showed that learning can fail even when there exists an approximate inference algorithm with strong approximation guarantees and argued that, to ensure efficient learning under approximate inferece, it is crucial to choose compatible inference and learning algorithms. As an example, they showed that a linear programming based approximate inference algorithm is compatible with the structured perceptron. \citet{Martins/etal/2009} provided risk bounds for learning with relaxations of integer linear programming based inference that is common in natural language applications. Training structured SVMs with approximate inference was considered in the works of \citet{Finley/Joachims/08} and \citet{Klein/etal/08} with mixed (negative and positive) results. The conclusion of \citet{Klein/etal/08}'s empirical study was that structured SVMs trained with exact inference resulted in improved performance when compared to those trained with approximate inference. \citet{Finley/Joachims/08} considered two classes of approximate inference algorithms --- undergenerating (e.g., greedy methods) and overgenerating (e.g., relaxation methods like linear programming and graph cuts) algorithms --- and showed that models trained with overgenerating methods have theoretical and empirical advantages over undergenerating methods. The aforementioned mixed results motivated us to consider efficient learning methods for structured prediction that tries to avoid using any inference algorithm, be it exact or approximate, during training.

\section{Hardness Results}\label{sc:hard}
In the following, we will also be interested in the set of hypotheses potentially occurring as solutions to the optimisation problem \eq{eq:cvx} and denote it as $\Hcal_{\mathrm{opt}} = \{Q(D)\mid D\subseteq \Xcal\times\Ycal\}$.

First, we show that the assumptions described in the previous section do not hold for several relevant output sets. In particular, we show that they do not hold if the non-optimality decision problem for a given $(\Ycal,\Hcal_{\mathrm{opt}})$ is NP-hard. This decision problem is the complement of the \emph{optimality} problem and is defined as deciding if for a given $h\in\Hcal_\mathrm{opt}$, $x\in\Xcal$ and $y\in\Ycal$, a $z\in\Ycal$ exists such that $h(x,z) > h(x,y)$. Second, we show that for the specific case of undirected cycles (route prediction), non-optimality is indeed NP-complete.  Our hardness result gains further significance as we can also show that this case can indeed occur for a specific set of observations. Third, we turn to a class of problems for which the output forms particular set systems, show that in this case the assumptions \emph{decoding}, \emph{separation}, and \emph{optimality} are not contained in P, and note that \emph{decoding} is often hard as it corresponds to edge deletion problems \cite{Yannakakis78}.

\subsection*{Representation in Output Space}\label{su:rep}
%In this section, we will use route prediction as an example and show that the assumptions  made by state-of-the-art structured output training algorithms do not hold in this case.
As described in the previous section, state-of-the-art structured output learning algorithms  assume (at least) that deciding if an output structure with higher score than a given one exists is in NP. With the definitions given above, we formally define $(\Ycal,\Hcal_\mathrm{opt})-$ \textsc{Optimality} as:
\begin{equation*}
h\in \Hcal_\mathrm{opt}, x\in\Xcal,  y\in\Ycal \mapsto \left(\nexists z\in\Ycal: h(x,z) > h(x,y)\right)\ .
\end{equation*}
Let $k_\Xcal:\Xcal\times\Xcal\to\R$ be the kernel function on the input space $\Xcal$. We assume that a polynomial time computable map $\psi:\Ycal\to\R^d$ is defined and will refer to the inner product under this map by $k_\Ycal:y,y'\mapsto\ip{\psi(y)}{\psi(y')}$. For the joint kernel of inputs and outputs, we consider the tensor product kernel $k[(x,y), (x',y')] = k_\Xcal(x,x') k_\Ycal(y,y')$, which has found applications in many important domains \cite{jacob_protein-ligand_2008,erhan_collaborative}. We refer to the Hilbert spaces corresponding to $k_\Xcal, k_\Ycal, k$ as $\Hcal_\Xcal,\Hcal_\Ycal,\Hcal$, respectively.

The strong representer theorem \cite{Schoelkopf/etal/01} holds for all minimisers $h^*\in\Hcal_{\mathrm{opt}}$ of the optimisation problem \eq{eq:cvx}. % (a short proof is included in Section~\ref{su:trep}).
It shows that $h^*\in\mathbf{span}\{k[(x_i,z), (\cdot,\cdot)]\mid i\in\numset{m}, z\in\Ycal\}$. That is,
\begin{align*}
h^*(x,y) &= \sum_{i\in\numset{m}, z\in\Ycal} \balpha_{iz} k[(x_i,z),(x,y)]%\\
         = \sum_{i\in\numset{m}, z\in\Ycal} \balpha_{iz} k_\Xcal(x_i,x) \ip{\psi(z)}{\psi(y)}\\
         &= \ip{\sum_{i\in\numset{m}, z\in\Ycal} \balpha_{iz} k_\Xcal(x_i,x) \psi(z)}{\psi(y)}\ .
\end{align*}
This motivates us to first consider $(\Ycal,\psi)-$ \textsc{Optimality}, defined as:
\begin{equation*}
w\in\mathbf{span}\{\psi(y) \mid y\in \Ycal\},
%\mathbf{span}(\phi[\Ycal]),  
y\in\Ycal \mapsto \left(\nexists z\in\Ycal: \ip{w}{\psi(z)} > \ip{w}{\psi(y)}\right)\ .
\end{equation*}
To show that $(\Ycal,\psi)-$ \textsc{Optimality} is not in NP, it suffices to show that $(\Ycal,\psi)-$ \textsc{Non-Optimality} is NP-hard. Formally, we define $(\Ycal,\psi)-$ \textsc{Non-Optimality} as:
\begin{equation*}
w\in\mathbf{span}\{\psi(y) \mid y\in \Ycal\},
%\mathbf{span}(\phi[\Ycal]),  
y\in\Ycal \mapsto \left(\exists z\in\Ycal: \ip{w}{\psi(z)} > \ip{w}{\psi(y)} \right)\ ,
\end{equation*}
and correspondingly $(\Ycal,\Hcal_\mathrm{opt})-$ \textsc{Non-Optimality}.

\subsection*{Route Prediction --- Hardness of Finding Cyclic Permutations}
We now consider the route prediction problem introduced in Section~\ref{sc:whypredictcs} as an instance for which $(\Ycal,\Hcal_\mathrm{opt})-$ \textsc{Non-Optimality} is NP-complete and for which thus the assumptions made by state-of-the-art structured prediction algorithms do not hold.

In the route prediction problem, each $x_i$ represents a sequence of features such as an individual person, a day, and a time of the day; $\Sigma^\mathrm{cyc}$ is a set containing points of interest; $\Ycal^\mathrm{cyc}$ is the set of cyclic permutations of subsets of $\Sigma$ (we are interested in cyclic permutations as we assume that each individual starts and ends each route in the same place, e.g., his/her home); and $d^\mathrm{cyc}=\Sigma\times\Sigma$. We represent a cyclic permutation by the set of all neighbours. For instance, the sequences $\{abc,bca,cab,cba,acb,bac\}$ are equivalent and we use $\{\{a,b\},\{b,c\},\{c,a\}\}$ to represent them. Furthermore, we define $\psi^\mathrm{cyc}_{(u,v)}(y)=1$ if $(u,v)\in y$, i.e., $v$ and $u$ are neighbours in $y$, and $0$ otherwise. 

\hide{
\begin{proposition}\label{prop:ham}
With all constants and functions defined as above, $(\Ycal^\mathrm{cyc},\psi^\mathrm{cyc})-$ \textsc{Non-Optimality} is not in P unless P=NP.
\end{proposition}
\begin{proof}
Repeated calls to a polynomial time $(\Ycal^\mathrm{cyc},\psi^\mathrm{cyc})-$ \textsc{Non-Optimality} subroutine could be used to solve the Hamiltonian cycle problem by the following algorithm:
\begin{algorithm}[h!]
\begin{algorithmic}[1]
\REQUIRE Graph $G=(V,E)$
\ENSURE G has a Hamiltonian cycle
\STATE Let $c$ be any cyclic permutation of $V$
\STATE Let $w$ be a function mapping a graph to its adjacency matrix
\WHILE{$(\Ycal^\mathrm{cyc},\psi^\mathrm{cyc})-$ \textsc{Non-Optimality}$(w(G),c)$}
\STATE Let $G' \leftarrow G$
\FOR{$e \in E$}
\IF{$(\Ycal^\mathrm{cyc},\psi^\mathrm{cyc})-$ \textsc{Non-Optimality}$(w(G'-e),c)$}
\STATE $G' \leftarrow G' - e$
\ENDIF
\ENDFOR
\STATE Let $c$ be any cyclic permutation of $V$ containing $G'$
\ENDWHILE
\end{algorithmic}
\end{algorithm}
\\ Here, for a graph $(V,E)$ we denote $(V,E)-e =(V,E \setminus \{e\})$. To see that this algorithm provides a polynomial time Turing reduction, it is sufficient to observe that $|\{|\ip{w(G)}{\psi^\mathrm{cyc}(c)}|: c \in \Ycal^\mathrm{cyc}\}| < |V|^2$.
\end{proof}

We will now prove the stronger lemma:
}
\begin{lemma}\label{le:w}
With all constants and functions defined as above, $(\Ycal^\mathrm{cyc},\psi^\mathrm{cyc})-$ \textsc{Non-Optimality} is NP-complete.
\end{lemma}
\begin{proof}
The proof is given by a Karp reduction of the Hamiltonian path problem. Let $G=(V,E)$ be an arbitrary graph. Wlog, assume that \mbox{$V \cap \numset{3} = \emptyset$}. Construct a weighted graph $\tilde{G}$ on the vertices of $V \cup \numset{3}$ with adjacency matrix 
\begin{equation*}
\tilde{w}= (|V|-2) \cdot \psi^\mathrm{cyc}(\{\{1,2\},\{2,3\},\{3,1\}\}) + \sum\limits_{e \in E}\psi^\mathrm{cyc}(e) \ .
\end{equation*}
Now, $(\Ycal^\mathrm{cyc},\psi^\mathrm{cyc})-$ \textsc{Non-Optimality}($\tilde{w},\{\{1,2\},\{2,3\},\{3,1\}\}$) holds iff there is a cyclic permutation on $V \cup \numset{3}$ that has an intersection with $G$ of size $|V|-1$. The intersection of any graph with a graph is a set of paths in $G$. As the total number of edges in the intersection is $|V|-1$, the intersection is a path in $G$. A path in $G$ with $|V| - 1$ edges is a Hamiltonian path.
\end{proof}

\begin{theorem}\label{th:subopt}
With all constants and functions defined as above, $(\Ycal^\mathrm{cyc},\Hcal_{\mathrm{opt}}^\mathrm{cyc})-$ \textsc{Non-Optimality} is NP-complete.
\end{theorem}
\begin{proof}
We consider $\Hcal_{\mathrm{opt}}^\mathrm{cyc}$ for $\lambda=0$ and construct training data and a function that satisfies all but one constraint if and only if the graph has a Hamiltonian cycle.

For an arbitrary graph $G=(V,E)$ let $\tilde{G}, \tilde{w}$ be defined as above and let $\Xcal =2^E$ with $k_{\Xcal}(e,e')=|e \cap e'|$. With $m =|E|+1$, $\{x_i \mid  i \in \numset{|E|}\} = E$, and $x_m = E$ we choose the training data $D$ as
\begin{equation*}
\{ (\{e\}, \{e\}) \mid  e \in E \} \cup \{(E, \{\{1, 2\}, \{2, 3\}, \{3, 1\}\}) \} \ .
\end{equation*}
For $i \in \numset{|E|}$, let $\balpha_{iz}=1/2$ for $z=y_i$ and $\balpha_{iz}=0$ otherwise. For $i=m$, let $\balpha_{iz}=(|V|-2)/2$ for $z=y_i$ and $\balpha_{iz}=0$ otherwise. For $i \in \numset{|E|}$, we then have
\begin{equation*}
h(x_i,y_i) = \ip{\sum_{j\in\numset{m}, z\in\Ycal^\mathrm{cyc}} \balpha_{jz} k_\Xcal(x_j,x_i) \psi(z)}{\psi(y_i)} = 1
\end{equation*}
and $y \neq y_i \dann h(x_i,y)=0$. Thus there are no violated constraints for $i\in\numset{|E|}$ and $h$ is indeed optimal on this part of the data. Furthermore, for all $y \in \Ycal^\mathrm{cyc}$:
\begin{equation*}
h(x_{m},y) = \ip{\sum_{i\in\numset{m}, z\in\Ycal^\mathrm{cyc}} \balpha_{iz} k_\Xcal(x_i,x_m) \psi(z)}{\psi(y)} = \ip{\tilde{w}}{\psi(y)}\ .
\end{equation*}
Together with Lemma~\ref{le:w} this gives the desired result. 
\end{proof}

\hide{
\subsection*{Partially Ordered Sets}\label{su:poset}
\todo{check}
Consider $\Sigma=\numset{N}\times\numset{N}$; $\Ycal\subseteq2^\Sigma$ such that all $y\in\Ycal$ form a poset (an acyclic directed graph closed under transitivity) $(\numset{N},y)$; as well as $\psi:\Ycal\to\R^{\Ycal}$ with $\psi_{uv}(z)=1$ if $(u,v)\in z$, $\psi_{uv}(z)=-1$ if $(v,u)\in z$, and $\psi_{uv}(z)=0$ otherwise.

\begin{theorem}
The optimality problem for posets is coNP-complete.
\end{theorem}
\begin{proof}
We will show that the problem of finding the minimum feedback arc set in tournaments (FAST) \cite{Charbit/etal/07}, which is known to be NP-hard, is reducible to the non-optimality problem for posets. We know that the non-optimality problem is in NP due to the existence of a short certificate. This would then imply that the problem is NP-complete.

Consider directed graphs $G=(V,E)$ with unit edge weights. Suppose the non-optimality problem is in P. The weight of the optimal (maximal) poset can be found by testing for non-optimality for all weights $k \in \numset{|E|}$ and counting the number ($K$) of 'yes' instances. The optimal poset can then be found by deleting edges in succession from the given directed graph and testing every resulting graph for non-optimality with weight $K$. This procedure can also be used to solve FAST in polynomial time. This completes the reduction.
\end{proof}
}
\subsection*{Connections to other Assumptions}\label{su:conn}
Recall the assumptions made by existing structured prediction algorithms, namely, decoding, separation, and optimality (cf. Section~\ref{sc:eff_learn}). Define $(\Ycal,\Hcal)-$ \textsc{Decoding} as
\begin{equation*}
h\in\Hcal, x\in\Xcal \mapsto \argmax_{z\in\Ycal} h(x,z) \ ,
\end{equation*}
$(\Ycal,\Hcal)-$ \textsc{Separation} as
\begin{equation*}
h\in\Hcal, x\in\Xcal,  y\in\Ycal \mapsto 
\begin{cases}
z &\text{for some $z\in\Ycal$ with } h(x,z)>h(x,y) \\
\emptyset &\text{otherwise}
\end{cases}\ ,
\end{equation*}
%if one exists or proof that none exists otherwise. 
and $(\Ycal,\Hcal)-$ \textsc{Optimality} as
\begin{equation*}
h\in\Hcal, x\in\Xcal,  y\in\Ycal \mapsto \nexists z\in\Ycal: h(x,z) > h(x,y) \ .
\end{equation*}

\begin{proposition}\label{prop:conn}\ 
\begin{enumerate}
\item[\phantom{$\dann$}] $(\Ycal,\Hcal)-$ \textsc{Non-Optimality} is NP-complete.
\item[$\dann$] $(\Ycal,\Hcal)-$ \textsc{Optimality} is coNP-complete.
\item[$\dann$] $(\Ycal,\Hcal)-$ \textsc{Separation} is coNP-hard.
\item[$\dann$] $(\Ycal,\Hcal)-$ \textsc{Decoding} is coNP-hard.
\end{enumerate}
\end{proposition}
\begin{proof}
The result follows immediately by observing that (1) optimality is the complement of non-optimality, (2) any separation oracle can be used to decide optimality, and (3) any algorithm for decoding can be used as a separation oracle. 
\end{proof}

\begin{corollary}
Unless coNP=NP, $(\Ycal^\mathrm{cyc},\Hcal^\mathrm{cyc})-$ \textsc{Optimality} is not contained in NP. Unless P=coNP, there is no polynomial time algorithm for
\begin{enumerate}
\item $(\Ycal^\mathrm{cyc},\Hcal^\mathrm{cyc})-$ \textsc{Separation}, and
\item $(\Ycal^\mathrm{cyc},\Hcal^\mathrm{cyc})-$ \textsc{Decoding}.
\end{enumerate}
\end{corollary}
\begin{proof}
The result follows immediately from Theorem~\ref{th:subopt} and Proposition~\ref{prop:conn}. 
\end{proof}

Lastly, the decision problem $(\Ycal,\Hcal)-$ \textsc{Optimal-Value}
\begin{equation*}
\beta\in\R, h\in\Hcal, x\in\Xcal,  y\in\Ycal \mapsto \left( \beta = \max \{h(x,z) \mid z\in\Ycal\} \right)
\end{equation*}
is also of interest.
Following the proof that \textsc{Exact-TSP}\footnote{\textsc{Exact-TSP} is the problem of finding a Hamiltonian cycle of a given lenth.} is $D^P$-complete \cite{Papadimitriou94}\footnote{$D^P$, introduced by \citet{Papadimitriou/Yannakakis/82}, is the class of languages that are the intersection of a language in NP and a language in coNP, i.e., $D^P = \{L_1 \cap L_2: L_1 \in NP, L_2 \in coNP\}$. Note that this is not the same as $NP \cap coNP$, and that $NP, coNP \subseteq D^P$.}, it can be shown that $(\Ycal^\mathrm{cyc},\Hcal^\mathrm{cyc})-$ \textsc{Optimal-Value} is $D^P$-complete. The significance of this result is that optimal-value can be reduced to separation and decoding, providing even stronger evidence that neither of these problems can be solved in polynomial time.

%Following the derivation in Section \ref{su:rep} we can correspondingly define $(\Ycal,\phi)-$ \textsc{Decoding}, $(\Ycal,\phi)-$ \textsc{Separation}, and $(\Ycal,\phi)-$ \textsc{Optimality}. 

\subsection*{Set Systems}\label{su:sesy}
We now consider set systems $(\Sigma,\Ycal^\pi)$ with $\Ycal^\pi=\{y\in 2^\Sigma\mid \pi(y)=1\}$ where $\pi:2^\Sigma\to\{\pm1\}$, and let $\psi^\in:\Ycal\to\{0,1\}^{\Sigma}$ be the indicator function $\psi^\in_e(y)=1$ if $e\in y$ and $0$ otherwise. An independence system is a set system for which $y'\subseteq y \in \Ycal^\pi$ implies $y' \in \Ycal^\pi$, i.e., for which $\pi$ is hereditary. A particularly interesting family of independence systems corresponds to hereditary properties on the edges of a graph \mbox{$G=(V,E)$} where $E\subseteq\Sigma=\{U\subseteq V\mid |U|=2\}$. \hide{A property is called hereditary if $F\subseteq F'$ implies $\pi(F)\geq \pi(F')$, for all $F,F'\subseteq E$, i.e., hereditary properties correspond to independence systems on the edges of the graph.} In this case, decoding corresponds to minimum edge deletion problems, which are NP-hard for several important hereditary graph properties like planar, outerplanar, bipartite graph, and many more \cite{Yannakakis78}. \hide{As closure under subsets implies closure under intersection, the algorithm given in the proof of Proposition~\ref{prop:ham} is sufficient to show that optimality, separation, and decoding are not contained in P (unless P=NP) for all NP-hard hereditary properties on the edges of a graph.}

\begin{proposition}\label{prop:set}
With all constants and functions defined as above, $(\Ycal^\pi,\psi^\in)-$ \textsc{Non-Optimality} is in $P$ if and only if the minimum edge deletion problem corresponding to the hereditary property $\pi$ is in $P$.
\end{proposition}

\begin{algorithm}[h!] 
\begin{algorithmic}[1]
\REQUIRE Graph $G=(V,E)$
\ENSURE A maximum subgraph of $G$ that satisfies $\pi$
\STATE Let $Y \leftarrow \emptyset$
\WHILE{$(\Ycal^\pi,\psi^\in)-$ \textsc{Non-Optimality}$(w(E),Y)$}
\STATE Let $F \leftarrow E$;
\FOR{$e \in E \setminus Y$}
\IF{$(\Ycal^\pi,\psi^\in)-$ \textsc{Non-Optimality}$(w(F\setminus\{e\}),Y)$}
\STATE $F \leftarrow F \setminus \{e\}$
\ENDIF
\ENDFOR
\FOR{$e \in E $}
\IF{$(\Ycal^\pi,\psi^\in)-$ \textsc{Non-Optimality}$\left( \frac{|F\setminus Y|-|Y\setminus F|-1}{|F\setminus Y|}w(F\setminus Y) + w(Y \cap F \setminus \{e\}), Y\right)$}
\STATE $F \leftarrow F \setminus \{e\}$
\ENDIF
\ENDFOR
\STATE Let $Y \leftarrow F$
\ENDWHILE
\end{algorithmic}
\caption{An algorithm for solving the minimum edge deletion problem for some hereditary property $\pi$ given a (non-)optimality oracle.}\label{alg:reduction}
\end{algorithm}
\begin{proof}
( $\Rightarrow$ ) For the graph $G = (V,E)$, let $w$ be a function mapping a set $F \subseteq E$ to the adjacency matrix of $(V,F)$. We give a Turing reduction from the minimum edge deletion problem to non-optimality in Algorithm~\ref{alg:reduction}. To see this, it is sufficient to observe that: (i) the \emph{while} loop continues as long as $E$ contains a subset that is larger than $|Y|$ and satisfies $\pi$; as $\ip{w(E)}{\psi^\in(Y)}$ increases in each iteration by one, the number of iterations is bounded from above by $|\{|\ip{w(E)}{\psi^\in(Z)}|~:~Z \in \Ycal^\pi\}| < |E|$, (ii) the first \emph{for} loop (lines 4--8) removes edges not contained in $Y$ from $F$ while ensuring that $F$ contains a subset that is larger than $|Y|$ and satisfies $\pi$; therefore $F \setminus Y \neq \emptyset$; and because of hereditary $\max\{|X|~:~X \subseteq F \land \pi(X)\} = |Y|+1$, and (iii) the second \emph{for} loop (lines 9--13) removes edges of $Y$ from $F$ while ensuring that $F$ still contains a subset that is larger than $|Y|$ and satisfies $\pi$. As the removal of these edges from the adjacency matrix $(w)$ will shrink $\ip{w}{\psi(Y)}$, the weight of the edges in $F \setminus Y$ is reduced accordingly.

( $\Leftarrow$ ) It remains to observe that to decide whether $Y$ is non-optimal, it suffices to find a maximum structure and compare its size with $Y$.
\end{proof}

For properties $\pi$ that are non-hereditary, two modifications of the algorithm arise: (a) The constant in line $10$ may be larger than 1; its precise value can be found by first increasing it from $c = 0$ until $\lnot (\Ycal^\pi, \psi^\in)-$ \textsc{Non-Optimality} $\left( \frac{|F\setminus Y|-|Y\setminus F|-c}{|F\setminus Y|}w(F\setminus Y) + w(Y \cap F), Y\right)$. (b) The structure $F$ itself is not necessarily contained in $\Ycal^\pi$; a polynomial time subroutine  for finding a $X \supseteq F$ such that $X \in \Ycal^\pi$ is needed additionally.

\section{Two New Assumptions}\label{sc:assumptions}
We have thus far discussed the assumptions mabe by existing structured prediction algorithms to ensure efficient learning. We have also seen, in the form of hardness results, that these assumptions do not hold for several combinatorial structures thereby exposing the limitations of existing algorithms to learn efficiently to predict combinatorial structures. We are now ready to introduce two new assumptions, and provide several examples of combinatorial structures and applications in machine learning where these assumptions hold. These assumptions are based on counting and sampling combinatorial structures and will be elucidated in the following sections.

\subsection{The Counting Assumption} 

The major difficulty in structured output learning is to handle the exponentially many constraints in the optimisation problem \eq{eq:cvx}. While successively adding violated constraints is feasible under several assumptions, in the previous section we discussed cases like route prediction where none of these assumptions hold.

In the following, we will first show, considering again cyclic permutations as a concrete example, that counting the number of super-structures can be feasible even if there are exponentially many of them and even if the assumptions of decoding, separation, and optimality do not hold. The counting assumption is stated as follows:
\begin{assumption}
Denote by $\psi: \Ycal \to \R^d$ the finite dimensional embedding of the output space $\Ycal$. It is possible to efficiently compute the quantities
\begin{equation*}
|\Ycal|, \quad \Psi = \sum_{y\in\Ycal} \psi (y), \quad \text{and}\quad C = \sum_{y\in\Ycal}  \psi(y) \psi^\top (y) \ .
\end{equation*}
\end{assumption}
\hide{
Following this insight, we aim at a structured output learning algorithm that is efficient whenever super-structure counting is feasible. Indeed, whenever we are minimising a quadratic upper bound on the loss, we can derive an unconstrained optimisation problem accessing the output set only by means of $|\Ycal^\mathrm{cyc}|$, $\Psi^\mathrm{cyc}$ and $C^\mathrm{cyc}$. To solve this optimisation problem, we give the gradient as well as the Hessian-vector multiplication.  
}
\hide{We now present several combinatorial structures for which the above assumption holds.}

\subsubsection{Route Prediction --- Counting Cyclic Permutations}
% counting cyclic permutations
For a given alphabet $\Sigma$, we are now interested in computing $|\Ycal^\mathrm{cyc}|$, the number of cyclic permutations of subsets of $\Sigma$. For a subset of size $i$ there are $i!$ permutations of which $2i$ represent the same cyclic permutation. That is, there are $(i-1)!/2$ cyclic permutations of each subset of size $i$, and for an alphabet of size $N=|\Sigma|$ there are
\begin{equation*}
|\Ycal^\mathrm{cyc}| = \sum_{i=2}^N\begin{pmatrix}N\\i\end{pmatrix} \frac{(i-1)!}{2}
\end{equation*}
different cyclic permutations of subsets of $\Sigma$.

Computing $\Psi^\mathrm{cyc}$ is simple. For each pair of neighbours, there are $N-2$ remaining vertices, and for each subset of these of size $i$, there are $2(i+1)!$ permutations, of which $2(i+1)$ represent the same cyclic permutation:
\begin{equation*}
\Psi^\mathrm{cyc} = \one \sum_{i=0}^{N-2}\begin{pmatrix}N-2\\i\end{pmatrix} i! \ ,
\end{equation*}
where $\one$ is the vector of all ones.

It remains to compute $C^\mathrm{cyc}$. Each element of this matrix is computed as 
\begin{equation*}
C^\mathrm{cyc}_{e,e'} = \sum_{y\in \Ycal^\mathrm{cyc}} \psi^\mathrm{cyc}_e (y) \psi^\mathrm{cyc}_{e'}(y) \ .
\end{equation*}
For $|e \cap e'| > 0$, we have
\begin{equation*}
C^\mathrm{cyc}_{e,e'} = \one \sum_{i=0}^{N-|e \cup e'|}\begin{pmatrix}N-|e \cup e'|\\i\end{pmatrix} i! \ ,
\end{equation*}
and for $|e \cap e'| = 0$, we have
\begin{equation*}
C^\mathrm{cyc}_{e,e'} = \one \sum_{i=0}^{N-4}\begin{pmatrix}N-4\\i\end{pmatrix} 2(i+1)! \ .
\end{equation*}

%We are now ready to investigate structured output training algorithms based on these counting formulae.

%\subsubsection{Counting Super-Structures}\label{sc:counting}
\hide{
Having introduced an algorithm that can be trained whenever super-structure counting is feasible, we now show that this assumption not only holds for route prediction but also for a large class of other output sets. For some of these output sets efficient decoding is trivial, whereas for others decoding is infeasible, showing that our super-structure counting assumption is orthogonal to the assumptions made in the structured prediction literature.
%\todo{fit in}
%as before
}
\subsubsection{Simple Set System Cases}
We first consider the general $\ell$-label prediction problem where $\Ycal=\{Z\subseteq\Sigma \mid |Z|=\ell\}$ with $\psi:\Ycal\to\R^{\Sigma}$ defined as $\psi_i(Z) = 1$ if $i\in Z$ and $0$ otherwise. This setting generalises the usual multi-class ($\ell=1$) and multi-label (by summing over all $\ell\leq|\Sigma|$) settings. For general $\ell\in\numset{\Sigma}$ we have (with $|\Sigma|=d$),
\begin{equation*}
|\Ycal|=\begin{pmatrix}d\\\ell\end{pmatrix};\hspace{\stretch{1}}\Psi=\one\begin{pmatrix}d-1\\\ell-1\end{pmatrix};\hspace{\stretch{1}}
C=\one\begin{pmatrix}d-2\\\ell-2\end{pmatrix} + \iden\begin{pmatrix}d-1\\\ell-1\end{pmatrix}\ . \hspace{\stretch{2}}
\end{equation*}

As special cases we have for multi-class ($\Ycal=\Sigma$) that $\Psi = \one, C = \iden$, and for multi-label ($\Ycal=2^\Sigma$) that $\Psi =  2^{|\Sigma|-1}\one, C = 2^{|\Sigma|-2}\iden +  2^{|\Sigma|-2}\one$.

For both of these simple cases, exact decoding is very simple. For a given (test) instance $x$, let $\kappa\in\R^m$ with $\kappa_i=k_{\Xcal}(x_i,x)$. For the multi-class case decoding is $\hat y = \argmax_{e\in\Sigma} [\balpha \kappa]_e$. For the multi-label case it is $\hat y = \{ e\in\Sigma \mid [\balpha \kappa]_e \geq 0 \}$. Hence, we could in this case also apply separation based learning algorithms.

\hide{Note, however, that this simple treatment of multi-label prediction has several disadvantages. To see this, consider the simple example of predicting what (mixed) drinks a certain person likes. Say, one likes good scotch and good wine. In reality this hardly implies that the same person likes mixing good wine with good scotch. This, more sophisticated approach is tackled later in Section~\ref{su:clique} as \emph{clique prediction}.}

\subsubsection{Simple Non-Set System Cases}
We now consider poset regression. Let $\Ycal\subset\Sigma$, $\psi:\Ycal\to\R^{\Sigma}$ and let $(\Sigma,\succ)$ be a poset. With $\psi_i(z) = 1$ if $z \succ i$ and $\psi_i(z) = 0$ otherwise, we have $\Psi_i = |\{k\in\Sigma\mid k\succ i\}|$ and $C_{ij} = |\{k\in\Sigma\mid k\succ i \wedge k\succ j\}|$.
As a special case, we have the ordinal regression problem where $\Ycal=\Sigma=\numset{d}$ (with $d$ ordinal values), $\psi:\Ycal\to\R^{\Sigma}$ with $\psi_i(z) = 1$ if $z \geq i$ and $\psi_i(z) = 0$ otherwise. In this case $\Psi_i = |\Sigma|-i$ and $C_{ij} = |\Sigma|-\max(i,j)$. Note that hierarchical classification is also a special case of poset regression where $\succ$ forms a directed tree. In both cases, decoding can be done by exhaustively testing only $|\Sigma|$ alternatives.

\subsubsection{Permutations}\label{su:larank}
Let $\Ycal$ be the set of permutations of $\Sigma$ and let $\psi:\Ycal\to\R^{\Sigma\times\Sigma}$. Then $|\Ycal| = |\Sigma| !$ and with $\psi_{(uv)}(z)=1$ if $u \succ_z v$, $\psi_{(uv)}(z)=-1$ if $v \succ_z u$, and $\psi_{(uv)}(z)=0$ otherwise, we have $\Psi = \zero$, and
\begin{align*}
C_{(uv)(u'v')} = 
\begin{cases}
-|\Sigma|! &\text{if}\ u = v' \wedge u'=v  \\ 
+|\Sigma|! &\text{if}\ u = u' \wedge v =v' \\ 
\frac{+|\Sigma|!}{3} &\text{if}\ u = u' \text{ xor } v = v' \\ 
\frac{-|\Sigma|!}{3} &\text{if}\ u = v' \text{ xor } v = u' \\ 
0 &\text{otherwise}
\end{cases}
\end{align*}
The assumptions made in the literature are unlikely to hold in this case as the `without any cycle of length $\leq \ell$' edge deletion problem is NP-complete \cite{Yannakakis78} for any fixed $\ell \geq 4$.

\subsubsection{Posets and Partial Tournaments}\label{su:poset}
Consider $\Sigma=\numset{N}\times\numset{N}$; $\Ycal\subseteq2^\Sigma$ such that all $y\in\Ycal$ form a poset (an acyclic directed graph closed under transitivity) $(\numset{N},y)$; as well as $\psi:\Ycal\to\R^{\Ycal}$ with $\psi_{uv}(z)=1$ if $(u,v)\in z$, $\psi_{uv}(z)=-1$ if $(v,u)\in z$, and $\psi_{uv}(z)=0$ otherwise. To the best of our knowledge, no exact way to compute $C_{e,e'}$ is known\hide{ but it is strongly related to so called `correlation inequalities' that could be used to bound and hence approximate $C_{e,e'}$}. However, we can relax $\Ycal$ to $\tilde{\Ycal}\subseteq2^\Sigma$ such that all $y\in\Ycal$ form a partial tournament (a directed graph with no cycles of length two). With $\eta=N(N-1)/2$ we have $|\tilde{\Ycal}|=3^\eta$, $\Psi=\zero$, and
\begin{align*}
\tilde{C}_{(uv),(u'v')} &= 
\begin{cases}
-2 \cdot 3^{|\eta|-1} &\text{if}\ u = v' \wedge u'=v  \\ 
+2 \cdot 3^{|\eta|-1} &\text{if}\ u = u' \wedge v =v' \\ 
+2 \cdot 3^{|\eta|-2} &\text{if}\ u = u' \text{ xor } v = v' \\ 
-2 \cdot 3^{|\eta|-2} &\text{if}\ u = v' \text{ xor } v = u' \\ 
0 &\text{otherwise}
\end{cases}
\end{align*}
The assumptions made in the literature are unlikely to hold in this case as the `transitive digraph' edge deletion problem is NP-complete \cite{Yannakakis78}.

\subsubsection{Graph Prediction}\label{su:graph}\label{su:clique}
Consider graphs on a fixed set of vertices, that is $\Sigma=\{U\subseteq\numset{N}\mid |U|=2\}$ and a property $\pi:2^\Sigma\to\Omega$ such as acyclicity, treewidth bounded by a given constant, planarity, outerplanarity bounded by a constant, clique etc.
Let $\Ycal^\pi$ and $\psi^\in$ be defined as in Section~\ref{su:sesy}.
For properties like clique, we can compute $\Ycal,\psi,C$:
\begin{equation*}
|\Ycal^\mathrm{clique}| = 2^N,\quad
\Psi^\mathrm{clique}_{\{u,v\}} =\sum_{i=2}^N \begin{pmatrix}N-2\\i-2\end{pmatrix} ,\quad
C^\mathrm{clique}_{e,e'} = 
\sum\limits_{i=|e \cap e'|}^{N} \begin{pmatrix}N-|e \cap e'|\\i-|e \cap e'|\end{pmatrix} \ .
\end{equation*}

For other properties, no way to compute $C$ might be known but we can always relax $\Ycal$ to $\tilde{\Ycal}=2^\Sigma$. We then have $\tilde{\Psi} = 2^{|\Sigma| - 1}$ and $\tilde{C}_{e,e'}=2^{|\Sigma| - |e\cup e'|}$.

\subsection{The Sampling Assumption}
The sampling assumption pertains to discriminative probabilistic models for structured prediction. The sampling assumption is stated as follows:
\begin{assumption}
It is possible to sample efficiently from the output space $\Ycal$ exactly and uniformly at random.
\end{assumption}
We now describe three combinatorial structures with their corresponding application settings in machine learning. For each of these structures, we show how to obtain exact samples uniformly at random.
\hide{
We describe three combinatorial structures with their corresponding application settings in machine learning. For each of these structures, we show how to obtain exact samples uniformly at random. Together with the `meta' approach presented in the previous section, it is then possible to obtain exact samples of these structures from exponential family distributions considered in this work. Therefore, we have all the necessary ingredients to approximate the partition function.
}
\subsubsection{Vertices of a hypercube}
The set of vertices of a hypercube is used as the output space in multi-label classification problems (see, for example, \citet{Elisseeff/Weston/01}). An exact sample can be obtained uniformly at random by generating a sequence (of length $d$, the number of labels) of bits where each bit is determined by tossing an unbiased coin. 

\subsubsection{Permutations}
The set of permutations is used as the output space in label ranking problems (see, for example, \citet{Dekel/etal/03}). An exact sample can be obtained uniformly at random by generating a sequence (of length $d$, the number of labels) of integers where each integer is sampled uniformly from the set $\numset{d}$ \emph{without} replacement.

\subsubsection{Subtrees of a tree}
Let $T=(V,E)$ denote a directed, rooted tree with root $r$. Let $T'$ denote a subtree of $T$ rooted at $r$. Sampling such rooted subtrees from a rooted tree finds applications in multi-category hierarchical classification problems as considered by \citet{Cesa-Bianchi/etal/06} and \citet{Rousu/etal/06}. We now present a technique to generate exact samples of subtrees uniformly at random. The technique comprises two steps. First, we show how to count the number of subtrees in a tree. Next, we show how to use this counting procedure to sample subtrees uniformly at random. The second step is accomplished along the lines of a well-known reduction from uniform sampling to exact/approximate counting \cite{Jerrum/etal/86}.

First, we consider the counting problem. Let $v \in V$ be a vertex of $T$ and denote its set of children by $\delta^+(v)$. Let $f(v)$ denote the number of subtrees rooted at $v$. Now, $f$ can be computed by using the following recursion:
\begin{equation}\label{eq:tree_count}
f(r) = 1 + \prod\limits_{c \in \delta^+(r)} f(c) \ .
\end{equation}

Next, we consider the sampling problem. Note that any subtree can be represented by a $d$-dimensional vector in $\{0,1\}^d$, where $d=|V|$. A na\"ive approach to generate samples uniformly at random would be the following: generate a sequence of $d$ bits where each bit is determined by tossing an unbiased coin; accept this sequence if it is a subtree (which can be tested in polynomial time). Clearly, this sample has been generated uniformly at random from the set of all subtrees. Unfortunately, this na\"ive  approach will fail if the number of acceptances (subtrees) form only a small fraction of the total number of sequences which is $2^d$, because the probability that we encounter a subtree may be very small. This problem can be rectified by a reduction from sampling to counting, which we describe in the sequel.

We will use the term \emph{prefix} to denote a subtree $T'$ included by another subtree $T''$, both rooted at $r$. Let $L(T')$ denote the set of leaves of $T'$. We will reuse the term \emph{prefix} to also denote the corresponding bit representation of the induced subtree $T'$. The number of subtrees with $T'$ as \emph{prefix} can be computed using the recursive formula \eq{eq:tree_count} and is given (with a slight abuse of notation) by $f(T')=(\prod_{v \in L(T')} f(v)) - |L(T')|$. Now, we can generate a sequence of $d$ bits where each bit $u$ with a prefix $v$ is determined by tossing a biased coin with success probability $f(u)/f(v)$ and is accepted only if it forms a tree with its prefix. The resulting sequence is an exact sample drawn uniformly at random from the set of all subtrees.

\section{Summary}
The purpose of this chapter was to highlight the limitations of existing structured prediction algorithms in the context of predicting combinatorial structures. We have shown how even the weakest assumption made by these algorithms --- \textsc{Optimality} --- is coNP-complete for several classes of combinatorial structures. We then introduced two new assumptions based on counting and sampling combinatorial structures. We have also seen how these assumptions hold for several combinatorial structures and applications in machine learning. As we will see in the next chapters, these two assumptions will result in (i) the design of a new learning algorithm  and (ii) a new analysis technique for discriminative probabilistic models for structured prediction.

\clearemptydoublepage
\chapter{Structured Ridge Regression}\label{ch:csop}
In this chapter, we design a new learning algorithm for structured prediction using the counting assumption introduced in the previous chapter. The algorithm, as we will see in the following sections, is a generalisation of ridge regression for structured prediction. The algorithm can be trained by solving an unconstrained, polynomially-sized quadratic program, and does not assume the existence of polynomial time algorithms for decoding, separation, or optimality. The crux of our approach lies in the polynomial time computation of the vector $\Psi=\sum_{z \in \Ycal} \psi(z)$ and the matrix $C=\sum_{z \in \Ycal} \psi (z) \psi^\top(z)$ leading to a tractable optimisation problem for training structured prediction models.

\section{Ridge Regression}
Given a set of training examples $(x_1,y_1), \ldots, (x_m,y_m) \in \Xcal \times \R$, the goal of regularised least squares regression (RLRS)\cite{Rifkin/Lippert/07} or ridge regression is to find a solution to the regression problem\footnote{The same technique can be used for binary classification. The term regularised least squares classfication (RLSC) was coined by \citet{Rifkin02}.} via Tikhonov regularisation in a reproduding kernel Hilbert space. The following optimisation problem is considered:
\begin{equation}\label{eq:rlsr}
\begin{array}{c}
f^* =
\argmin\limits_{f\in\Hcal}
\frac{\lambda}{2} \|f\|^2 + \frac{1}{2} \sum\limits_{i=1}^m (f(x_i)-y_i)^2 \ ,
\end{array}
\end{equation}
where $\lambda > 0$ is the regularisation parameter. According to the representer theorem \cite{Schoelkopf/etal/01}, the solution to this optimisation problem can be written as 
\begin{equation*}
f^* = \sum\limits_{i=1}^m c_i k(x_i,\cdot) 
\end{equation*}
for some $c \in \R^m$ and a positive definite kernel $k : \Xcal \times \Xcal \to \R$ on the input space. Using this fact, the optimisation problem \eq{eq:rlsr} can be rewritten as
\begin{equation*}
\begin{array}{c}
c^* =
\argmin\limits_{c\in\R^m}
\frac{\lambda}{2} c^\top K c + \frac{1}{2}\|y - Kc\|^2 \ .
\end{array}
\end{equation*}
\hide{Setting the derivative w.r.t. $c$ to $0$,}The optimal solution $c^*$ can be found by solving a system of linear equations
\begin{equation*}\label{eq:rlsr_sol}
(K+\lambda \iden)c^* = y \ .
\end{equation*}

\citet{Rifkin02} discusses the pros and cons of regularised least squares regression in comparison with SVMs. Training an SVM requires solving a convex quadratic optimisation problem, whereas training an RLSR requires the solution of a single system of linear equations. However, the downside of training a non-linear RLSR is that it requires storing ($O(m^2)$ space) and also inverting ($O(m^3)$ time) the entire kernel matrix. Also, the solution of an RLSR is not sparse unlike the solution of an SVM thereby demanding huge computations at test time. In the case of linear RLSR, the optimisation problem \eq{eq:rlsr_sol} can be written as
\begin{equation*}
(XX^\top+\lambda \iden)c^* = y \ ,
\end{equation*}
where $X$ is the input matrix of size $m \times n$. If $n \ll m$, it is possible to solve this system in $O(mn^2)$ operations using the Sherman-Morrison-Woodbury formula \cite{Rifkin02}. Empirically, RLSC was shown to perform as well as SVMs \cite{Rifkin02}.
\hide{\todo{expand - Nystr\"om apx, etc?}}

\section{Training Combinatorial Structures}\label{sc:training}
We now present a learning algorithm for predicting combinatorial structures. Interestingly, the connection to ridge regression is incidental that is established due to manipulating a particular structured loss function in such a way that the resulting optimisation problem remains tractable under the counting assumption.

\subsubsection{Problem Setting}
Let $\Xcal \subseteq \R^n$ be an input space and  $\Ycal$ be the output space of a combinatorial structure. Given a set of training examples $(x_1,Y_1), \ldots, (x_m,Y_m) \in \Xcal \times 2^\Ycal$, the goal is to learn a scoring function $h : \Xcal \times \Ycal \to \R$ that, for each $x_i\in\Xcal$, orders (ranks) $\Ycal$ in such a way that it assigns a higher score to all $y\in Y_i$ than to all $z\in\Ycal\setminus Y_i$. Let $\psi: \Ycal \to \R^d$ be a finite dimensional embedding of $\Ycal$ with the dot-product kernel $k_{\Ycal} = \ip{\psi(y)}{\psi(y')}$. Let $k_\Xcal:\Xcal\times\Xcal\to\R$ be a kernel on $\Xcal$. Denote the joint scoring function on input-output pairs by $h\in\Hcal=\Hcal_\Xcal\otimes\Hcal_\Ycal$ where $\otimes$ denotes the tensor product and $\Hcal_\Xcal, \Hcal_\Ycal$ are the reproducing kernel Hilbert spaces (RKHS) of $k_\Xcal, k_\Ycal$, respectively. Note that the reproducing kernel of $\Hcal$ is then $k[(x,y),(x',y')]=k_\Xcal(x,x')k_\Ycal(y,y')$. We aim at solving the optimisation problem
\begin{equation}\label{eq:basicopt}
\begin{array}{c}
h^* =
\argmin\limits_{h\in\Hcal}
\lambda \|h\|^2 + \sum\limits_{i\in\numset{m}} \ell(h,i) \ ,
\end{array}
\end{equation}
where $\ell:\Hcal\times\numset{m} \to \R$ is the empirical risk on a training instance and $\lambda > 0$ is the regularisation parameter.

\subsubsection{Loss Functions}
\hide{We propose to directly minimise the number of misordered pairs as minimising ranking measures is often more robust, i.e.,}For each $x_i$ we aim at ordering all elements of $Y_i$ before all elements of $\Ycal\setminus Y_i$. Note that traditional (label) ranking methods cannot be applied due to the huge (exponential) size of $\Ycal\setminus Y_i$. We use AUC-loss as the empirical error of the optimisation problem \eq{eq:basicopt}:
\begin{equation}\label{eq:aucloss}
\begin{array}{c}
\ell\hide{^\Delta}_\mathrm{auc}(h,i) = \sum\limits_{y\in Y_i} \sum\limits_{z\in\Ycal\setminus Y_i} \sigma[h(x_i,z)-h(x_i,y)]\hide{ \Delta (z, y_i) }
\end{array}
\end{equation}
where $\sigma$ is the modified step function: $\sigma(a)=+1$ if $a>0$,  $\sigma(a)=0$ if $a<0$, and $\sigma(a)=1/2$ if $a=0$. Our definition of $\ell\hide{^\Delta}_\mathrm{auc}$ differs from the `area under the ROC curve' measure only in the sense that it is not normalised.
To obtain a convex function we bound it from above by the exponential loss 
\begin{equation}\label{eq:exploss}
\begin{array}{c}
\ell_{\exp}(h,i) 
= \sum\limits_{y\in Y_i} \sum\limits_{z\in\Ycal\setminus Y_i} \exp \left[ 1 + h(x_i,z) - h(x_i,y)  \right]%\\
\geq 
%& 
\ell_\mathrm{auc}(h,i)\ .
\end{array}
\end{equation}
Despite being convex the exponential loss does not allow compact formulations in our setting, but using its second order Taylor expansion at $0$, i.e., 
$\exp(a)\approx1+a+\frac{1}{2}a^2$, does.
Ignoring constants that can be accommodated by the regularisation parameter, we get
\begin{equation}\label{eq:texploss}
\begin{array}{ll}
\ell(h,i) & = \sum\limits_{y\in Y_i} \sum\limits_{z\in\Ycal\setminus Y_i} \hide{c_i} [ h(x_i,z) - h(x_i,y) + \frac{1}{2}h^2(x_i,z) \\
& \quad\quad - h(x_i,z) h(x_i,y) + \frac{1}{2}h^2(x_i,y) ]\ .
\end{array}
\end{equation}

The above loss function can be seen as a generalisation of the square loss for structured prediction, and hence the connection to ridge regression is established. Similar loss functions were considered in previous work on structured prediction problems. \citet{Altun/etal/02} introduced the ranking loss \eq{eq:aucloss} for structured prediction and minimised an upper bound given by the exponential loss \eq{eq:exploss} for discriminative sequence labeling. For this specific problem, dynamic programming was used to explicity compute the sums over all possibls sequences efficiently \cite{Altun/etal/02}. A closely related loss function to our approximation \eq{eq:texploss} is the one minimised by least-squares SVM \cite{Suykens/Vandewalle/99} and also its multi-class extension \cite{Suykens99}. Our approach can therefore be seen as an extension of least-squares SVM for structured prediction problems. The main reason behind deviating from the standard max-margin hinge loss is to make the problem tractable. As will become clear in the following sections, using the loss function \eq{eq:texploss} results in a polynomially-sized unconstrained optimisation problem.

\subsubsection{Representer}
The standard representer theorem \cite{Schoelkopf/etal/01} states that there is a minimiser $h^*$ of the optimisation problem \eq{eq:basicopt} with 
\begin{equation*}
h^*\in\Fcal=\lhull\{k[(x_i,z),(\cdot,\cdot)]\mid i\in\numset{m},z\in\Ycal\} \ .
\end{equation*}
It also holds in our case: Without loss of generality we consider $h=f+g$ where $f\in\Fcal$ and $g\in\Hcal$ with $g\bot\Fcal$. Now $h(x_i,z)=\ip{h(\cdot,\cdot)}{k[(x_i,z),(\cdot,\cdot)]}=\ip{f(\cdot,\cdot)}{k[(x_i,z),(\cdot,\cdot)]} + 0 = f(x_i,z)$ as well as $\|h(\cdot,\cdot)\|^2=\|f(\cdot,\cdot)\|^2+\|g(\cdot,\cdot)\|^2$. This shows that for each $h\in\Hcal$ there is an $f\in\Fcal$ for which the objective function of \eq{eq:basicopt} is no larger.

As usually $|\Ycal|$ grows exponentially with the input of our learning algorithm, it is intractable to optimise over functions in $\Fcal$ directly. Let $e_1,\ldots, e_d$ be the canonical orthonormal bases of $\R^d$. We then have that $\{k_\Xcal(x_i,\cdot)\otimes \ip{e_l}{.}\mid i\in\numset{m},l\in\numset{d}\}$ spans $\{k_{\Xcal}(x_i,\cdot) \otimes \ip{\psi(z)}{.} \mid i\in\numset{m},z\in\Ycal\}$. Therefore it is sufficient to optimise over only $md$ variables. We hence consider
%
%With 
%\begin{equation}\label{eq:form}
%\end{equation}
%we hence consider 
\begin{equation}\label{eq:texpoptform}
\balpha^* =
\argmin\limits_{\balpha\in\R^{m\times d}}
\lambda  \|f_{\balpha}\|^2 + \sum_{i\in\numset{m}} \ell(f_{\balpha},i)
\end{equation}
with
\begin{equation*}
f_{\balpha}(x,z)=\sum\limits_{i\in\numset{m},l \in \numset{d}} \balpha_{il} k_{\Xcal}(x_i,x) \ip{e_l}{\psi(z)}%\ip{ u_l(\cdot)}{k_{\Ycal}(z,\cdot)}
\ .
\end{equation*}
%\sum_{\stackrel{ \scriptstyle{i,i'\in\numset{n} } }{ l,l'\in\numset{m} } } \balpha_{il} \balpha_{i'l'}

\subsubsection{Optimisation}\label{su:opt}
\hide{
We now consider cases in which we have a finite dimensional output embedding $\psi:\Ycal\to\R^d$ with $k_\Ycal(y,y')=\fsip{y}{y'}$ and the vector $\Psi=\sum_z \psi(z)$ as well as the matrix $C=\sum_z \psi(z) \psi^\top(z)$ can be computed in polynomial time for all of $\Ycal$. We present several examples in Section~\ref{sc:examples}. If $\psi:\Ycal\to\R^d$ can be computed in polynomial time but computing $C, \Psi$ is hard, we may resort to approximations to $C$ and $\Psi$. Another option would be to relax the set $\Ycal$ to a superset $\tilde{\Ycal}$ such that $\tilde{C},\tilde{\Psi}$ can be computed in polynomial time for $\tilde{\Ycal}$. For set systems we can (in the worst case) relax to $\tilde{\Ycal}=2^\Sigma$ such that $C_{a,b}=2^{|\Sigma|-|a\cup b|}$ with $k_\Ycal(z,z')=|z\cap z'|$. For approximate decoding it might however sometimes be necessary to choose a different kernel. We also give several examples for this case in Section~\ref{sc:examples}.
}
We now exploit the fact that the vector $\Psi=\sum_{z \in \Ycal} \psi(z)$ and the matrix $C=\sum_{z \in \Ycal} \psi(z) \psi^\top(z)$ can be computed in polynomial time for a wide range of combinatorial structures as shown in Section~\ref{sc:assumptions}. If $\psi:\Ycal\to\R^d$ can be computed in polynomial time but computing $C, \Psi$ is hard, we may resort to approximations to $C$ and $\Psi$. Another option would be to relax the set $\Ycal$ to a superset $\tilde{\Ycal}$ such that $\tilde{C},\tilde{\Psi}$ can be computed in polynomial time for $\tilde{\Ycal}$. For set systems we can (in the worst case) relax to $\tilde{\Ycal}=2^\Sigma$ such that $C_{a,b}=2^{|\Sigma|-|a\cup b|}$ with $k_\Ycal(z,z')=|z\cap z'|$.

Denote $f_{\balpha}(x_i,\cdot) = \sum_{j\in\numset{m}, l\in\numset{d}} \balpha_{jl} k_{\Xcal}(x_i,x_j) e_l$ by $f_{\balpha}^i$. Let $Y$ be the matrix $Y\in\R^{m \times d}$ such that $Y_{i\cdot}=\sum_{y\in Y_i}\psi^\top (y)$ and $K$ be the kernel matrix such that $K_{ij}=k_{\Xcal}(x_i,x_j)$. We then have $f^i_{\balpha} = \balpha K_{i\cdot}$ and $F^i_{\balpha} = Y_{i\cdot} f^i_{\balpha}$. We can now express $\ell(f_{\balpha},i)$ using $\Psi$, $C$, and 
\begin{equation*}
\sum_{z\in\Ycal} f(x_i,z) = \ip{f_{\balpha}^i}{\Psi} \ ,
\end{equation*}
\begin{equation*}
\sum_{z\in\Ycal\setminus Y_i} f(x_i,z) = \sum_{z\in\Ycal} f(x_i,z) -  \sum_{y\in Y_i} f(x_i,y) \ ,
\end{equation*}
\begin{equation*}
\sum_{z\in\Ycal} f^2(x_i,z) = f_{\balpha}^i C f_{\balpha}^i \ ,
\end{equation*}
\begin{equation*}
\sum_{z\in\Ycal\setminus Y_i} f^2(x_i,z) = \sum_{z\in\Ycal} f^2(x_i,z) -  \sum_{y\in Y_i} f^2(x_i,y)\ .
\end{equation*}
For the simple setting where $|Y_i|=1$, for all $i \in \numset{m}$, we have
\begin{equation*}
\ell(f_{\balpha},i)
 = \frac{1}{2} f_{\balpha}^i C f_{\balpha}^i + \ip{f_{\balpha}^i}{\Psi} - |\Ycal| F_{\balpha}^i
- F_{\balpha}^i \left( \ip{f_{\balpha}^i}{\Psi} - \frac{|\Ycal|}{2} F_{\balpha}^i \right)\ .
\end{equation*}
We have thus expressed $\ell(f_{\balpha},i)$ explicity in terms of $\Psi$ and $C$, and computing these quantities will depend on the specific combinatorial structure (cf. Section~\ref{sc:assumptions}).

Let $\circ$ denote the Hadamard product, let $\tr$ denote the trace operator, and let $\diag$ be the operator that maps a square matrix to the column vector corresponding to its diagonal as well as a column vector to the corresponding diagonal matrix. Using the canonical orthonormal basis of $\R^d$, we can write the optimisation problem \eq{eq:texpoptform} as
\begin{equation}\label{eq:finiteopt}
\begin{aligned}
\argmin_{\balpha\in\R^{d\times m}}
 \ & \lambda \tr \balpha K \balpha^\top + \frac{1}{2} \tr K \balpha^\top C \balpha K 
% \\ &
       +  \Psi^\top \balpha K \one     +   \frac{|\Ycal|}{2} \left\| \diag (Y\balpha K) \right\|^2 \\
   &   -  |\Ycal| \tr Y \balpha K     -    \Psi^\top \balpha K \diag (Y\balpha K)\ .
\end{aligned}
\end{equation}
We can use iterative methods like Newton conjugate gradient for training with the gradient
\begin{equation*}\label{eq:gradient}
\begin{aligned}
2 \lambda  \balpha K + C \balpha K^2  +  \Psi \one^\top K  - Y^\top\diag(\Psi^\top \balpha K)K \\
    \quad
-  |\Ycal| Y^\top K  + (|\Ycal| Y^\top - \Psi \one^\top)(\iden \circ Y\balpha K)K
\end{aligned}
\end{equation*}
and the product of the Hessian with vector $v$ 
\begin{equation*}\label{eq:hessvec}
2 \lambda  v K + C v K^2 + |\Ycal| Y^\top(\iden \circ Y v K) K 
%\\ & \quad  
-  \Psi \diag (Y v K) K - Y^\top\diag(\Psi^\top v K)K\ .
\end{equation*}

\subsubsection{Computing $\ell(f_{\balpha},i)$ with Infinite Dimensional Output Embedding}
In the case of infinite dimensional output embeddings, we assume that $\lhull\{k_\Ycal(z,\cdot)\mid z\in\Ycal\}$ has a basis $u_1(\cdot),\ldots u_k(\cdot)$ with $k$ polynomial in the input of our learning problem. Therefore it is sufficient to optimise over $mk$ variables as $\lhull\{k_\Xcal(x_i,\cdot)\otimes u_l(\cdot)\mid i\in\numset{m},l\in\numset{k}\}=\Fcal$ resuting in the following optimisation problem:
\begin{equation*}
\balpha^* =
\argmin\limits_{\balpha\in\R^{m\times k}}
\lambda  \|f_{\balpha}\|^2 + \sum_{i\in\numset{m}} \ell(f_{\balpha},i)
\end{equation*}
with
\begin{equation*}
f_{\balpha}(x,z)=\sum\limits_{i\in\numset{m}, l\in\numset{k}} \balpha_{il} k_{\Xcal}(x_i,x) u_l(z)%\ip{ u_l(\cdot)}{k_{\Ycal}(z,\cdot)}
\ .
\end{equation*}

We now introduce bra-ket notation for convenience and denote $k_\Ycal(z,\cdot)$ by $\ket{z}$ and its dual by $\bra{z}$.  General elements of $\Hcal_\Ycal$ and its dual will be denoted by kets and bras with letters other than $z$ and $y$. Note that hence $\ket{a}\in\Hcal_\Ycal$ does not imply that there exists $z\in\Ycal$ with $\ket{a}=\ket{z}$. In particular, we denote $f_{\balpha}(x_i,\cdot) = \sum_{j\in\numset{m}, l\in\numset{k}} \balpha_{jl} k_{\Xcal}(x_i,x_j) u_l(\cdot)$ by $\ket{f_{\balpha}^i}$. The product $\ket{a}\bra{b}$ is a linear operator $\Hcal_{\Ycal}\to\Hcal_{\Ycal}$, the product $\bra{b}\ket{a}$ is just the inner product $\ip{a}{b}$.
Note that for $\psi:\Ycal\to\R^d$ with \hide{$d<\infty$ and} $\fsip{z}{z'}=k_\Ycal(z,z')$ it holds that $\ket{z}=\psi(z)$ and $\bra{z}=\psi^\top(z)$. 
With $\ket{\Psi}=\sum_{z\in\Ycal}\ket{z}$, $C=\sum_{z\in\Ycal}\ket{z} \bra{z}$, and  $F_{\balpha}^i=\sum_{y\in Y_i} f(x_i,y)$ we obtain 
\begin{equation*}
\sum_{z\in\Ycal} f(x_i,z) = \bra{f_{\balpha}^i} \ket{\Psi}\ ,
\end{equation*}
%\begin{equation*}
%\sum_{z\in\Ycal\setminus Y_i} f(x_i,z) = \sum_{z\in\Ycal} f(x_i,z) -  \sum_{y\in Y_i} f(x_i,y)\ ,
%\end{equation*}
\begin{equation*}
\sum_{z\in\Ycal} f^2(x_i,z) = \bra{f_{\balpha}^i} C \ket{f_{\balpha}^i}\ ,
\end{equation*}
%\begin{equation*}
%\sum_{z\in\Ycal\setminus Y_i} f^2(x_i,z) = \sum_{z\in\Ycal} f^2(x_i,z) -  \sum_{y\in Y_i} f^2(x_i,y)\ ,
%\end{equation*}
and hence
\begin{equation*}
\ell(f_{\balpha},i)
= \frac{1}{2} \bra{f_{\balpha}^i} C \ket{f_{\balpha}^i} + \bra{f_{\balpha}^i} \ket{\Psi} - |\Ycal| F_{\balpha}^i
- F_{\balpha}^i \left( \bra{f_{\balpha}^i} \ket{\Psi} - \frac{|\Ycal|}{2} F_{\balpha}^i \right)\ .
\end{equation*}

\section{Scalability Issues}\label{sc:scale}
The optimisation problem \eq{eq:finiteopt} suffers from scalability issues similar to those that arise in ridge regression. Also, the gradient and Hessian computations involve dense matrix-matrix and matrix-vector computations that may prove to be detrimental for use in large-scale structured prediction problems. We now present a couple of techniques to address these issues. First, we reformulate the problem using a linear scoring function. This is not a serious restriction as we will see in the following section that it is indeed possible to solve an equivalent problem to \eq{eq:finiteopt} using linear models and techniques from low-dimensional embeddings. Second, we show how to solve the problem \eq{eq:finiteopt} using online optimisation and RKHS updates very much similar in spirit to the kernel perceptron \cite{Freund/Schapire/99}.
\subsection{Linear models}
Consider a linear model with scoring function $f(x,y) = \ip{w}{\phi(x,y)}$ where $\phi(x,y) = \psi(y) \otimes x$, and the following problem:
\begin{equation} \label{eqn:linopt}
\argmin\limits_{w \in \R^{nd}} \lambda \|w\|^2 + \sum\limits_{i \in \numset{m}}\ell(f, i) \ ,
\end{equation}
where the loss function is the same as \eq{eq:texploss}. If we let $\psi: \Ycal \to \{0,1\}^d$, we can again express $\ell(f, i)$ using $\Psi$, $C$, and 
\begin{equation*}
\begin{array}{ll}
\sum\limits_{z \in \Ycal} f(x_i, z) = & \ip{w}{\Psi \otimes x_i} \ ,
\end{array}
\end{equation*}
\begin{equation*}
\begin{array}{rl}
\sum\limits_{z \in \Ycal} f^2(x_i, z) = & w^\top [\sum\limits_{z \in \Ycal}\phi(x_i,z)\phi^\top(x_i,z)]w \\
= & w^\top [C \otimes x_ix_i^\top] w \ .
\end{array}
\end{equation*}
Under the assumption that $\Hcal = \Hcal_{\Xcal} \otimes \Hcal_{\Ycal}$, the optimisation problem \eq{eqn:linopt} is equivalent to \eq{eq:finiteopt} with $k_\Xcal(x,x') = \ip{x}{x'}$. Given any arbitrary kernel on the input space $\Xcal$ and a set of training examples, we can extract a corresponding low-dimensional embedding of the inputs (see Appendix~\ref{app:lde} for an overview on kernel functions and low-dimensional mappings) and still be able to solve the problem \eq{eqn:linopt}. The advantage of such a formulation is that it is straightforward to apply online optimisation techniques like stochastic gradient descent and its extensions \cite{Zinkevich03,Shwartz/etal/07,Hazan/etal/07} resulting in scalable algorithms for structured prediction.

\subsection{Online Optimisation}\label{sc:online}
Online methods like stochastic gradient descent (SGD) and stochastic meta descent (SMD) \cite{Schraudolph99} incrementally update their hypothesis using an approximation of the true gradient computed from a single training example (or a set of them). While this approximation leads to slower convergence rates (measured in number of iterations) when compared to batch methods, a common justification to using these methods is that the computational cost of each iteration is low as there is no need to go through the entire data set in order to take a descent step. Consequently, stochastic methods outperform their batch counterparts on redundant, non-stationary data sets \cite{Vishy/etal/06}. The central issue in using stochastic methods is choosing an appropriate step size of the gradient descent, and techniques like stochastic meta descent \cite{Schraudolph99,Vishy/etal/06} have emerged as powerful means to address this issue.

\subsubsection{Stochastic Gradient Descent in Feature Space}
It is rather straightforward to use stochastic gradient descent for linear models with a parameter vector $w$. The update rule at any iteration is $w \gets w - \eta \nabla$, where $\nabla$ and $\eta$ are the instantaneous gradient and step size\footnote{The step size is often chosen to be time-dependent.} respectively. However, for non-linear models such as kernel methods, we have to perform gradient descent in RKHS and update the \emph{dual} parameters. We illustrate this with an example following the online SVM algorithm of \citet{Kivinen/etal/01}. Consider regularised risk minimisation with loss function $\ell : \Xcal \times \Ycal \times \Ycal \to \R$:
\begin{equation*}
\begin{array}{c}
R(f) = \lambda \Omega(f) + \frac{1}{m} \sum\limits_{i=1}^m \ell(x_i,y_i,f(x_i)) \ .
\end{array}
\end{equation*}
The stochastic approximation of the above functional is given as
\begin{equation*}\label{eq:online}
\begin{array}{c}
R_{\text{stoc}}(f,t) = \lambda \Omega(f) + \ell(x_t,y_t,f(x_t)) \ .
\end{array}
\end{equation*}
The gradient of $R_{\text{stoc}}(f,t)$ w.r.t. $f$ is 
\begin{equation*}
\begin{aligned}
\nabla_f R_{\text{stoc}}(f,t) & = \lambda \nabla_f\Omega(f) + \nabla_f\ell(x_t,y_t,f(x_t))\\
& = \lambda \nabla_f\Omega(f) + \ell'(x_t,y_t,f(x_t))k_{\Xcal}(x_t,\cdot) \ .
\end{aligned}
\end{equation*}
The second summand follows by using the reproducing property of $\Hcal_{\Xcal}$ to compute the derivate of $f(x)$, i.e., \mbox{$\nabla_f \ip{f(\cdot)}{k_{\Xcal}(x,\cdot)}=k_{\Xcal}(x,\cdot)$}, and therefore for the loss function which is differentiable in its third argument we obtain $\nabla_f \ell(x,y,f(x)) = \ell'(x,y,f(x))k_{\Xcal}(x,\cdot)$. The update rule is then given as \mbox{$f \gets f - \eta \nabla R_{\text{stoc}}(f,t)$}. For the commonly used regulariser $\Omega(f)=\frac{1}{2}\|f\|^2$, we get
\begin{equation*}
\begin{array}{rl}
f & \gets f - \eta_t \nabla_t (\lambda f + \ell'(x_t,y_t,f(x_t))k_{\Xcal}(x_t,\cdot)) \\
& = (1-\lambda \eta_t)f - \eta_t \ell'(x_t,y_t,f(x_t))k_{\Xcal}(x_t,\cdot) \ .
\end{array}
\end{equation*}
Expressing $f$ as a kernel expansion $f(\cdot) = \sum_i c_i k_{\Xcal}(x_i,\cdot)$, where the expansion is over the examples seen until the current iteration, we get
\begin{equation*}
\begin{array}{rll}
c_t & \gets (1-\lambda \eta)c_t - \eta_t \ell'(x_t,y_t,f(x_t)) &\\
& = \eta \ell'(x_t,y_t,f(x_t)) & \text{for } c_t = 0 \\
c_i & = (1-\lambda \eta)c_i & \text{for } i \neq t \ . \\
\end{array}
\end{equation*}

We are now ready to derive SGD updates for the optimisation problem \eq{eq:finiteopt}.

\subsubsection{Online Structured Ridge Regression}
At iteration t, let $K_t = K_{\numset{t}\numset{t}} \in \R^{t \times t}$, $k_t = (K_{\numset{t}\numset{t}})_{.t}$, $y_t = Y_{t.}$, and let $\balpha_t \in \R^{d \times t}$ be the parameter matrix. In the following, we omit the subscript $t$. The instantaneous objective of the optimisation problem \eq{eq:finiteopt} can be written as
\hide{
Let $I_t \in \R^{n \times n}$ such that $(I_t)_{ij}=1$ if $i,j \in \numset{t}$ or $0$ otherwise. At iteration $t$, define $K_t = K \circ I_t$, $k_t = (K_t)_{.t}$, $y_t = Y_{t.}$, and let $\balpha_t$ be the parameter vector. In the following, we omit the subscript $t$. The instantaneous objective of \eq{eq:finiteopt} can be written as
} \begin{equation}\label{eq:insfiniteopt}
\begin{aligned}
\argmin_{\balpha\in \R^{d \times t}} \lambda \tr \balpha K \balpha^\top + \frac{1}{2} k^\top \balpha^\top C \balpha k  +  \Psi^\top \balpha k \\
\quad\quad
  +   \frac{|\Ycal|}{2} ( y \balpha k )^2 -  |\Ycal| y \balpha k    -   \Psi^\top \balpha k y \balpha k
\end{aligned}
\end{equation}
with gradient $\nabla$,
\begin{equation}\label{eq:insgradient}
\begin{aligned}
 2 \lambda  \balpha K + C \balpha k k^\top  +  \Psi k^\top  + |\Ycal| y \balpha k y^\top k^\top \\
\quad
 -  |\Ycal| y^\top k^\top   - \Psi k^\top y \balpha k - \Psi^\top \balpha k y^\top k^\top \ .
\end{aligned}
\end{equation}
product of Hessian $\nabla^2$ with vector $v$,
\begin{equation}\label{eq:inshessvec}
  2 \lambda  v K + C v k k^\top +  |\Ycal| y v k y^\top k^\top  - \Psi k^\top y v k - \Psi^\top v k y^\top k^\top \ .
\end{equation}

With step size $\eta$, we obtain the following update rule: $\balpha \gets \balpha - \eta \nabla$. We now discuss a couple of implementation aspects pertaining to the optimisation problem \eq{eq:insfiniteopt}.

\subsubsection{Truncating Kernel Expansion Coefficients}
As the function \eq{eq:insfiniteopt} is minimised in RKHS, the parameter matrix $\balpha$ grows incrementally with time by adding a single row in every iteration. In order to speed up computations, we truncate all parameters that were updated before time $\tau$. This is justified for regularised risk minimisation problems because at every iteration, $\balpha_i.$ with $i < t$ is shrunk by a factor $(1- \lambda \eta)$ and therefore the contribution of old parameters in the computation of the kernel expansion \hide{eq:form} decreases with time \cite{Kivinen/etal/01,Vishy/etal/06}. We use this simple technique to speed up computations in our experiments. It is also possible to apply other techniques (see, for example, \cite{Dekel/etal/05}), to discard less important coefficients. Note that the learning rate is set to ensure $0 \leq 1- \lambda \eta < 1$.

\subsubsection{Step Size Adaptation}
The step size plays an important role in the convergence of stochastic approximation techniques and has been the focus of recent research \cite{Vishy/etal/06}. We set the step size $\eta_t = \frac{p}{\lambda t}$ where $p$ is a parameter that has to be fine tuned to obtain good performance. A better way to set the step size would be to consider SMD updates \cite{Schraudolph99,Vishy/etal/06}. \hide{or perform SGD followed by the greedy projection method where the step size is cleary defined with theoretical guarantees}

\section{Approximate Inference} \label{ch_:inference}
Thus far we have described a training algorithm for structured prediction, but have not discussed how to predict structures using the learned model. We now design (inference) algorithms for predicting combinatorial structures. As exact inference is in general hard (cf. Chapter \ref{ch:complexity}), we have to resort to approximations. We therefore design approximation algorithms using the notion of z-approximation \cite{Hassin/Khuller/01,AusMar80} with provable guarantees.

\subsection{Approximate Decoding}\label{su:dec}
In order to construct (predict) structures for a given input using the model described in Section~\ref{sc:training}, we have to solve the decoding problem 
\begin{equation*}
\argmax_{y\in\Ycal} f(x,y) \ .
\end{equation*}
In cases where \emph{exact decoding} is not possible, we resort to approximate decoding methods and aim at finding $\hat y$ such that  
\begin{equation*}
f(x,\hat y)\approx \max_{y\in\Ycal} f(x,y) \ .
\end{equation*}
\hide{
The general idea is to first construct a complete directed graph and assign a weight to each edge using the trained model. Given a directed edge $(u,v)$, its weight $w(u,v)$ is given by $f(x,(u,v)) = \ip{f_{\balpha}^i}{\phi((u,v))}$. The next step is to apply an appropriate approximation algorithm on this graph. This idea was first introduced by \citet{McDonald/etal/05} to predict maximum spanning trees for dependency parsing in natural language processing applications. We illustrate with an example that this idea does not work for the kind of combinatorial structures considered in this work and thereby motivate the use of z-approximation algorithms.

Consider the prediction of posets. At first glance, it seems plausible to use existing approximation algorithms for the maximum acyclic subgraph problem \cite{Hassin/Rubinstein/94} or the complementary minimum feedback arc set problem \cite{Even/etal/95}. We now give a counter-example where this is not possible. Consider the digraph $V = \numset{4}, E = \{(1,2),(2,3),(3,4),(1,3),(2,4)\}$ with $w((2,4))=-8$ and $w(e)=1$ for all $e \in E \setminus \{(2,4)\}$. The maximum acyclic subgraph is $\{(1,2),(2,3),(3,4),(1,3)\}$ with total weight $4$. However, if we also consider edges included by transitivity (as needed for the decoding problem for posets), the weight reduces to $-4$. On the other hand, the subgraph $\{(3,4),(1,3)\}$ has weight only $2$, but if we also consider the edges included by transitivity, the weight is still 2!. The problem is due to negative weights on edges which makes these algorithms inapplicable in our setting because we cannot guarantee that the weights returned by our model are non-negative.

As another example, consider the min-cut problem on a graph with non-negative edge weights. Now, if we change the sign of every edge weight, we see that one has to solve the max-cut problem which is a hard problem. Changing the sign of the weights has dramatically affected the optimisation problem (min-cut to max-cut)! In the following section, we describe the notion of $z$-measure for approximation algorithms and prove approximation guarantees for the decoding problem of various classes of combinatorial structures.
}
\subsubsection{z-approximation}
$z$-approximation algorithms are particularly suitable for optimisation problems involving negative weights. The reader is referred to \citet{Hassin/Khuller/01} for examples.
\hide{
For prediction, we resort to approximate decoding methods for $\hat y$ such that  $f(x,\hat y)\approx \max_{y\in\Ycal} f(x,y)$. For output kernels $k_\Ycal(y,y')=\fsip{y}{y'}$ we have $f(x,y)=\ip{w_x}{\phi(y)}$ for some $w_x\in\lhull\{\phi(y)\mid y\in \Ycal\}$ determined\hide{ for finite dimensional output spaces} by the above described optimisation \eq{eq:finiteopt} and the representation of the function \eq{eq:form} as $w_x=\balpha \left(k_{\Xcal}(x_1,x)\ k_{\Xcal}(x_2,x)\ \cdots\ k_{\Xcal}(x_n,x)\right)^\top$. 

We now give two cases that are as simple as they are general and in which a $z$-approximation for decoding can be guaranteed.}$z$-approximation was proposed by \citet{Zemel/81} instead of the more common ``percentage-error'' $f(x,\hat y)/\max_{y\in\Ycal}f(x,y)$. $\hat y$ has $z$-approximation factor $\nu$ for a maximisation problem if
\begin{equation}\label{eq:z}
f(x,\hat y) \geq (1-\nu) \max_{y\in\Ycal} f(x,y) + \nu \min_{y\in\Ycal}f(x,y)\ .
\end{equation}
An algorithm with $\nu=0$ is optimal whereas an algorithm with $\nu=1$ is trivial.
$z$-approximation has several advantages \cite{Hassin/Khuller/01,AusMar80} including ($i$) $\max_{y\in\Ycal} f(x,y)$ has the same $z$-approximation as $\max_{y\in\Ycal} (f(x,y) + c)$ where $c$ is a constant; ($ii$) $\max_{y\in\Ycal} f(x,y)$ has the same $z$-approximation as $\min_{y\in\Ycal} ( - f(x,y))$ ($z$-approximation for minimisation is defined by exchanging $\min$ and $\max$ in \eq{eq:z}); and ($iii$) the $z$-approximation factor is invariant under replacing binary variables by their complement.

\subsubsection{Decoding Sibling Systems}\label{su:dss}
A sibling system is an output space $\Ycal$ with a sibling function $r:\Ycal\to\Ycal$ and an output map $\psi$ such that $\forall z \in \Ycal: \psi(z) + \psi(r(z)) = \mathbf{c}$ as well as $\forall z \in \Ycal: \ip{\mathbf{c}}{\psi(z)}=0$ with fixed $\mathbf{c}$. In other words, it is an output space in which each structure can be complemented by its sibling. 
\begin{proposition} \label{pr:z-sibling}
There is a $1/2$-factor $z$-approximation algorithm for decoding sibling systems.
\end{proposition}
\begin{proof}
Choose some $y\in\Ycal$ at random. If $f(x,y)\geq 0$, then $\hat y = y$; otherwise $\hat y = r(y)$. This completes the description of the algorithm.

We know that $f(x,y)=\ip{w_x}{\psi(y)}$ for some $w_x \in \R^d$. Now $\forall y \in\Ycal: f(x,y) + f(x,r(y)) = \ip{w_x}{\psi(z)} + \ip{w_x}{\psi(r(z))} = \ip{w_x}{\mathbf{c}} = 0$ and as $\min_{y\in\Ycal}f(x,y) = - \max_{y\in\Ycal} f(x,y)$, we have
\begin{equation*} 
f(x,\hat y)\geq 0 = \frac{1}{2}\max_{y\in\Ycal} f(x,y) + \frac{1}{2}\min_{y\in\Ycal}f(x,y)
\end{equation*}
thereby satisfying \eq{eq:z} with $\nu=1/2$.
\end{proof}

Notice that if $r$ is bijective and $\mathbf{c}=\zero$, then also $\Psi=\zero$, thus significantly simplifying the optimisation problem \eq{eq:finiteopt}. For $\Ycal=2^\Sigma$ an example of a bijective sibling function is $r:y\mapsto\Sigma\setminus y$.

\subsubsection{Decoding Independence Systems}\label{su:dic}
An independence system $(\Sigma,\Ycal)$ is an output space $\Ycal$ such that $\forall y \in \Ycal: z\subset y \dann z\in\Ycal$ and membership in $\Ycal$ can be tested in polynomial time. \hide{If $\Ycal$ is defined by a property $\pi:\Ycal\to\Omega$, this property is called hereditary.}Consider an output map $\psi:\Ycal\to\R^{|\Sigma|}$ with $\psi_u(z)=\sqrt{\mu(u)}$ if $u\in z$ and $\psi_u(z)=0$ otherwise, for some measure $\mu$. Here we find a polynomial time $1-(\log_2|\Sigma|)/|\Sigma|$ factor $z$-approximation following \cite{Halldorsson/99}.
\begin{proposition} \label{pr:z-indy}
There is a $1 - \log_2|\Sigma|/|\Sigma|$ factor $z$-approximation algorithm for decoding independence systems.
\end{proposition}
\begin{proof}
Partition $\Sigma$ into $\lceil|\Sigma|/\log_2|\Sigma|\rceil$ many subsets $E_i$ of size $|E_i|\leq\lceil\log_2|\Sigma|\rceil$. Choose $\hat y = \argmax_{y\in \Scal} f(x,y)$ where $\Scal = \{\argmax_{y\subseteq E_i\cap\Ycal} f(x,y) \mid i \in \numset{\lceil|\Sigma|/\log_2|\Sigma|\rceil}\}$. We again choose $w_x$ such that $f(x,y)=\ip{w_x}{\psi(y)}$. Now we can find $\hat y = \argmax_{y\in \Scal} \ip{w_x}{\psi(y_i)}$ in polynomial time by exhaustively testing $2^{|E_i|}\leq 2^{\lceil\log_2|\Sigma|\rceil}\leq|\Sigma|+1$ alternatives in each of the $\lceil|\Sigma|/\log_2|\Sigma|\rceil\leq|\Sigma|$ subsets. This completes the description of the algorithm.

For the $z$-approximation, suppose there was $y'\in\Ycal$ with $f(x,y') > f(x,\hat y) |\Sigma|/\log_2 |\Sigma|$. As $\{E_i\}$ partitions $\Sigma$, we have 
\begin{equation*}
\begin{aligned}
f(x,y') & = \ip{w_x}{\psi(y')} \\
& = \sum_i \ip{w_x}{\psi(y'\cap E_i)} \\
& \leq  \frac{|\Sigma|}{\log_2 |\Sigma|} \max_i \ip{w_x}{\psi(y'\cap E_i)} \\
& \leq \frac{|\Sigma|}{\log_2 |\Sigma|} f(x,\hat y) 
\end{aligned}
\end{equation*}
contradicting the assumption. Together with $\min_{y\in\Ycal} f(x,y) \leq f(x,\emptyset) =0$ this proves the stated approximation guarantee.
\end{proof}

\subsection{Approximate Enumeration}\label{su:sampling}
If the \emph{non-optimality} decision problem $\exists y\in\Ycal: f(x,y) > \theta$ is NP-hard then there is no algorithm for enumerating any set $\Scal_x'\supseteq \Scal_x^* = \left\{\hat y \in \Ycal \mid f(x,\hat y) > \theta \right\}$ of cardinality polynomial in $|\Scal_x^*|$ in output polynomial time (unless P=NP).
To see this, suppose there was an output polynomial algorithm for listing $\Scal_x'$. This algorithm can be used to obtain an output polynomial algorithm for listing $\Scal_x^*$ with additional complexity only for testing $s\in\Scal_x^*$ for all $s\in\Scal_x^*$.
Now if the algorithm terminates in input polynomial time then it is sufficient to check the cardinality of $\Scal_x^*$ to decide non-optimality. If on the other hand the algorithm does not terminate in input polynomial time, then $\Scal_x^*$ can not be empty.

Hence, we consider enumerating approximate solutions, i.e., we want to list
\begin{equation}\label{eq:zset}
\left\{\hat y \in \Ycal \mid f(x,\hat y) \geq (1-\nu) \max_{y\in\Ycal} f(x,y) + \nu \min_{y\in\Ycal}f(x,y)\right\}\ .
\end{equation}

If $\Ycal$ is such that listing the whole of $\Ycal$ is hard, this precludes a general algorithm for the trivial case ($\nu=1$) and makes it rather unlikely that an algorithm for $\nu>0$ exists. We now assume that we know how to list (sample uniformly from) $\Ycal$ and that the situation is similar to sibling systems: We have a bijective function $r:\Ycal\to\Ycal$ and a map $\psi$ such that $\forall z \in \Ycal: \psi(z) + \psi(r(z)) = \mathbf{c}$ as well as $\forall z \in \Ycal: \ip{\mathbf{c}}{\psi(z)}=0$ for fixed $\mathbf{c}$. Then we can list the set~\eq{eq:zset} in incremental polynomial time by a simple algorithm that internally lists $y$ from $\Ycal$ but instead of outputting $y$ directly, it only outputs $y$ if $f(x,y)>f(x,r(y))$ and otherwise outputs $r(y)$.\hide{ Note that the case $f(x,y)=f(x,r(y))$ does not deserve special attention as the subroutine for listing $\Ycal$ will also list $r(y)$.}

\section{Empirical Results}\label{sc:exp}
In all experiments, we fixed the regularisation parameter of our algorithm $\lambda= |\Ycal| \cdot \sum_{i \in \numset{m}}|Y_i|/m$.
\subsubsection{Multi-label Classification}
We compared the performance of our algorithm with the kernel method proposed by \citet{Elisseeff/Weston/01} that directly minimises the ranking loss (number of pairwise disagreements). We followed the same experimental set up (dataset and kernels) as described in \cite{Elisseeff/Weston/01}. We used the \emph{Yeast} dataset consisting of 1500 (training) and 917 (test) genes with 14 labels, and trained our algorithm with polynomial kernel of varying degree (2-9). Figure~\ref{fig:results_mlp} shows the results for Hamming loss and ranking loss. \hide{Our algorithm consistently outperforms the kernel method for multi-label classification.}
\begin{figure}[tb]
\begin{center}
\epsfig{file=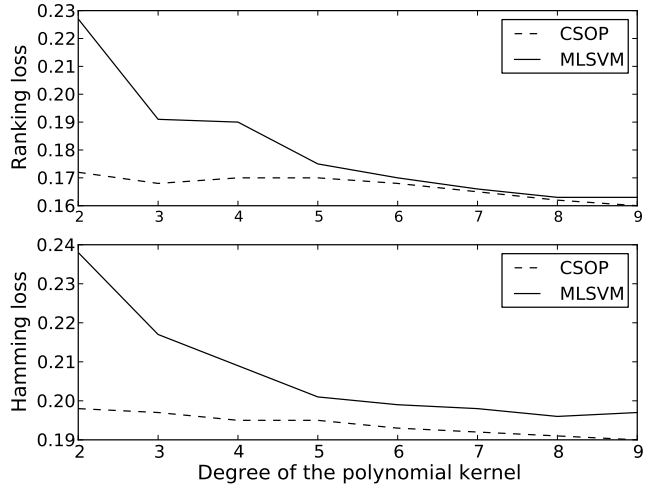,width=.75\textwidth}
\caption{Comparison of multi-label SVM (MLSVM) and our algorithm (CSOP) on multi-label classification.}
\label{fig:results_mlp}
\end{center}
\vspace{-.5cm}
\end{figure}

\subsubsection{Hierarchical Classification}
We trained our algorithm on the WIPO-alpha patent dataset\footnote{Available at one of the authors \cite{Rousu/etal/06} webpage: http://users.ecs.soton.ac.uk/cjs/downloads/} consisting of 1352 training and 358 test documents. The number of nodes in the hierarchy is 188 with maximum depth of 3. Each document belongs to exactly one leaf category and hence contains no multiple paths. The performance of several algorithms (results are taken from \cite{Rousu/etal/05}) is shown in Table~\ref{tbl:hierarchical}.
\begin{table}[t]
\caption{Comparison of various algorithms on hierarchical classification.}
\begin{center}
\begin{tabular}{|c||c|c|c|}
\hline
Algorithm & $\ell_{0/1}$ & $\ell_\Delta$ & $\ell_H$ \\
\hline
SVM&87.2&1.84&0.053 \\
H-SVM&76.2&1.74&0.051\\
H-RLS&72.1&1.69&0.050 \\
H-$\mathrm{M}^3-\ell_\Delta$&70.9&1.67&0.050\\
H-$\mathrm{M}^3-\ell_{\tilde{H}}$&65.0&1.73&0.048\\
CSOP&51.1&1.84&0.046\\
\hline
\end{tabular}\label{tbl:hierarchical}
\end{center}
\end{table}
$\ell_{0/1}$ denotes the zero-one loss in percentage, $\ell_\Delta$ is the average Hamming loss per instance, and $\ell_H$ is the hierarchical loss (average per instance). The hierarchical loss is based on the intuition that if a mistake is made at node $i$, then further mistakes made in the subtree rooted at $i$ are unimportant \cite{Cesa-Bianchi/etal/06}. Formally, for any pair of hierarchical labels $z$ and $y$,
\begin{equation*}
\ell_H(z,y) = \sum\limits_{i=1}^{|\Sigma|} I(\psi_i(z) \neq \psi_i(y) \land \psi_j(z) = \psi_j(y), ~ j \in ANC(i)) \ ,
\end{equation*}
where $ANC(i)$ is the set of ancestors of $i$, and $I(p)$ is 1 if $p$ is true and $0$ otherwise. In this way the hierarchy or taxonomy of the problem domain is taking into account. SVM denotes an SVM trained for each microlabel independently, H-SVM denotes an SVM trained for each microlabel independently and using only those samples for which the ancestor labels are positive, H-RLS is the hierarchical least squares algorithm described in \cite{Cesa-Bianchi/etal/06} and H-$\mathrm{M}^3$ is the kernel-based algorithm proposed by \citet{Rousu/etal/06} which uses the maximum margin Markov network framework. The two different versions of this algorithm correspond to using the Hamming loss and the hierarcical loss during training. While the performance on the Hamming loss was comparable to the baseline SVM, our algorithm resulted in best performance on the $\ell_{0/1}$ loss and the hierarchical loss. \hide{We note that on this particular dataset there is no positive correlation between the Hamming loss and the other two losses. Our algorithm was able to output the true label for around half of the test instances. Among the rest, it was making a lot of mistakes in the subtrees of incorrectly labeled nodes that are anyway considered to be unimportant and therefore not penalised by the hierarchical loss.}

\subsubsection{Dicycle policy estimation}
We experimented with an artificial setting due to lack of real world data sets on dicycle prediction\footnote{Note that we considered directed cyclic permutations as opposed to undirected ones described in Section~\ref{sc:hard} due to the fact that $\Psi^\mathrm{cyc} = \zero$ for dicycles, thereby reducing the number of computations involved in optimising \eq{eq:finiteopt}. See Appendix~\ref{app:dicycle} for more details on counting dicyclic permutations.}. We simulate the problem of predicting the cyclic tour of different people. We assume that there is a hidden policy for each person and he/she takes the route that (approximately) maximises the reward of the route. In the setting of dicycle prediction, the learned function is linear in the output space ($f(x_i,y)=\ip{f_{\balpha}^i}{\psi(y)}$) and for testing we can check how well the estimated policy $f_{\balpha}^i$ approximates the hidden policy in the test ($i\in\numset{m+1, m'}$) set. The data is constructed as follows: 
($i$) generate $n$ matrices $A^{(i)}\in\R^{\Sigma\times \Sigma}$ uniformly at random with entries in the interval $[-1,1]$ and $A^{(i)}_{uv} = -A^{(i)}_{vu}$; ($ii$) generate $(m+m')n$ random numbers uniformly between $0$ and $1$ to form the inputs $x_i\in\R^n$; ($iii$) create the output structures $y_i\approx\argmax_{y\in\Ycal} \sum_{(u,v)\in y, j\in\numset{n}} x_{ij} A^{(j)}_{uv}$ for training, that is $i\in\numset{n}$.
On the test set, we evaluated our algorithm by cosine similarity of the learned policy and the true policy: 
\begin{equation*}
\begin{array}{c}
\sum\limits_{i\in\numset{m,m+m'}} \ip{f_{\balpha}^i}{\sum\limits_{j\in\numset{n}} x'_{ij}  A^{(j)}} \cdot \left\|f_{\balpha}^i\right\|^{-1} \cdot \left\|\sum\limits_{j\in\numset{M}} x'_{ij}  A^{(j)} \right\|^{-1}
\end{array}
\end{equation*}

Figure~\ref{fig:results_dcp} shows a plot of the cosine measure on an experiment ($m'=500, n=15, |\Sigma|=10$) for varying number of training instances.
\begin{figure}[tb]
\begin{center}
\epsfig{file=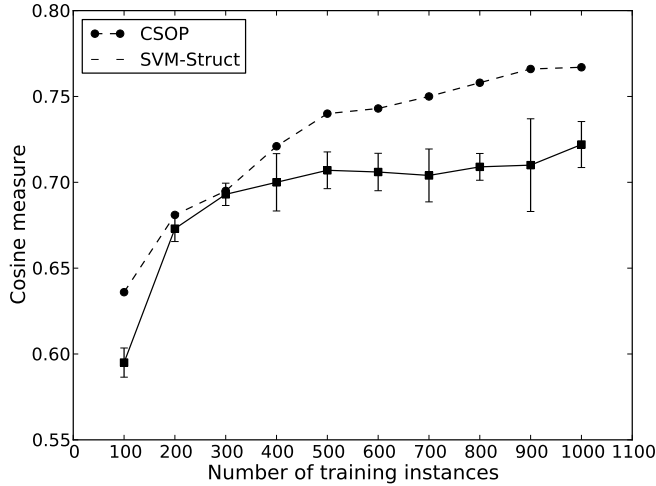,width=.75\textwidth}
\caption{Comparison of structured SVM and CSOP on dicycle policy estimation.}
\label{fig:results_dcp}
\end{center}
\vspace{-.5cm}
\end{figure}
\hide{We generated 5 cycles per training instance, i.e. $|Y_i|=5$.} As expected, we see that our algorithm is able to estimate the true policy with increasing accuracy as the number of training instances increases. The plot also shows the performance of structured SVM using approximate decoding during training. Approximate decoding is performed by randomly sampling a couple of cyclic permutations and using the best scoring one (the number of repetitions used in our experiments was $25$). The results were unstable due to approximate decoding and the results shown on the plot are averages from 5 trials. For structured SVM, we experimented with several values of the regularisation parameter and report the best results obtained on the test set --- though this procedure gives an advantage to structured SVM, our algorithm still resulted in better performance.

\subsubsection{Stochastic Gradient Descent}
We performed a couple of experiments using artificially generated data to compare the performances of online and batch learning. In the first experiment, we trained a simple multi-label classification model to learn the identity function $f: \{0,1\}^d \to \{0,1\}^d$. The goal was to compare batch and online learning, using Newton conjugate gradient (NCG) and stochastic gradient descent (SGD) respectively, in terms of the final objective value of the optimisation problem \eq{eq:finiteopt} and training time. We also studied the effects of the truncation parameter $\tau$ on speed and final objective. We trained SGD on a single pass of the data set. Figures~\ref{fig:online_srr11} and \ref{fig:online_srr12} summarises the results for multi-label classification on an artificial data set with 5 features and labels. 
\begin{figure}[t!]
\begin{center}
\epsfig{file=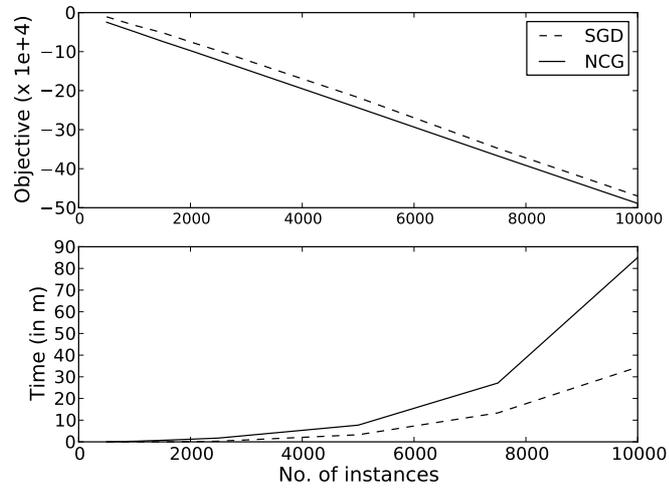,width=.75\textwidth}
\caption{Comparison of SGD and NCG training on multi-label classification.}
\label{fig:online_srr11}
\end{center}
\vspace{-.5cm}
\end{figure}
We set the truncation parameter $\tau$ to $0.15 \times m$, where $m$ is the number of training instances. We see that the final solution of SGD is comparable to that of NCG, and the speed up achieved by SGD is apparent for large data sets. The effect of $\tau$ on training time and objective is shown in Figure~\ref{fig:online_srr12}. The training time of SGD increases with $\tau$ and attains the training time of NCG at around $19\%$ of $m$. Beyond this value, we found that SGD was taking longer time than NCG. This underlines the effect of $\tau$ when performing SGD updates in RKHS.
\begin{figure}[t!]
\begin{center}
\epsfig{file=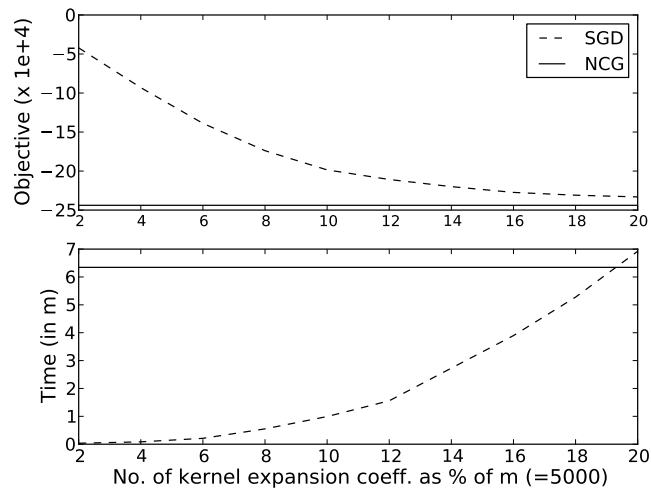,width=.75\textwidth}
\caption{Effect of truncation parameter $\tau$ on training time and final objective value on multi-label classification with SGD and NCG.}
\label{fig:online_srr12}
\end{center}
\vspace{-.5cm}
\end{figure}

In our second experiment, we considered dicycle policy estimation as before. We trained SGD and NCG on data sets of varying size from $100$ to $5000$ with $n=15$ and $\Sigma=10$. We fixed $\tau$ to $500$ kernel expansion coefficients. Figure~\ref{fig:online_srr2} shows a plot of final objective versus training time of SGD and NCG on the different data sets.
\begin{figure}[t!]
\begin{center}
\epsfig{file=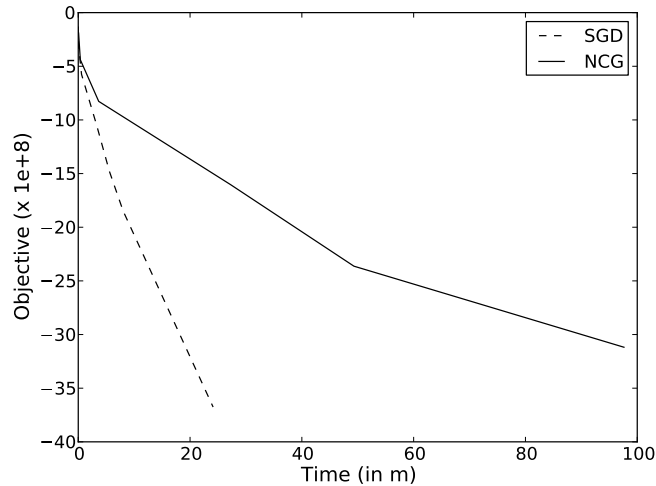,width=.75\textwidth}
\caption{Comparison of SGD and NCG training on dicycle policy estimation.}
\label{fig:online_srr2}
\end{center}
\vspace{-.5cm}
\end{figure}
The plot shows that NCG takes a much longer time to attain the same final objective value as SGD. Note that as we perform single pass training, with fixed amounts of training instances NCG attains a smaller value of the objective function than SGD. However, as SGD can deal with much more training instances in the same time, after a fixed amount of time, SGD attains a smaller value of the objective function than NCG.

\section{Summary}
We presented a learning algorithm for predicting combinatorial structures under the counting assumption introduced in Chapter~\ref{ch:complexity} . Our approach subsumes several machine learning problems including multi-class, multi-label and hierarchical classification, and can be used for training complex combinatorial structures. As for most combinatorial structures considered in this paper the inference problem is hard, an important aspect of our approach is that it obviates the need to use inference algorithms, be it exact or approximate, for training. 

We have also seen how to train a linear model using the counting assumption. Under some reasonable assumptions, a non-linear model can be approximated using a linear model using techniques from metric embedding theory. Furthermore, we addressed the scalability issues of our approach by presenting an online learning algorithm using stochastic gradient descent that can update model parameters (kernel expansion coefficients) in RKHS.

For prediction, inference can naturally not be avoided. Therefore, we have to rely on approximation algorithms described in Section~\ref{ch_:inference}. We note that it is non-trivial to design approximation algorithms for the decoding problem of the combinatorial structures considered in this work. Indeed, there are hardness of approximation results for the maximum acyclic subgraph problem \cite{Guruswami/etal/08} and the problem of finding longest directed cycles \cite{Bjorklund/etal/04}.

While we have seen how to minimise a squared loss function, it would be interesting to train a probabilistic model by minimising the negative log-likelihood $\grave{a}$~la conditional random fields (cf. Section~\ref{sc:algorithms}). This will be the focus of the next chapter.

\clearemptydoublepage
\chapter{Probabilistic Structured Prediction}\label{ch:psp}
Maximum a posteriori (MAP) estimation with exponential family models is a fundamental statistical technique for designing probabilistic classifiers (cf. logistic regression). In this chapter, we consider MAP estimators for structured prediction using the sampling assumption introduced in Chapter~\ref{ch:complexity}, i.e., we concentrate on the case that efficient algorithms for uniform sampling from the output space exist. We show that under this assumption (i) exact computation of the partition function remains a hard problem, and (ii) the partition function and the gradient of the log partition function can be approximated efficiently. The main result of this chapter is an approximation scheme for the partition function based on Markov chain Monte Carlo theory. We also design a Markov chain that can be used to sample combinatorial structures from exponential family distributions given that there exists an exact uniform sampler, and also perform a non-asymptotic analysis of its mixing time.

\section{Probabilistic Models and Exponential Families}
Let $\Xcal \times \Ycal$ be the domain of observations and labels, and \mbox{$X=(x_1,\dots,x_m) \in \Xcal^m$}, \mbox{$Y=(y_1,\dots,y_m) \in \Ycal^m$} be the set of observations. Our goal is to estimate $y \mid x$ using exponential families via
\begin{equation*}
p(y \mid x,w) = \exp(\ip{\phi(x,y)}{w} - \ln Z(w \mid x)) \ ,
\end{equation*}
where $\phi(x,y)$ are the joint sufficient statistics of $x$ and $y$, and $Z(w \mid x)= \sum_{y \in \Ycal} \exp(\ip{\phi(x,y)}{w})$ is the partition function. We perform MAP parameter estimation by imposing a Gaussian prior on $w$. This leads to optimising the negative joint likelihood in $w$ and $Y$:
\begin{equation} \label{eqn:opt}
\begin{aligned}
\hat{w} & = \argmin\limits_{w} \left[- \ln p(w,Y \mid X)\right] \\
& = \argmin\limits_{w} \left[\lambda \|w\|^2 + \frac{1}{m} \sum\limits_{i=1}^m [\ln Z(w \mid x_i) - \ip{\phi(x_i,y_i)}{w}] \right] \ ,
\end{aligned}
\end{equation}
where $\lambda > 0$ is the regularisation parameter. We assume that the $\ell_2$ norm of the sufficient statistics and the parameters are bounded, i.e., $\|\phi(x,y)\| \leq R$ and $\|w\| \leq B$, where $R$ and $B$ are constants. \hide{This is not a serious restriction as $B$ can be bounded from above as is proved in Appendix~\ref{sec:norm_bound}.}Note that it is possible to upper bound the norm of the parameter vector $w$ as shown below.
\begin{proposition}\label{prop:norm_param}
The norm of the optimal parameter vector $\hat{w}$ is bounded from above as follows:
\begin{equation*}
\|\hat{w}\| \leq \sqrt{\frac{\ln |\Ycal|}{\lambda}} \ .
\end{equation*}
\end{proposition}
\begin{proof}
Consider any $(x,y) \in \Xcal \times \Ycal$. Denote by $\ell(w,x,y)$ the loss function, where $\ell(w,x,y) = \ln Z(w \mid x) - \ip{\phi(x,y)}{w} \geq 0$, and note that $\ell(0,x,y) = \ln |\Ycal|$. Let $F(w) = - \ln p(w,Y\mid X)$. The true regularised risk w.r.t. $\hat{w}$ and an underlying joint distribution $D$ on $\Xcal \times \Ycal$ is
\begin{equation*}
E_{(x,y)\sim D}[\ell(\hat{w},x,y)] + \lambda \|\hat{w}\|^2 \leq F(0) = \ln |\Ycal| \ .
\end{equation*}
This implies that the optimal solution $\hat{w}$ of the above optimisation problem lies in the set \mbox{$\{w : \|w\| \leq \sqrt{\ln |\Ycal|/\lambda}\}$}.
\end{proof}

The difficulty in solving the optimisation problem \eq{eqn:opt} lies in the computation of the partition function. The optimisation is typically performed using gradient descent techniques and advancements thereof. We therefore also need to compute the gradient of the log partition function, which is the first order moment of the sufficient statistics, i.e., $\nabla_{w} \ln Z(w\mid x)=\E_{y \thicksim p(y\mid x,w)} [\phi(x,y)]$. 

Computing the log partition function and its gradient are in general NP-hard. In Section~\ref{sec:partition}, we will show that computing the partition function still remains NP-hard given a uniform sampler for $\Ycal$. We therefore need to resort to approximation techniques to compute these quantities. Unfortunately, application of concentration inequalities do not yield approximation guarantees with polynomial sample size. We present Markov chain Monte Carlo (MCMC) based approximations for computing the partition function and the gradient of the log partition function with provable guarantees. There has been a lot of work in applying Monte Carlo algorithms using Markov chain simulations to solve \mbox{$\#$P-complete} counting and NP-hard optimisation problems. Recent developments include a set of mathematical tools for analysing the rates of convergence of Markov chains to equilibrium (see \cite{Randall/03,Sinclair/Jerrum/96} for surveys). To the best of our knowledge, these tools have not been applied in the design and analysis of structured prediction problems, but have been referred to as an important research frontier \cite{Andrieu/etal/03} for MCMC based machine learning problems in general.

\section{Hardness of Computing the Partition Function} \label{sec:partition}
We begin with a hardness result for computing the partition function. Consider the following problem:
\begin{definition}
\textsc{Partition}: For a class of output structures $\Ycal$ over an alphabet $\Sigma$, an input structure $x\in\Xcal$, a polynomial time computable map $\psi:\Ycal\to\R^d$, and a parameter $w$, compute the partition function $Z(w)$.
\end{definition}
We now show that no algorithm can efficiently solve \textsc{Partition} on the class of problems for which an efficient uniform sampling algorithm exists. To show this, we suppose such an algorithm existed, consider a particular class of structures, and show that the algorithm could then be used to solve an NP-hard decision problem.
We use that
(a) cyclic permutations of subsets of the alphabet $\Sigma$ can be sampled uniformly at random in time polynomial in $|\Sigma|$; and
(b) there is no efficient algorithm for \textsc{Partition} for the set of cyclic permutations of subsets of the alphabet $\Sigma$ with $\psi_{uv}(y) = 1$ if $\{u,v\}\in y$ and $0$ otherwise.
Here (a) follows from Sattolo's algorithm \cite{Sattolo86} to generate a random cyclic permutation. To prove (b), we show that by applying such an algorithm to a multiple of the adjacency matrix of an arbitrary graph and comparing the result with $|\Sigma|^3$ we could decide if the graph has a Hamiltonian cycle or not.

\begin{theorem}
Unless P=NP, there is no efficient algorithm for \textsc{Partition} on the class of problems for which we can efficiently sample output structures uniformly at random.
\end{theorem}
\begin{proof}
Consider the output space of undirected cycles over a fixed set of vertices $\Sigma$, i.e., $\Ycal = \bigcup_{U \subset \Sigma} \mathrm{cyclic\_permutations}(U)$, for an input $x \in \Xcal$. Let $\psi: \Ycal \to \R^{\Sigma \times \Sigma}$ with $\psi_{uv}(y) = 1$ if $\{u,v\}\in y$ and $0$ otherwise.

Suppose we can compute \mbox{$\ln Z(w) = \ln \sum_{y\in\Ycal} \exp(\ip{\psi(y)}{w})$} efficiently. Given an arbitrary graph $G=(V,E)$ with adjacency matrix $\bar{w}$, let $\Sigma = V$ and $w = \bar{w} \times \ln (|V|! \times |V|)$. We will show that $G$ has a Hamiltonian cycle if and only if $\ln Z(w) \geq |V| \times \ln (|V|! \times |V|)$.

Necessity: As the exponential function is positive and $\ln$ is monotone increasing, it follows that \mbox{$\ln Z(w) \geq |V| \times \ln(|V|! \times |V|)$}.

Sufficiency: First, observe that $|\Ycal| < |V|! \times |V|$. Suppose G has no Hamiltonian cycle. Then
\begin{equation*}
\begin{aligned}
\ln Z(w) & \leq \ln [|\Ycal| \times \exp[ (|V|-1) \times \ln(|V|! \times |V|) ]] \\
& = \ln |\Ycal| + (|V|-1) \times \ln(|V|! \times |V|) \\
& < |V| \times \ln(|V|! \times |V|) \ .
\end{aligned}
\end{equation*}
This completes the proof.
\end{proof}

We are interested in the class of problems for which sampling uniformly at random is easy, and cyclic permutations is one example of these. The above result shows that computing the partition function is hard even if we restrict the problem to this class. Essentially, it transfers the general NP-hardness result of computing the partition function to the restricted class of problems that we are interested in. In the following section, we show how to approximate the partition function given that there exist efficient algorithms for uniform sampling.

\section{Approximating the Partition Function Using Uniform Samplers}\label{ssc:apx_partition}
As a first step towards approximating the partition function, let us consider using concentration inequalities. If we can sample uniformly at random from $\Ycal$, then we can apply Hoeffding's inequality to bound the deviation of the partition function $Z(w\mid x)$ from its finite sample expectation $\hat{Z}(w\mid x)$. Let $S$ denote the sample size. Then 
\begin{equation*}
\hat{Z}(w\mid x) = \sum\limits_{i=1}^S\frac{|\Ycal|}{S} \left[\exp(\ip{\phi(x,y_i)}{w})\right] \ .
\end{equation*}
\hide{By Cauchy-Schwarz's inequality, we have $|\ip{\phi(x,y)}{w}| \leq RB$.A direct application of Hoeffding's inequality gives us the following result:
\begin{equation*}
P(|Z(w\mid x) - \hat{Z}(w\mid x)| \geq \epsilon) \leq 2 \exp \left( \frac{-\epsilon^2 S}{|\Ycal|^2R^2B^2} \right) \ .
\end{equation*}
Unfortunately, this bound is not useful due to its dependence on the size of the output space $|\Ycal|$.} Unfortunately, the
bound obtained from using Hoeffding’s inequality is not useful due to its dependence on the size of the output space $|\Ycal|$. We now present an algorithm that is a \emph{fully-polynomial randomised approximation scheme} for computing the partition function.

\begin{definition}
Suppose $f : P \to \R^+$ is a function that maps problem instances $P$ to positive real numbers. A \emph{randomised approximation scheme} for $P$ is a randomised algorithm that takes as input an instance $p  \in P$ and an error parameter $\epsilon > 0$, and produces as output a number $Q$  such that
\begin{equation*}
\Pr[(1-\epsilon)f(p) \leq Q \leq (1+\epsilon)f(p)]\geq \frac{3}{4} \ .
\end{equation*}
A randomised approximation scheme is said to be \emph{fully polynomial} (FPRAS) if it runs in time polynomial in the length of $p$ and $1/\epsilon$.
\end{definition}

We exploit the intimate connection between counting and sampling problems \cite{Jerrum/etal/86} to approximately compute the partition function using sampling algorithms. The technique is based on a reduction from counting to sampling. \hide{Computing the partition function is basically a counting problem and if we are able to sample from the corresponding output space $\Ycal$, then it is also possible to solve the counting problem with approximation guarantees.}The standard approach \cite{Sinclair/Jerrum/96} is to express the quantity of interest, i.e., the partition function $Z(w\mid x)$, as a telescoping product of ratios of parameterised variants of the partition function. Let \mbox{$0=\beta_0 < \beta_1 \cdots < \beta_l=1$} denote a sequence of parameters also called as \emph{cooling schedule} and express $Z(w\mid x)$ as a telescoping product
\begin{equation*}
\frac{Z(w\mid x)}{Z(\beta_{l-1}w\mid x)} \times \frac{Z(\beta_{l-1}w\mid x)}{Z(\beta_{l-2}w\mid x)} \times \cdots \times \frac{Z(\beta_1w\mid x)}{Z(\beta_0w\mid x)} \times Z(\beta_0w\mid x) \ .
\end{equation*}
\hide{
\begin{equation*}
Z(\beta) = \frac{Z(\beta_l)}{Z(\beta_{l-1})} \times \frac{Z(\beta_{l-1})}{Z(\beta_{l-2})} \times \frac{Z(\beta_{l-2})}{Z(\beta_{l-3})} \cdots \frac{Z(\beta_1)}{Z(\beta_0)} \times Z(\beta_0) \ .
\end{equation*}
}
Define the random variable \mbox{$f_i(y) = \exp[(\beta_{i-1}-\beta_i)\ip{\phi(x,y)}{w}]$} (we omit the dependence on $x$ to keep the notation clear), for all $i \in \numset{l}$, where $y$ is chosen according to the distribution $\pi_{\beta_i}=p(y\mid x,\beta_iw)$. We then have
\begin{equation*}
\begin{aligned}
\E_{y \thicksim \pi_{\beta_i}} f_i & = \sum\limits_{y \in \Ycal} \exp[(\beta_{i-1}-\beta_i)\ip{\phi(x,y)}{w}] \frac{\exp[\beta_i \ip{\phi(x,y)}{w}]}{Z(\beta_iw\mid x)} \\
& = \frac{Z(\beta_{i-1}w\mid x)}{Z(\beta_iw\mid x)} \ ,
\end{aligned}
\end{equation*}
which means that $f_i(y)$ is an unbiased estimator for the ratio 
\begin{equation*}
\rho_i= \frac{Z(\beta_{i-1}w\mid x)}{Z(\beta_iw\mid x)} \ .
\end{equation*}
This ratio can now be estimated by sampling using a Markov chain according to the distribution $\pi_{\beta_i}$ and computing the sample mean of $f_i$. The desideratum is an upper bound on the variance of this estimator. Having a low variance implies a small number of samples $S$ suffices to approximate each ratio well. The final estimator is then the product of the reciprocal of the individual ratios in the telescoping product.

We now proceed with the derivation of an upper bound on the variance of the random variable $f_i$, or more precisely on the quantity $B_i = \Var f_i/(\E f_i)^2$. We first assume that $Z(\beta_0w\mid x)=|\Ycal|$ can be computed in polynomial time. This assumption is true for all combinatorial structures consider in this chapter. If it is not possible to compute $|\Ycal|$ in polynomial time, then we can approximate it using the same machinery described in this section. We use the following cooling schedule \cite{Stefankovic/etal/07}:
\begin{equation*}
l = p \lceil R \|w\|  \rceil; \quad  \beta_j = \frac{j}{pR\|w\|}, \quad  \forall j \in \numset{l-1} \ ,
\end{equation*}
where $p$ is a constant integer $\geq 3$, i.e., we let the cooling schedule to be of the following form:
\begin{equation*}
0, \frac{1}{q}, \frac{2}{q}, \frac{3}{q}, \ldots, \frac{p  \lfloor R \|w\| \rfloor }{q}, 1  \ ,
\end{equation*}
where $q = pR\|w\|$ (w.l.o.g. we assume that $R\|w\|$ is non-integer). Given this cooling schedule, observe that \mbox{$\exp(-1/p) \leq f_i \leq \exp(1/p)$}, which follows from the definition of the random variable $f_i$, and also that 
\begin{equation*}
\exp(-1/p) \leq \E f_i = \rho_i \leq \exp(1/p) \ .
\end{equation*}
We are now ready to prove the bound on the quantity $B_i$.

\begin{proposition}\label{lem:variance_bound}
$B_i = \frac{\Var f_i}{(\E f_i)^2} \leq \exp(2/p),  ~\forall~ i \in \numset{l}$.
\end{proposition}

We first need to prove the following lemma.
\begin{lemma}\label{lem:rho_bound}
%$\exp(1/p)-\exp(1/(2p)) \leq \rho_i \leq \exp(-1/p)+\exp(1/(2p)), ~\forall~ p \geq 2$.
\mbox{$\exp(1/p)-1 \leq \rho_i \leq \exp(-1/p)+1$.} %~\forall~ p \geq 3$.}
\end{lemma}
\begin{proof}
$a \leq b \dann \exp(a) - \exp(-a) \leq \exp(b) - \exp(-b)$ as the exponential function is monotone increasing. Thus $a \leq 1/3 \dann \exp(a) - \exp(-a) \leq \exp(1/3) - \exp(-1/3) < 1$. \hide{Now, $\exp (a) \geq 1, ~\forall~ a \geq 0$. Therefore, we have $0 \leq a\leq 1/3 \dann \exp(a) - \exp(-a) < \exp(a/2)$.} Setting $a=1/p$ with $p \geq 3$ and using the fact that $\exp(-1/p) \leq \rho_i \leq \exp(1/p)$ for all $i \in \numset{l}$ proves the lemma.
\end{proof}
\\\\\noindent
\begin{proof} (of Proposition~\ref{lem:variance_bound}) Consider \mbox{$\rho_i \geq \exp(1/p)-1 \geq f_i - 1$}. This implies \mbox{$f_i - \rho_i  \leq 1$}. Next, consider \mbox{$\rho_i \leq \exp(-1/p)+1 \leq f_i + 1$}. This implies \mbox{$f_i - \rho_i  \geq -1$}. Combining these, we get \mbox{$|f_i-\rho_i| \leq 1$}, which implies \mbox{$\Var f_i \leq 1$}, and therefore \mbox{$\Var f_i/(\E f_i)^2 \leq \exp(2/p)$.}
\end{proof}

Equipped with this bound, we are ready to design an FPRAS for approximating the partition function. We need to specify the sample size $S$ in each of the Markov chain simulations needed to compute the ratios.
\begin{theorem} \label{th:fpras}
Suppose the sample size \mbox{$S = \lceil 65 \epsilon^{-2}l \exp(2/p) \rceil$} and suppose it is possible to sample \emph{exactly} according to the distributions $\pi_{\beta_i}$, for all $i \in \numset{l}$, with polynomially bounded time. Then, there exists an FPRAS with $\epsilon$ as the error parameter for computing the partition function.
\end{theorem}
\begin{proof}
The proof uses standard techniques described in \cite{Sinclair/Jerrum/96}. Let $X^{(1)}_i, \ldots, X^{(S)}_i$ be a sequence of $S$ independent copies of the random variable $f_i$ obtained by sampling from the distribution $\pi_{\beta_i}$, and let $\bar{X}_i = S^{-1}\sum_{j=1}^S X_i^{(j)}$ be the sample mean. We have $\E\bar{X}_i = \E f_i=\rho_i$, and $\Var \bar{X_i} = S^{-1} \Var f_i$. The final estimator $\rho = Z(w\mid x)^{-1}$ is the random variable $X=\prod_{i=1}^l \bar{X}_i$ with $\E X = \prod_{i=1}^l \rho_i = \rho$. Now, consider 
\begin{equation*}
\begin{aligned}
\frac{\Var X}{(\E X)^2} & = \frac{\Var \bar{X_1} \bar{X_2} \cdots \bar{X_l}}{(\E \bar{X_1} \E \bar{X_2} \cdots \E \bar{X_l})^2} \\ %\quad \textrm{(denominator follows from independence of the r.v.)}\\
& = \frac{\E(\bar{X_1}^2 \bar{X_2}^2 \cdots \bar{X_l}^2) - [\E (\bar{X_1} \bar{X_2} \cdots \bar{X_l})]^2}{(\E \bar{X_1} \E \bar{X_2} \cdots \E \bar{X_l})^2} \\
& = \frac{\E \bar{X_1}^2 \E \bar{X_2}^2 \cdots \E \bar{X_l}^2}{(\E \bar{X_1} \E \bar{X_2} \cdots \E \bar{X_l})^2} - 1 \\
& = \prod\limits_{i=1}^l \left(1+ \frac{\Var \bar{X}_i}{(\E \bar{X}_i)^2}\right) - 1 \\
& \leq \left(1+ \frac{\exp(\frac{2}{p})}{S}\right)^l - 1 \\
& \leq \exp\left(\frac{l \exp(\frac{2}{p})}{S}\right) - 1 \\ 
& \leq \epsilon^2 / 64 \ ,
\end{aligned}
\end{equation*}
where the last inequality follows by choosing $S = \lceil 65 \epsilon^{-2}l \exp(2/p) \rceil$ and using the inequality $\exp(a/65) \leq 1 + a/64$ for $0 \leq a \leq 1$. By applying Chebyshev's inequality to $X$, we get
\begin{equation*}
\Pr[(|X-\rho|) > (\epsilon/4)\rho] \leq \frac{16}{\epsilon^2} \frac{\Var X}{(\E X)^2} \leq \frac{1}{4} \ ,
\end{equation*}
and therefore, with probability at least $3/4$, we have
\begin{equation*}
\left(1-\frac{\epsilon}{4}\right)\rho \leq X \leq \left(1+\frac{\epsilon}{4}\right)\rho \ .
\end{equation*}
Thus, with probability at least $3/4$, the partition function $Z(w\mid x)=X^{-1}$ lies within the ratio $(1 \pm \epsilon/4)$ of $\rho^{-1}$. Polynomial run time immediately follows from the assumption that we can sample exactly according to the distributions $\pi_{\beta_i}$ in polynomial time.
\end{proof}

We have shown how to approximate the partition function under the assumption that there exists an exact sampler for $\Ycal$ from the distribution $\pi_{\beta_i}$ for all $i \in \numset{l}$. A similar result can be derived by relaxing the exact sampling assumption and is proved in Appendix~\ref{sec:apx_samples}. In fact, it suffices to have only an exact \emph{uniform} sampler. As we will see in Section~\ref{sec:sampling}, it is possible to obtain exact samples from distributions of interest other than uniform if there exists an exact uniform sampler.

\section{Approximating the Partition Function Using Counting Formulae} \label{sec:part_count}
In this section, we describe how to approximate the partition function using counting formulae (cf. Section~\ref{sc:assumptions}), which obviates the need to use the sophisticated machinery described in the previous section. However, the downside of this technique is that it is not an FPRAS and works only for a specific feature representation.

Consider output spaces $\Ycal$ with a finite dimensional embedding $\psi: \Ycal \to \{0,+1\}^d$, and an input space $\Xcal \subseteq \R^n$. Define the scoring function $f(x,y) = \ip{w}{\phi(x,y)}$, where $\phi(x,y) = \psi(y) \otimes x$. Suppose that the size of the output space $|\Ycal|$, the vector $\Psi=\sum_{y \in \Ycal} \phi(y)$, and the matrix $C=\sum_{y \in \Ycal} \psi(y) \psi^\top(y)$ can be computed efficiently in polynomial time. We first observe that given these quantities, it is possible to compute $\sum_{y \in \Ycal} f(x, y)$ and $\sum_{y \in \Ycal} f^2(x, y)$ (as in the linear model described in Section~\ref{sc:scale}), and these are given as
\begin{equation*}
\sum\limits_{y \in \Ycal} f(x, y) = \ip{w}{\Psi \otimes x} ,
\end{equation*}
\begin{equation*}
\sum\limits_{y \in \Ycal} f^2(x, y)
=  w^\top (C \otimes xx^\top) w \ .
\end{equation*}
We then consider the second order Taylor expansion of the $\exp$ function at $0$, i.e., $\texp(a)=1+a+\frac{1}{2}a^2\approx\exp(a)$ and write the partition function as 
\begin{equation*}
\begin{aligned}
Z(w\mid x)  & = \sum\limits_{y \in \Ycal} \exp[f(x,y)] \\
& \approx \sum\limits_{y \in \Ycal} 1 + f(x,y) + f^2(x,y) \\
& = |\Ycal| + \ip{w}{\Psi \otimes x} + w^\top (C \otimes xx^\top) w \ .
\end{aligned}
\end{equation*}

At first glance, one may argue that using the second order Taylor expansion is a crude way to approximate the partition function. But observe that in a restricted domain around $[-1,1]$, the second order Taylor expansion approximates the $\exp$ function very well as illustrated in  Figure~\ref{fig:exp}.
\begin{figure}[tb]
\begin{center}
\epsfig{file=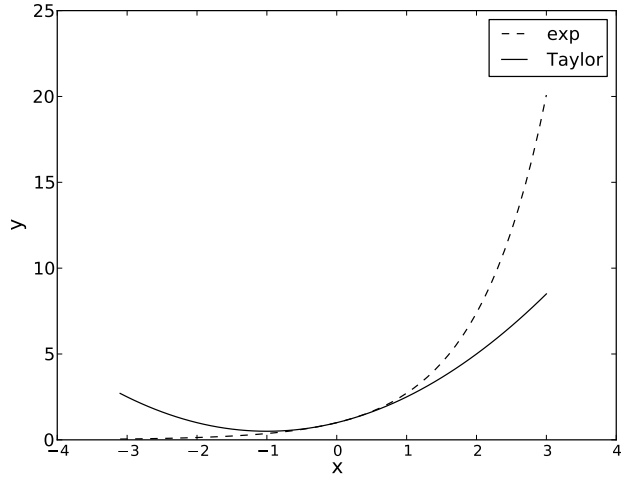,width=.75\textwidth}
\caption{Exponential function and its second Taylor approximation.}\label{fig:exp}
\end{center}
\end{figure}
In order to exploit this property, we have to constrain the scoring function $f(x,y)$ to lie in the range $[-1,1]$. But from Proposition~\ref{prop:norm_param}, we know that $\|w\| \leq \sqrt{\ln |\Ycal| / \lambda}$. A direct application of Cauchy-Schwarz's inequality gives us the following bound on the scoring function: $|f(x,y)| \leq R\sqrt{\ln |\Ycal| / \lambda}$.
An immediate consequence is a bound that relates the regularisation parameter and the scoring function.
\begin{proposition}
$\lambda  \geq R^2 \ln |\Ycal| \dann |f(x,y)| \leq 1$.
\end{proposition}

\section{Approximating the Gradient of the Log Partition Function}\label{sc:apx_gradient}
The optimisation problem \eq{eqn:opt} is typically solved using gradient descent methods which involves gradient-vector multiplications. We now describe how to approximate the gradient-vector multiplication with provable guarantees using concentration inequalities. Let $z$ be a vector in $\R^n$ (where $n$ is the dimension of the feature space $\phi(x,y)$) with bounded $\ell_2$ norm, i.e., $\|z\| \leq G$, where $G$ is a constant. The gradient-vector multiplication is given as
\begin{equation*}
\ip{\nabla_{w}\ln Z(w\mid x)}{z} = \E_{y \thicksim p(y\mid x,w)}[\ip{\phi(x,y)}{z}]  \ .
\end{equation*}
We use Hoeffding's inequality to bound the deviation of $\ip{\nabla_{w}\ln Z(w\mid x)}{z}$ from its estimate $\ip{d(w\mid x)}{z}$ on a finite sample of size $S$, where
\begin{equation*}
d(w\mid x) = \frac{1}{S}\sum\limits_{i=1}^S \phi(x,y_i)  \ ,
\end{equation*}
and the sample is drawn according to $p(y\mid x,w)$.

Note that by Cauchy-Schwarz's inequality, we have $|\ip{\phi(x,y_i)}{z}| \leq RG$, for all $i \in \numset{S}$. Applying Hoeffding's inequality, we then obtain the following exponential tail bound:
\begin{equation*}
\begin{aligned}
%& \Pr( |\nabla_{w} g_z(w\mid x) - \nabla_{w} \hat{g}_z(w\mid x))| \geq \epsilon) \leq 2 \exp\left( \frac{-\epsilon^2s^2}{2R^2G^2} \right) \ .
\Pr( |\ip{\nabla_{w}\ln Z(w\mid x)-d(w\mid x)}{z}| \geq \epsilon) \leq 2 \exp\left( \frac{-\epsilon^2S}{2R^2G^2} \right) \ .
\end{aligned}
\end{equation*}

For online optimisation methods like stochastic gradient descent and advancements thereof \cite{Hazan/etal/07,Shwartz/etal/07,Ratliff/etal/2007}, the optimisation problem is solved using plain gradient descent, and therefore it might be desirable to approximate the gradient and analyse the effects of the approximation guarantee on factors such as rates of convergence and regret bounds \cite{Ratliff/etal/2007}. In Appendix~\ref{sec:apx_gradient_mcmc}, we show how to approximate the gradient of the log partition function using the machinery described in Section~\ref{ssc:apx_partition}, i.e., using the reduction from counting to sampling.

\section{Sampling Techniques} \label{sec:sampling}
We now focus on designing sampling algorithms. These algorithms are needed (i) to compute the partition function using the machinery described in Section~\ref{ssc:apx_partition}, and (ii) to do inference, i.e., predict structures, using the learned model by solving the optimisation problem $\argmax_{y \in \Ycal} p(y\mid x,w)$ for any $x \in \Xcal$. Sampling algorithms can be used for optimisation using the Metropolis process \cite{Sinclair/Jerrum/96} and possibly other methods like simulated annealing for convex optimisation \cite{Adam/Vempala/06}. \hide{We will revisit these techniques in Chapter~\ref{ch:conclusions}.}Note that these methods come with provable guarantees and are not heuristics.

\subsection{Basics of Markov Chains} \label{sc:BasicsMCMC}
We start with some preliminaries on Markov chains. The exposition mostly follows the articles by \citet{Sinclair/Jerrum/96,Jerrum98,Randall/03}, and the lecture notes of \citet{Vigoda06}.

Let $\Omega$ denote the state space of a Markov chain $\M$ with transition probability matrix $P : \Omega \times \Omega \to [0,1]$. Let $P^t(u,\cdot)$ denote the distribution of the state space at time $t$ given that $u \in \Omega$ is the initial state. Let $\pi$ denote the stationary distribution of $\M$. A Markov chain is said to be \emph{ergodic} if the probability distribution over $\Omega$ converges asymptotically to $\pi$, regardless of the intial state. A Markov chain is said to be (a) \emph{irreducible} if for all $u,v \in \Omega$, there exists a $t$ such that $P^t(u,v) > 0$, and (b) \emph{aperiodic} if $\textrm{gcd}\{P(u,u) > 0 \} = 1$, for all $u \in \Omega$. Any finite, irreducible, aperiodic Markov chain is ergodic. A Markov chain is said to be \emph{time-reversible} with respect to $\pi$ if it satisfies the following condition also known as \emph{detailed balanced equations}:
\begin{equation*}
\pi(u)P(u,v) = \pi(v)P(v,u), \quad \forall u,v \in \Omega \ .
\end{equation*}
The \emph{mixing time} of a Markov chain is a measure of the time taken by the chain to converge to its stationary distribution. It is measured by the \emph{total variation distance} between the distribution at time $t$ and the stationary distribution. The total variation distance at time $t$ is
\begin{equation*}
\|P^{t},\pi\|_{tv} = \max\limits_{u \in \Omega} \frac{1}{2}\sum\limits_{v \in \Omega}|P^t(u,v) - \pi(u)| \ .
\end{equation*}
For any $\epsilon > 0$, the mixing time $\tau(\epsilon)$ is given by
\begin{equation*}
\tau(\epsilon) = \min\{t: \|P^{t'},\pi\|_{tv} \leq \epsilon, \ \forall \ t'\geq t\}\ .
\end{equation*}
A Markov chain is said to be \emph{rapidly mixing} if the mixing time is bounded by a polynomial in the input and $\ln \epsilon^{-1}$. We now describe techniques to bound the mixing time of Markov chains.

\subsubsection{Canonical Paths}
Let $\M$ be an ergodic, time-reversible Markov chain with state space $\Omega$, transition probabilties $P(\cdot,\cdot)$ and stationary distribution $\pi$. Given that $\M$ satisfies the detailed balanced condition, we may view $\M$ as an undirected graph $(\Omega, E)$ with vertex set $\Omega$ and edge set \mbox{$E = \{\{u,v\} \in \Omega \times \Omega : Q(u,v) > 0\}$}, where $Q(u,v) = \pi(u)P(u,v) = \pi(v)P(v,u)$.

For every ordered pair $(u,v) \in \Omega \times \Omega$, a canonical path $\gamma_{uv}$ from $u$ to $v$ in the graph corresponds to a sequence of legal transitions in $\M$ that leads from the initial state $u$ to the final state $v$. Let $\Gamma = \{\gamma_{uv}:u,v\in \Omega\}$ be the set of all canonical paths. In order to obtain good bounds on the mixing time, it is important to choose a set of paths $\Gamma$ that avoids the creation of ``hot spots:" edges that carry a heavy burden of canonical paths. The degree to which an even loading has been achieved is measured by a quantity known as \emph{congestion}, which is defined as follows:
\begin{equation*}
\rho({\Gamma}) = \max\limits_{e}\frac{1}{Q(e)}\sum\limits_{\gamma_{uv} \ni e} \pi(u)\pi(v) |\gamma_{uv}| \ ,
\end{equation*}
where the maximum is over oriented edges $e$ of $(\Omega, E)$ and $|\gamma_{uv}|$ denotes the length of the path $\gamma_{uv}$.  Intuitively, we expect a Markov chain to be rapidly mixing if it admits a choice of paths $\Gamma$ for which the congestion $\rho({\Gamma})$ is not too large. This intuition is formalised in the following result due to \citet{Sinclair/92}.
\begin{theorem} \cite{Sinclair/92} \label{th:maxflow}
Let $\M$ be an ergodic, time-reversible Markov chain with stationary distribution $\pi$ and self-loop probabilities $P(v,v) \geq 1/2$ for all states $v \in \Omega$. Let $\Gamma$ be a set of canonical paths with maximum edge loading $\bar{\rho} = \bar{\rho}(\Gamma)$. Then the mixing time of $\M$ satisfies 
\begin{equation*}
\tau_u(\epsilon) \leq \bar{\rho}(\ln \pi(u)^{-1} + \ln \epsilon^{-1}) \ ,
\end{equation*}
for any choice of initial state $u \in \Omega$.
\end{theorem}

\subsubsection{Coupling}
A coupling is a Markov process $(X_t,Y_t)_{t=0}^\infty$ on $\Omega \times \Omega$ such that each of the processes $P_t$ and $Q_t$, considered in isolation, is a faithful copy of $\M$, i.e.,
\begin{equation*}
\begin{aligned}
\Pr(X_{t+1}& =x' \mid  X_t = x \wedge Y_t = y) = P(x,x') \\
\Pr(Y_{t+1}&=y' \mid  X_t = x \wedge Y_t = y) = P(y,y') \ ,
\end{aligned}
\end{equation*}
and also 
\begin{equation*}
X_t=Y_t \dann X_{t+1}=Y_{t+1} \ .
\end{equation*}
If it can be arranged that the processes $(X_t)$ and $(Y_t)$ \emph{coalesce} rapidly independent of the initial states $X_0$ and $Y_0$, we may deduce that the Markov chain $\M$ is rapidly mixing. The key result is what is called the ``Coupling Lemma" due to \citet{Aldous/83}.
\begin{lemma}\cite{Aldous/83} \label{lem:coupling_lemma}
\textbf{(Coupling lemma)} Suppose $\M$ is a countable, ergodic Markov chain with transition probabilities $P(\cdot,\cdot)$ and let $(X_t,Y_t)_{t=0}^\infty$ be a coupling. Suppose $t: (0,1] \to \N$ is a function such that $\Pr(X_{t(\epsilon)}\neq Y_{t(\epsilon)}) \leq \epsilon$, for all $\epsilon \in (0,1]$, uniformly over the choice of initial state $(P_0,Q_0)$. Then the mixing time $\tau(\epsilon)$ of $\mathcal{M}$ is bounded from above by $t(\epsilon)$.
\end{lemma}

\subsubsection{Path Coupling}
\hide{
\begin{definition}
A coupling is a Markov chain on $\Omega \times \Omega$ defining a stochastic process $(P_t,Q_t)_{t=0}^\infty$ such that each of the processes $P_t$ and $Q_t$ is a faithful copy of $\M$ and $P_t=Q_t \dann P_{t+1}=Q_{t+1}$.
\end{definition}
\begin{definition}
For initial states $p$ and $q$, the coupling time $T = \max_{p,q} \E[\min \{t:P_t=Q_t \mid  P_0=p,Q_t=q\}]$.
\end{definition}
The \emph{coupling time} $T$ provides a good bound on the mixing time of $\M$.
\begin{theorem} (version from \cite{Randall/03})
The mixing time of $\M$ is bounded from above by $\tau(\epsilon) \leq  \lfloor T \mathrm{e} \ln \epsilon^{-1} \rfloor$, where $\mathrm{e}$ is the base of the natural logarithm.
\end{theorem}
}
Although coupling is a powerful technique to bound the mixing time, it may not be easy to measure the expected change in distance between two arbitrary configurations. The idea of path coupling introduced by \citet{Bubley/Dyer/97} is to consider only a small set of pairs of configurations $U \subseteq \Omega \times \Omega$ that are \emph{close} to each other w.r.t. a distance metric $\delta: \Omega \times \Omega \to \R$. Suppose we are able to measure the expected change in distance for all pairs in $U$. Now, for any pair $P,Q \in \Omega$, we define a shortest path $P=r_0,r_1, \ldots, r_s=Q$ of length $s$ from $P$ to $Q$ (sequence of transitions of minimal weight from $P$ to $Q$), where $(r_l,r_{l+1}) \in U$ for all $0 \leq l < s$. If we define $U$ appropriately, then $\delta(P,Q) = \sum_{l=0}^{s-1}\delta(r_l,r_{l+1})$, and by linearity of expectation, the expected change in distance between the pair $(P,Q)$ is just the sum of the expected change between the pairs $(z_l,z_{l+1})$. We now state the ``path coupling lemma" of \citet{Bubley/Dyer/97}.
\hide{
\begin{lemma} \cite{Bubley/Dyer/97}
\textbf{(Coupling lemma)} Suppose $P$ and $Q$ be a random process (the coupling) such that marginally, $P$ and $Q$ are both copies of $\mathcal{M}$. Moreover, suppose $Q_0$ is chosen from $\pi$, and $\mu_t$ is the distribution of $P_t$, then $d_{TV}(\mu_t,\pi) = Pr(P_t \neq Q_t)$, where $d_{TV}$ is the total variation distance metric on measures.
\end{lemma}
}
\begin{lemma}\cite{Bubley/Dyer/97} \label{th:pathcoupling}
\textbf{(Path Coupling lemma)} Let $\delta$ be an integer valued metric defined on $\Omega \times \Omega$, which takes values in $\numset{B}$. Let $U$ be a subset of $\Omega \times \Omega$ such that for all $(P_t,Q_t) \in \Omega \times \Omega$ there exists a path $P_t = r_0,r_1, \ldots, r_s = Q_t$ between $P_t$ and $Q_t$ where $(r_l,r_{l-1}) \in U$ for $0 \leq l < s$ and $\sum_{l=0}^{s-1} \delta(r_l,r_{l+1}) = \delta(P_t,Q_t)$. Suppose a coupling $(P,Q) \mapsto (P',Q')$ of the Markov chain $\M$ is defined on all pairs of $(P,Q) \in U$ such that there exists a $\beta < 1$ such that $\E[\delta(P',Q')] \leq \beta \E[\delta(P,Q)]$ for all $(P,Q) \in U$. Then the mixing time $\tau(\epsilon)$ is bounded from above as follows:
\begin{equation*}
\tau(\epsilon) \leq \frac{\ln B\epsilon^{-1}}{1-\beta} \ .
\end{equation*}
\end{lemma}

\subsubsection{Coupling from the Past}
Coupling from the past (CFTP) \cite{Propp/Wilson/96,Huber98} is a technique used to obtain an exact sample from the stationary distribution of a Markov chain. The idea is to simulate Markov chains forward from times in the past, starting in all possible states, as a coupling process. If all the chains coalesce at time $0$, then \citet{Propp/Wilson/96} showed that the current sample has the stationary distribution.

Suppose $\M$ is an ergodic (irreducible, aperiodic) Markov chain with (finite) state space $\Omega$, transition probabilties $P(\cdot,\cdot)$ and stationary distribution $\pi$. Suppose $\F$ is a probability distribution on functions $f: \Omega \to \Omega$ with the property that for every $u \in \Omega$, its image $f(u)$ is distributed according to the transition probability of $\M$ from state $u$, i.e.,
\begin{equation*}
\Pr_{\F}(f(u)=v) = P(u,v), ~ \forall u,v \in \Omega \ .
\end{equation*}

To sample $f \in \F$, we perform the following steps: (i) sample, independently for each $u \in \Omega$, a state $v_u$ from the distribution $P(v,\cdot)$, and (ii) let $f:\Omega \to \Omega$ be the function mapping from $u$ to $v_u$ for all $u \in \Omega$. Now, let $f_s, \ldots, f_{t-1}: \Omega \to \Omega$ with $s < t$ be an indexed sequence of functions, and denote by $F_s^t: \Omega \to \Omega$ the iterated function composition
\begin{equation*}
F_s^t = f_{t-1} \circ f_{t-2} \circ \cdots \circ f_{s+1} \circ f_{s} \ .
\end{equation*}

A $t$-step simulation of $\M$ can be performed as follows starting from some initial state $u_0 \in \Omega$: (i) select $f_0, \ldots, f_{t-1}$ independently from $\F$, (ii) compute the composition $F_0^t = f_{t-1} \circ f_{t-2} \circ \cdots \circ f_{1} \circ f_{0}$, and (iii) return $F_0^t(u_0)$. Of course, this is very inefficient requiring about $|\Omega|$ times the work of a direct simulation. However, this view will be convenient to explain the conceptual ideas behind CFTP as given below.

For fixed transition probabilities $P(\cdot,\cdot)$, there is a considerable flexibility in the choice of distributions $\F$, allowing us to encode uniform couplings over the entire state space. The Coupling Lemma can be stated in this setting. Suppose $f_0, \ldots, f_{t-1}$ are sampled independently from $\F$. If there exists a function $t:(0,1] \to \N$ such that
\begin{equation*}
\Pr[F_0^{t(\epsilon)}(\cdot) ~ \text{is not a constant function}] \leq \epsilon \ ,
\end{equation*}
then the mixing time $\tau(\epsilon)$ of $\M$ is bounded by $t(\epsilon)$. Thus, in principle, it is possible to estimate the mixing time of $\M$ by observing the coalesence time of the coupling defined by $\F$, and thereby obtain samples from an \emph{approximation} to the stationary distribution of $\M$ by simulating the Markov chain for a number of steps comparable with the empirically observed mixing time. However, in practice, the explicit evaluation of $F_0^t$ is computationally infeasible.

The first idea underlying Propp and Wilson's proposal was to work with $F_{-t}^0$ instead of $F^{t}_0$, i.e., by ``coupling from the past", it is possible to obtain samples from the \emph{exact} stationary distribution.

\begin{theorem}
Suppose that $f_{-1},f_{-2}, \ldots$ is a sequence of independent samples from $\F$. Let the stopping time $T$ be defined as the smallest number $t$ for which $F_{-t}^0(\cdot)$ is a constant function, and assume that $\E T < \infty$. Denote by $\hat{F}_{-\infty}^0$ the unique value of $F_{-T}^0$ (which is defined with probability 1). Then $\hat{F}_{-\infty}^0$ is distributed according to the stationary distribution of $\M$.
\end{theorem}

The second idea underlying Propp and Wilson's proposal is that in certain cases, specifically when the coupling $\F$ is ``monotone", it is possible to evaluate $F_{-T}^0$ without explicity computing the function composition $f_1 \circ f_2 \circ \cdots \circ f_{-T+1} \circ f_{-T}$. Suppose that the state space $\Omega$ is partially ordered $\succeq$ with maximal and minimal elements $\top$ and $\bot$ respectively. A coupling $\F$ is monotone if it satisfies the following property:
\begin{equation*}
u \succeq v \dann f(u) \succeq f(v), \quad \forall u,v \in \Omega \ .
\end{equation*}
If $\F$ is montone, then $F_{-t}^0 (\top) = F_{-t}^0 (\bot)$ implies $F_{-t}^0$ is a constant function and that $F_{-t}^0 (\top) = F_{-t}^0 (\bot) = \hat{F}^0_{\infty}$. Therefore, it suffices to track the two trajectories starting at $\top$ and $\bot$ instead of tracking $|\Omega|$ trajectories. Since only an upper bound on $T$ is needed for computing $\hat{F}^0_{\infty}$, a doubling scheme $t = 1,2,4,8,16, \dots$ is typically used rather than iteratively computing $F_{-t}^0$ for $t=0,1,2,3,4, \dots$.

\subsection{A Meta Markov chain}
The main contribution of this section is a generic, `meta' approach that can be used to sample structures from distributions of interest given that there exists an exact uniform sampler. We start with the design of a Markov chain based on the Metropolis process \cite{Metropolis/etal/53} to sample according to exponential family distributions $p(y\mid x,w)$ under the assumption that there exists an exact uniform sampler for $\Ycal$. Consider the following chain \textsc{Meta}: If the current state is $y$, then
\begin{enumerate}
\item select the next state $z$ uniformly at random, and
\item move to $z$ with probability $\min [1, p(z \mid  x, w) / p(y \mid  x, w) ]$.
\end{enumerate}

\subsubsection{Exact Sampling Using Coupling from the Past}
As described in the previous section, CFTP is a technique to obtain an exact sample from the stationary distribution of a Markov chain. To apply CFTP for \textsc{Meta}, we need to bound the expected number of steps $T$ until all Markov chains are in the same state. For the chain \textsc{Meta}, this occurs as soon as we update all the states, i.e., if we run all the parallel chains with the same random bits, once they are in the same state, they will remain coalesced. This happens as soon as they all accept an update (to the same state $z$) in the same step. First observe that, using Cauchy-Schwarz and triangle inequalities, we have
\begin{equation*}
\forall y,y' \in \Ycal: |\ip{\phi(x,y)-\phi(x,y')}{w}| \leq 2BR \ .
\end{equation*}
The probability of an update is  given by
\begin{equation*}
\min_{y,y'} \left[1, \frac{p(y\mid x,w)}{ p(y'\mid x,w)} \right] \geq \exp(-2BR) \ .
\end{equation*}
We then have
\begin{equation*}
\begin{aligned}
\E T \leq & 1 \times \exp[-2BR]  + \\
 &  2 \times (1-\exp[-2BR]) \times \exp[-2BR]  + \\
&  3 \times (1-\exp[-2BR])^2 \times \exp[-2BR]  +  \cdots
\end{aligned}
\end{equation*}
By using the identity $\sum_{i=0}^\infty i \times a^{i} = a^{-1}/(a^{-1}-1)^2$ with $a = (1-\exp[-2BR])$, we get $\E T = \exp[2BR]$. We now state the main result of this section.
\begin{theorem} \label{th:meta_cftp}
The Markov chain \textsc{Meta} can be used to obtain an exact sample according to the distribution $\pi = p(y\mid x,w)$ with expected running time that satisfies $\E T \leq \exp(2BR)$.
\end{theorem}
Note that the running time of this algorithm is random. To ensure that the algorithm terminates with a probability at least $(1 - \delta)$, it is required to run it for an additional factor of $\ln(1/\delta)$ time \cite{Huber98}. In this way, we can use this algorithm in conjunction with the approximation algorithm for computing the partition function resulting in an FPRAS. The implication of this result is that we only need to have an exact uniform sampler in order to obtain exact samples from other distributions of interest. \hide{As we will see in the next section, designing an exact uniform sampler is possible for several combinatorial structures that are of importance in machine learning problems.}

We conclude this section with a few remarks on the bound in Theorem~\ref{th:meta_cftp} and its practical implications. At first glance, we may question the usefulness of this bound because the constants $B$ and $R$ appear in the exponent. But note that we can always set $R=1$ by normalising the features. Also, the bound on $B$ (cf. Proposition~\ref{prop:norm_param}) could be loose in practice as observed recently by \citet{Do/etal/09}, and thus the value of $B$ could be way below its upper bound $\sqrt{\ln |\Ycal|/\lambda}$. We could then employ techniques similar to those described by \citet{Do/etal/09} to design optimisation strategies that work well in practice. Also, note that the problem is mitigated to a large extent by setting $\lambda \geq \ln|\Ycal|$ and $R=1$.

While in this section we focused on designing a `meta' approach for sampling, we would like to emphasise that it is possible to derive improved mixing time bounds by considering each combinatorial structure individually. As an example, we design a Markov chain to sample from the vertices of a hypercube and analyse its mixing time using path coupling. Details are delegated to Appendix~\ref{sec:mc_cube}.

\subsubsection{Mixing Time Analysis using Coupling} \label{sec:meta_coupling}
\hide{We first consider \textsc{Meta} with an exact uniform sampler. We later generalise it to the approximate uniform sampling case. We will use the coupling lemma (cf. Lemma~\ref{lem:coupling_lemma}) in our analysis.}We now analyse the Markov chain \textsc{Meta} using coupling and the coupling lemma (cf. Lemma~\ref{lem:coupling_lemma}).
\begin{theorem} \label{th:meta_coupling}
The mixing time of \textsc{Meta} with an exact uniform sampler is bounded from above as follows:
\begin{equation*}
%\lceil (\ln \epsilon^{-1}) / \ln (\exp(-2BR) -1) \rceil \ .
\lceil (\ln \epsilon^{-1}) / \ln (1-\exp(-2BR))^{-1} \rceil \ .
\end{equation*}
\end{theorem}
\begin{proof}
Using Cauchy-Schwarz and triangle inequalities, we have
\begin{equation*}
\forall y,y' \in \Ycal: |\langle \phi(x,y)-\phi(x,y'), w \rangle| \leq 2BR \ .
\end{equation*}
The probability of an update is 
\begin{equation*}
\min_{y,y'} \left[1, \frac{p(y|x,w)}{ p(y'|x,w)} \right] \geq \exp(-2BR) \ .
\end{equation*}
The probability of not updating for $T$ steps is therefore less than $(1 - \exp(-2BR))^T$. Let 
\begin{equation*}
%t(\epsilon)=\lceil (\ln \epsilon^{-1}) / \ln (\exp(-2BR) -1) \rceil \ .
t(\epsilon)=\lceil (\ln \epsilon^{-1}) / \ln (1-\exp(-2BR))^{-1} \rceil \ .
\end{equation*}
We now only need to show that $\Pr(P_{t(\epsilon)} \neq Q_{t(\epsilon)}) = \epsilon$. Consider
\begin{equation*}
\begin{aligned}
& \Pr(P_{t(\epsilon)} \neq Q_{t(\epsilon)}) \\
& \leq (1 - \exp(-2BR))^{ (\ln \epsilon) / \ln (1 - \exp(-2BR) ) } \\
& = \exp[   \ln (  1 - \exp(-2BR)    +   \epsilon   -   1 + \exp(-2BR)  )   ] \\
& = \epsilon \ .
\end{aligned}
\end{equation*}
The bound follows immediately from the Coupling Lemma.
\end{proof}
\hide{
We now consider the case that there exists only an approximate uniform sampler for the output space $\Ycal$. Let $\pi$ be the stationary distribution of a Markov chain that can be used to obtain approximate samples uniformly at random, and let the total variation distance between $\pi$ and uniform distribution $U$ be at most $\epsilon_u$, i.e., $\|\pi, U\|_{tv} \leq \epsilon_u$. Consider the following modification of the Markov chain \textsc{Meta}: If the current state is $y$, then
\begin{enumerate}
\item select the next state $z$ uniformly at random, and
\item move to $z$ with probability 
\begin{equation*}
\min \left[1, \frac{\pi(y\mid x,w)p(z\mid x,w)}{\pi(z\mid x,w)p(y\mid x,w)} \right] \ .
\end{equation*}
\end{enumerate}
We are essentially running Metropolis-Hastings algorithm with $\pi$ as the proposal density. We may not be able to compute $\pi(\cdot)$, but we can use the following lower bound in the second step of \textsc{Meta}:
\begin{equation*}
\min_{y,y'} \left[ \frac{\pi(y\mid x,w)}{\pi(y'\mid x,w)} \right] \geq \frac{\frac{1}{|\Ycal|}-\epsilon_u}{\frac{1}{|\Ycal|}+\epsilon_u}  \ .
\end{equation*}
Let $c$ denote the r.h.s. of the above inequality. The probability of an update is now given as
\begin{equation*}
\min_{y,y'} \left[1, \frac{\pi(y'\mid x,w)p(y\mid x,w)}{\pi(y\mid x,w)p(y'\mid x,w)} \right] \ .
\end{equation*}
Note that the above expression is bounded from below by $c\exp(-2BR)$. Using an analysis similar to the one described in the proof of Theorem~\ref{th:meta_coupling} for the exact uniform sampler, we arrive at the following result:
\begin{theorem} \label{th:meta_coupling_apx}
The mixing time of \textsc{Meta} with an approximate uniform sampler is bounded from above as follows:
\begin{equation*}
\lceil (\ln \epsilon^{-1}) / \ln (1-c\exp(-2BR))^{-1} \rceil \ .
\end{equation*}
\end{theorem}
If $\epsilon_u=0$, then $c=1$, and we recover \textsc{Meta} with an exact uniform sampler. Note that in order to have a non-trivial bound, we need to ensure that $\epsilon_u < 1 / |\Ycal|$.
}
\section{Summary}
The primary focus of this chapter was to rigorously analyse probabilistic structured prediction models using tools from MCMC theory. We designed algorithms for approximating the partition function and the gradient of the log partition function with provable guarantees. We also presented a simple Markov chain based on Metropolis process that can be used to sample according to exponential family distributions given that there exists an exact uniform sampler. \hide{While we were able to design an exact uniform sampler for combinatorial structures like vertices of a hypercube, permutations and subtrees of a tree (cf. Section~\ref{sc:assumptions}), we note that this may not be feasible in general for all machine learning applications. In such cases, we can design a Markov chain to obtain approximate samples from distributions of interest and also bound its mixing time. This is possible using coupling technique. Indeed, we show how to obtain approximate samples given that there exists an exact uniform sampler and the analysis is given in \eq{sec:meta_coupling}. We note that the coupling technique is much more amenable than the coupling from the past technique to the problem of obtaining approximate samples from a non-uniform distribution given only approximate uniform samples.}

If we were to solve the optimisation problem \eq{eqn:opt} using iterative techniques like gradient descent, then we have to run Markov chain simulations for every training example in order to compute gradients in any iteration of the optimisation routine. We therefore argue for using online convex optimisation techniques \cite{Hazan/etal/07,Shwartz/etal/07} as these would result in fast, scalable algorithms for structured prediction.

\clearemptydoublepage
\chapter{Conclusions}\label{ch:conclusions}
State-of-the-art approaches for structured prediction like structured SVMs \cite{Tsochantaridis/etal/05} have the flavour of an \emph{algorithmic template} in the sense that if, for a given output structure, a certain assumption holds, then it is possible to train the learning model and perform inference efficiently. The standard assumption is that the argmax problem 
\begin{equation*}
\hat{y} = \argmax_{z \in \Ycal} f(x,z)
\end{equation*}
is tractable for the output set $\Ycal$. Thus, all that is required to learn and predict a new structure is to design an algorithm to solve the argmax problem efficiently in polynomial time. The primary focus of this thesis was on predicting combinatorial structures such as cycles, partially ordered sets, permutations, and several other graph classes. We studied the limitations of existing structured prediction models \cite{Collins02,Tsochantaridis/etal/05,Taskar/etal/05} for predicting these structures. The limitations are mostly due to the argmax problem being intractable. In order to overcome these limitations, we introduced new assumptions resulting in two novel \emph{algorithmic templates} for structured prediction.

Our first assumption is based on counting combinatorial structures. We proposed a kernel method that can be trained efficiently in polynomial time and also designed approximation algorithms to predict combinatorial structures. The resulting algorithmic template is a generalisation of the classical regularised least squares regression for structured prediction. It can be used to solve several learning problems including multi-label classification, ordinal regression, hierarchical classification, and label ranking in addition to the aforementioned combinatorial structures.

Our second assumption is based on sampling combinatorial structures. Based on this assumption, we designed approximation algorithms with provable guarantees to compute the partition function of probabilistic models for structured prediciton where the class conditional distribution is modeled using exponential families. Our approach uses classical results from Markov chain Monte Carlo theory \cite{Sinclair/Jerrum/96,Randall/03}. It can be used to solve several learning problems including but not limited to multi-label classification, label ranking, and multi-category hierarchical classification.

We now point to some directions for future research. In Section~\ref{ch_:inference}, we designed approximation algorithms for solving the argmax problem for combinatorial output sets, but could do so with only weak approximation guarantees. While the approximation factor has no effect in training models such as structured ridge regression (cf. Chapter~\ref{ch:csop}), we believe there is a need to further investigate the possibilities of designing approximation algorithms with improved guarantees. We describe two promising directions for future work below.

\subsubsection{Approximate Inference using Simulated Annealing}
If we restrict ourselves to linear models, then the argmax problem reduces to a linear programming problem:
\begin{equation*}
\hat{y} = \argmax_{y \in \Ycal} \ip{w}{\phi(x,y)} \ .
\end{equation*}
Recently, \citet{Adam/Vempala/06} considered using simulated annealing \cite{Kirkpatrick/etal/83} to minimise a linear function over an arbitrary convex set. More precisely, they considered the following linear minimisation problem: for a unit vector $c \in \R^n$ and a convex set $K \subset \R^n$:
\begin{equation*}
\min_{ z \in K} \ip{c}{z} \ ,
\end{equation*}
under the assumption that there exists a membership oracle for the set $K$. The algorithm, described below, goes through a series of decreasing \emph{temperatures} as in simulated annealing, and at each temperature, it uses a random walk based sampling algorithm \cite{Lovasz/Vempala/04} to sample a point from the stationary distribution whose density is proportional to $\exp(-\ip{c}{z}/T)$, where $T$ is the current temperature. Let $R$ be an an upper bound on the radius of a ball that contains the convex set $K$, and $r$ be a lower bound on the radius of a ball around the starting point contained in $K$. At every temperature (indexed by $i$), the algorithm performs the following steps:
\begin{enumerate}
\item Set the temperature $T_i = R(1-1/\sqrt{n})^i$.
\item Move the current point to a sample from a distribution whose density is proportional to $\exp(-\ip{c}{z}/T_i)$, obtained by executing $k$ steps of a biased random walk (refer \cite{Lovasz/Vempala/04} for details).
\item Using $O^*(n)$\footnote{The $O^*$ notation hides logarithmic factors.} samples observed during the above walk and estimate the covariance matrix  of the above distribution.
\end{enumerate}
\begin{theorem} \cite{Adam/Vempala/06} \label{th:sa}
For any convex set $K \in \R^n$, with probability $1-\delta$, the above procedure given a membership oracle for the convex set $K$, starting point $z_0$, $R$, $r$, $I=O(\sqrt{n}\log(Rn/\epsilon\delta))$, $k=O^*(n^3)$, and $N=O^*(n)$, outputs a point $z_I \in K$ such that
\begin{equation*}
\ip{c}{z_I} \leq \min\limits_{z \in K} \ip{c}{z} + \epsilon \ .
\end{equation*}
The total number of calls to the membership oracle is $O^*(n^{4.5})$.
\end{theorem}

While the combinatorial structures considered in this work do not form a convex set, it would be interesting to consider the convex hull of such sets, for instance, the convex hull of the vertices of a hypercube and derive results similar in spirit to Theorem~\ref{th:sa}. For sampling, we can use the algorithms described in Section~\ref{sec:sampling} to sample from exponential family distributions.

\subsubsection{Approximate Inference using Metropolis Process}
The Metropolis process \cite{Metropolis/etal/53} can be seen as a special case of simulated annealing with fixed temperature, i.e., the Markov chain on the state space $\Omega$ is designed to be \emph{time-homogeneous}. This chain can be used to maximise (or minimise) objective functions $f: \Omega \to \R$ defined on the combinatorial set $\Omega$ \cite{Sinclair/Jerrum/96}. Consider the following chain $\mathrm{MC}(\alpha)$ which is similar to the `meta' Markov chain described in Section~\ref{sec:sampling}. If the current state is $y$, then
\begin{enumerate}
\item select the next state $z$ uniformly at random, and
\item move to $z$ with probability $\min (1, \alpha^{f(y)-f(z)} )$,
\end{enumerate}
where $\alpha \geq 1$ is a fixed parameter. The stationary distribution of this chain is
\begin{equation*}
\pi_{\alpha}(\cdot) = \frac{\alpha^{f(\cdot)}}{Z(\alpha)} \ ,
\end{equation*}
where $Z(\alpha)$ is the partition function.

Note that $\pi_{\alpha}$ is a monotonically increasing function of $f(\cdot)$ as desired. It is now crucial to choose the parameter $\alpha$ appropriately. Having a small $\alpha$ would make the distribution $\pi_{\alpha}$ well concentrated whereas having a large $\pi_{\alpha}$ would make the chain less mobile and susceptible to locally optimal solutions. Note than we can analyse the mixing time of this chain using the techniques described in Section~\ref{sec:sampling}. The probability of finding the optimal solution is at least the weight of such solutions in the stationary distribution $\pi_{\alpha}$. Therefore, if we can derive non-trivial upper bounds on the weights of optimal solutions, then we can be certain of hitting such solutions using the chain with high probability. The success probability can be boosted by running the chain multiple times. Choosing an appropriate $\alpha$ and deriving non-trivial upper bounds on the the weights of optimal solutions is an open question. Note that if we set $\alpha = \exp(1)$, we are essentially sampling from exponential family distributions.

\hide{
\section{Learning with Minimal Supervision}
In supervised machine learning, if labeling training data is expensive, then it becomes imperative to exploit information present in auxiliary data sources. To this end, several new paradigms of learning like semi-supervised learning \cite{Blum/Mitchell/98}, transfer learning \cite{Thrun95} , multi-task learning \cite{Caruana97}, and active learning \cite{Angluin87} have produced a plethora of learning algorithms that use unlabeled data and data from multiple tasks. A natural extension of our results described in Chapter~\ref{ch:psp} would be to design a semi-supervised learning algorithm using hybrids of generative and discriminative models \cite{Lasserre/etal/06,Agarwal/Daume/09}. Also, having a probabilistic model for structured prediction makes it appealing to design semi-supervised and multi-task learning algorithms using the machinery of Bayesian statistics (see, for example, \cite{Xue/etal/07}).

\section{Large-scale Learning}
While there has been a considerable amount of work in scaling up binary classifiers for large data sets \cite{Langford/etal/07,Shwartz/etal/07}, research on large-scale structured prediction is still in its infancy. Investigating online convex optimisation techniques \cite{Hazan/etal/07,Shwartz/etal/07} for training probabilistic models for structured prediction (cf. Chapter~\ref{ch:psp}) is an interesting direction for future research.
}

\clearemptydoublepage
\appendix
\hide{
\chapter{Concentration Inequalities}
We state the concentration inequalities used in Chapter~\ref{ch:psp}.
\hide{
\begin{theorem}(Cauchy-Schwarz's Inequality)
For vectors $x,y$ in an inner product space, $|\ip{x}{y}| \leq \|x\| \cdot \|y\|$.
\end{theorem}
}
\begin{theorem}(Chebyshev's Inequality)
Let $X$ be a random variable with expected value $\mu$ and variance $\sigma^2$. Then for any $\epsilon > 0$,
\begin{equation*}
\Pr(|X-\mu| \geq \epsilon\sigma) \leq \frac{1}{\epsilon^2} \ .
\end{equation*}
\end{theorem}

\begin{theorem}(Hoeffding's Inequality \cite{Hoeffding63})
Let $X_1, X_2, \cdots, X_n$ be $n$ independent random variables such that $ a_i \leq X_i \leq b_i$ with probability $1$ for all $i \in \numset{n}$, where $a_i$ and $b_i$ are real constants. Then, the sum of these random variables $S = \sum_{i=1}^n X_i$ satisfies the following inequalities for any $\epsilon >0$:
\begin{equation*}
\Pr(S-\E S \geq n\epsilon) \leq \exp\left(-\frac{2n^2\epsilon^2}{\sum_{i=1}^n (b_i-a_i)^2}\right) \ ,
\end{equation*}
and
\begin{equation*}
\Pr(|S-\E S| \geq n\epsilon) \leq 2 \exp\left(-\frac{2n^2\epsilon^2}{\sum_{i=1}^n (b_i-a_i)^2}\right) \ .
\end{equation*}
\end{theorem}
}

\chapter{Kernels and Low-dimensional Mappings} \label{app:lde}
The Johnson-Lindenstrauss lemma \cite{Johnson/Lindenstrauss/84,Dasgupta/Gupta/99} states that any set of $m$ points in high dimensional Euclidean space can be mapped to an $n=O(\ln m / \epsilon^2)$ Euclidean space such that the distance between any pair of points are preserved within a ($1 \pm \epsilon$) factor, for any $0 < \epsilon < 1$. Such an embedding is constructed by projecting the $m$ points onto a random $k$-dimensional hypersphere. A similar result was proved by \citet{Arriaga/Vempala/99} and \citet{Balcan/etal/04} in the context of kernels. Given a kernel matrix $K$ in the (possibly) infinite dimensional $\phi$-space, it is possible to compute a low-dimensional mapping of the data points that approximately preserves linear separability. Specifically, if the data points are separated with a margin $\gamma$ in the $\phi$-space, then a random linear projection of this space down to a space of dimension $n= O(\frac{1}{\gamma^2} \ln \frac{1}{\epsilon\delta})$ will have a linear separator of error at most $\epsilon$ with probability at least $1-\delta$.

One of the drawbacks of using this random projection method for low-dimensional mapping of kernels is the need to store a dense random matrix of size $n \times m$ which is prohibitively expensive for large-scale applications. To circumvent this problem, one could resort to Nystr\"om approximation of the eigenvectors and eigenvalues of the kernel matrix using random sampling. This technique is a fast, scalable version of the classical multi-dimensional scaling (MDS) (see \cite{Platt05} and references therein), and is referred to as NMDS in the following.

Let $K$ be a positive definite matrix. Choose $k \ll m$ samples (or data points) at random, re-order the matrix such that these samples appear at the beginning, and partition it as
\begin{equation*}
K = 
\begin{pmatrix}
A & B \\
B^\top & C
\end{pmatrix}\ .
\end{equation*}
Let $A = U \Lambda U^\top$, $n$ be the dimension of the embedding, $U_{[n]}$ and $\Lambda_{[n]}$ be the submatrices corresponding to the $n$ largest eigenvectors and eigenvalues respectively. Then the low-dimensional embedding (matrix) is
\begin{equation*}
\begin{pmatrix}
U_{[n]}\Lambda_{[n]}^{1/2}\\
B^\top U_{[n]}\Lambda_{[n]}^{-1/2}
\end{pmatrix}\ .
\end{equation*}
The computational complexity of this approach is $O(nkm + k^3)$. 

The NMDS approach cannot be applied if the input matrix is indefinite. For example, in learning on graphs, the input is a sparse adjacency matrix but not a kernel. In such cases, we would still like to find a low-dimensional embedding of this graph without the need to compute a kernel as an intermediate step. Note that if we use the (normalised) Laplacian as the input, then the algorithm would output the leading eigenvectors, but we need to compute the trailing eigenvectors of the Laplacian. \citet{Fowlkes/etal/04} proposed a two-step approach to solve the problem even when the input matrix is indefinite. First, NMDS is applied to the input matrix. Let $\bar{U}^\top = [U^\top \Lambda^{-1}U^\top B]$, $Z = \bar{U}\Lambda^{1/2}$ and let $V \Sigma V^\top$ denote the orthogonalisation of $Z^\top Z$. Then the matrix $ZV\Sigma^{-1/2}$ contains the leading orthonormalised approximate eigenvectors of $K$ and the low-dimensional embedding is given by $ZV$. The computational complexity is still in the order of $O(nkm + k^3)$, but with an additional $O(k^3)$ orthogonalisation step.

\chapter{Counting Dicyclic Permutations}\label{app:dicycle}
% counting cyclic permutations
We represent a directed cyclic permutation by the set of (predecessor, successor)-pairs. For instance, the sequences $\{abc,bca,cab\}$ are equivalent and we use $\{(a,b),(b,c),(c,a)\}$ to represent them. Furthermore, we define $\psi^\mathrm{cyc}_{(u,v)}(y)=1$ if $(u,v)\in y$, i.e., $v$ follows directly after $u$ in $y$; $\psi^\mathrm{cyc}_{(u,v)}(y)=-1$ if $(v,u)\in y$, i.e., $u$ follows directly after $v$ in $y$; and $0$ otherwise. 

For a given alphabet $\Sigma$, we are now interested in computing $|\Ycal^\mathrm{cyc}|$, the number of cyclic permutations of subsets of $\Sigma$. For a subset of size $i$ there are $i!$ permutations of which $i$ represent the same cyclic permutation. That is, there are $(i-1)!$ cyclic permutations of each subset of size $i$, and for an alphabet of size $N=|\Sigma|$ there are
\begin{equation*}
|\Ycal^\mathrm{cyc}| = \sum_{i=3}^N\begin{pmatrix}N\\i\end{pmatrix} (i-1)!
\end{equation*}
different cyclic permutations of subsets of $\Sigma$.

Computing $\Psi^\mathrm{cyc}$ is simple. Observe that, for each cyclic permutation containing a (predecessor, successor)-pair $(u,v)$, there is also exactly one cyclic permutation containing $(v,u)$. Hence the sum over each feature is zero and
%\begin{equation*}
$\Psi^\mathrm{cyc} = \zero$ (where $\zero$ is the vector of all zeros).
%\end{equation*}

It remains to compute $C^\mathrm{cyc}$. Each element of this matrix is computed as 
\begin{equation*}
C^\mathrm{cyc}_{(u,v),(u',v')} = \sum_{y\in \Ycal^\mathrm{cyc}} \psi^\mathrm{cyc}_{(u,v)} (y) \psi^\mathrm{cyc}_{(u',v')}(y) \ .
\end{equation*}
As seen above, for $u=u'$ and $v=v'$  there are as many cyclic permutations for which $\psi^\mathrm{cyc}_{(u,v)}=+1$ as there are cyclic permutations for which $\psi^\mathrm{cyc}_{(u,v)}=-1$. In both cases, \mbox{$\psi^\mathrm{cyc}_{(u,v)} (y) \psi^\mathrm{cyc}_{(u',v')}(y)=+1$}, and to compute $C^\mathrm{cyc}_{(u,v),(u,v)}$ it suffices to compute the number of cyclic permutations containing $(u,v)$ or $(v,u)$. There are  $(i-2)!$ different cyclic permutations of each subset of size $i$ that contain $(u,v)$. We thus have 
\begin{equation*}
C^\mathrm{cyc}_{(u,v),(u,v)} = 2 \sum\limits_{i=3}^{N} \begin{pmatrix}N-2\\i-2\end{pmatrix} (i-2)!\ .
\end{equation*}
Similarly, for $u=v'$ and $v=u'$, we have $\psi^\mathrm{cyc}_{(u,v)} (y) \psi^\mathrm{cyc}_{(u',v')}(y)=-1$ and 
\begin{equation*}
C^\mathrm{cyc}_{(u,v),(v,u)} = - 2 \sum\limits_{i=3}^{N} \begin{pmatrix}N-2\\i-2\end{pmatrix} (i-2)!\ .
\end{equation*}

For $u\neq u' \neq v \neq v' \neq u$, we observe that there are as many cyclic permutations containing $(u,v)$ and $(u',v')$ as there are cyclic permutations containing $(u,v)$ and $(v',u')$, and hence in this case $C^\mathrm{cyc}_{(u,v),(u',v')}=0$. Finally, we need to consider $C^\mathrm{cyc}_{(u,v),(u,v')}$, $C^\mathrm{cyc}_{(u,v),(u',v)}$, $C^\mathrm{cyc}_{(u,v),(v,v')}$, and $C^\mathrm{cyc}_{(u,v),(u',u)}$. Here we have
\begin{equation*}
\begin{array}{rclr}
C^\mathrm{cyc}_{(u,v),(u,v')} =& C^\mathrm{cyc}_{(u,v),(u',v)} =& -2 \sum\limits_{i=3}^{N} \begin{pmatrix}N-3\\i-3\end{pmatrix} (i-3)! &\quad\text{and}\\
C^\mathrm{cyc}_{(u,v),(v,v')} =& C^\mathrm{cyc}_{(u,v),(u',u)} =& +2 \sum\limits_{i=3}^{N} \begin{pmatrix}N-3\\i-3\end{pmatrix} (i-3)! \ .
\end{array}
\end{equation*}

\chapter{Appendix for Chapter~\ref{ch:psp}}
\section{Approximating the Partition Function using Approximate Samples}\label{sec:apx_samples}
In Section~\ref{ssc:apx_partition}, we designed an FPRAS for approximating the partition function under the assumption that there exists an exact sampler. We now consider the case where we only have approximate samples resulting from a truncated Markov chain. Let $T_i$ denote the simulation length of the Markov chains, for all $i \in \numset{l}$. The main result of this section is as follows:
\begin{theorem}\label{th:fpras_bias}
Suppose the sample size \mbox{$S = \lceil 65 \epsilon^{-2}l \exp(2/p) \rceil$} and suppose the simulation length $T_i$ is large enough that the variation distance of the Markov chain from its stationary distribution $\pi_{\beta_i}$ is at most $\epsilon/5l\exp(2/p)$. Under the assumption that the chain is rapidly mixing, there exists an FPRAS with $\epsilon$ as the error parameter for computing the partition function.
\end{theorem}
\begin{proof}
The proof again uses techniques described in \cite{Sinclair/Jerrum/96}. The bound \mbox{$\Var f_i/(\E f_i)^2\leq \exp(2/p)$} (from Proposition~\ref{lem:variance_bound}) w.r.t. the random variable $f_i$ will play a central role in the proof. We cannot use this bound per se to prove approximation guarantees for the partition function $Z(w\mid x)$. This is due to the fact that the random variable $f_i$ is defined w.r.t. the distribution $\pi_{{\beta_i}}$, but our samples are drawn from a distribution $\hat{\pi}_{{\beta_i}}$ resulting from a truncated Markov chain, whose variation distance satisfies \mbox{$|\hat{\pi}_{{\beta_i}} - \pi_{{\beta_i}}| \leq \epsilon/5 l \exp(2/p)$}.
\hide{
\begin{equation*}
|\hat{\pi}_{{\beta_i}} - \pi_{{\beta_i}}| \leq \frac{\epsilon}{5 l \exp(2/p)} \ .
\end{equation*}
}
Therefore, we need to obtain a bound on $\Var \hat{f}_i/(\E \hat{f}_i)^2$ w.r.t. the random variable $\hat{f}_i$ defined analogously to $f_i$ with samples drawn from the distribution $\hat{\pi}_{\beta_i}$. An interesting observation is the fact that Lemma~\ref{lem:rho_bound} still holds for $\hat{\rho_i}$, i.e., \mbox{$\exp(1/p)-1 \leq \hat{\rho}_i \leq \exp(-1/p)+1,$}
\hide{
\begin{equation*}
%(1-\frac{\epsilon}{5l}) \rho_i \leq \hat{\rho}_i \leq (1+\frac{\epsilon}{5l}) \rho_i \ ,
\exp(1/p)-1 \leq \hat{\rho}_i \leq \exp(-1/p)+1, 
\end{equation*}
}
for all integers $p \geq 3$, and using similar analysis that followed Lemma~\ref{lem:rho_bound}, we get 
\begin{equation*}
\frac{\Var \hat{f_i}}{(\E \hat{f}_i)^2} \leq \exp(2/p), ~\forall~ i \in \numset{l} \ .
\end{equation*}

Also, note that $|\hat{\pi}_{{\beta_i}} - \pi_{{\beta_i}}| \leq \epsilon / 5 l \exp(2/p)$ implies \mbox{$|\hat{\rho}_i-\rho_i| \leq \epsilon/5 l \exp(1/p)$} (using the fact that $\exp(-1/p) \leq \rho_i \leq \exp(1/p)$). Therefore,
\begin{equation} \label{eqn:rho_bounds}
(1-\frac{\epsilon}{5l}) \rho_i \leq \hat{\rho}_i \leq (1+\frac{\epsilon}{5l}) \rho_i \ .
\end{equation}

Equipped with these results, we are ready to compute the sample size $S$ needed to obtain the desired approximation guarantee in the FPRAS. Let $X^{(1)}_i, \ldots, X^{(S)}_i$ be a sequence of $S$ independent copies of the random variable $\hat{f}_i$ obtained by sampling from the distribution $\hat{\pi}_{\beta_i}$, and let $\bar{X}_i = S^{-1}\sum_{j=1}^S X_i^{(j)}$ be the sample mean. We have $\E\bar{X}_i = \E\hat{f}_i=\hat{\rho_i}$, and $\Var \bar{X_i} = S^{-1} \Var \hat{f_i}$. The final estimator $\hat{\rho} = Z(w\mid x)^{-1}$ is the random variable $X=\prod_{i=1}^l \bar{X}_i$ with $\E X = \prod_{i=1}^l \hat{\rho}_i = \hat{\rho}$. From \eq{eqn:rho_bounds}, we have
\begin{equation}\label{eqn:rhohat_bounds}
(1-\frac{\epsilon}{4})\rho \leq \hat{\rho} \leq (1+\frac{\epsilon}{4})\rho \ .
\end{equation}
Now, consider 
\begin{equation*}
\begin{aligned}
\frac{\Var X}{(\E X)^2} & = \prod\limits_{i=1}^l \left(1+ \frac{\Var \bar{X}_i}{(\E \bar{X}_i)^2}\right) - 1 \\
& \leq \left(1+ \frac{\exp(\frac{2}{p})}{S}\right)^l - 1 \\
& \leq \exp\left(\frac{l \exp(\frac{2}{p})}{S}\right) - 1 \\ 
& \leq \epsilon^2 / 64 \ ,
\end{aligned}
\end{equation*}
if we choose $S = \lceil 65 \epsilon^{-2}l \exp(2/p) \rceil$ (because $\exp(a/65) \leq 1 + a/64$ for $0 \leq a \leq 1$). By applying Chebyshev's inequality to $X$, we get
\begin{equation*}
\Pr[(|X-\hat{\rho}|) > (\epsilon/4)\hat{\rho}] \leq \frac{16}{\epsilon^2} \frac{\Var X}{(\E X)^2} \leq \frac{1}{4} \ ,
\end{equation*}
and therefore, with probability at least $3/4$, we have
\begin{equation*}
(1-\frac{\epsilon}{4})\hat{\rho} \leq X \leq (1+\frac{\epsilon}{4})\hat{\rho} \ .
\end{equation*}
Combining the above result with \eq{eqn:rhohat_bounds}, we see that with probability at least $3/4$, the partition function $Z(w \mid x)=X^{-1}$ lies within the ratio $(1 \pm \epsilon/4)$ of $\rho^{-1}$. Polynomial run time follows from the assumption that the Markov chain is rapidy mixing.
\end{proof}

\section{Approximating the Gradient of the Log Partition Function using a Reduction from Counting to Sampling}\label{sec:apx_gradient_mcmc}
In this section, we show how to approximate the gradient of the log partition function using the reduction from counting to sampling described in Section~\ref{ssc:apx_partition}. Recall that the gradient of the log partition function generates the first order moment (mean) of $\phi(x,y)$, i.e.,
\begin{equation*}
\begin{aligned}
\nabla_{w} \ln Z(w\mid x) & = \frac{ \sum\limits_{y \in \Ycal} \phi(x,y) \exp(\ip{w}{\phi(x,y)})} {\sum\limits_{y \in \Ycal} \exp(\ip{w}{\phi(x,y)})} \\
& = \E_{y \thicksim p(y\mid x,w)}[\phi(x,y)] \ .
\end{aligned}
\end{equation*}
The numerator in the above expression is the quantity of interest and it can be seen as a \emph{weighted} variant of the partition function $Z(w\mid x)$ where the weights are the features $\phi(x,y)$. We will use $\phi_j(x,y)$ to denote the $j$th component of $\phi(x,y)$ and let $Z_j(w\mid x) = \sum_{y \in \Ycal}\phi_j(x,y) \exp(\ip{\phi(x,y)}{w})$. Consider again the random variable $f_i(y) = \exp[(\beta_{i-1}-\beta_i)\ip{\phi(x,y)}{w}]$, where $y$ is now chosen according to the distribution 
\begin{equation*}
\pi^j_{\beta_i}= \frac{\phi_j(x,y) \exp(\ip{\phi(x,y)}{\beta_iw})}{\sum_{y \in \Ycal}\phi_j(x,y) \exp(\ip{\phi(x,y)}{\beta_iw})} \ , 
\end{equation*}
i.e, we use the same random variable as defined in Section~\ref{ssc:apx_partition}, but sample according to a slightly different distribution as given above. It is easy to verify that $f_i(y)$ is an unbiased estimator for the ratios in the telescoping product of the quantity of interest, i.e,
\begin{equation*}
\E_{y \thicksim \pi^j_{\beta_i}} f_i = \frac{Z_j(\beta_{i-1}w\mid x)}{Z_j(\beta_{i} w \mid  x)}, ~ \forall i \in \numset{l} \ ,
\end{equation*}
where $l$ is the length of the cooling schedule (cf. Section~\ref{ssc:apx_partition}). It remains to analyse the mixing time of the Markov chain with stationary distribution $\pi^j_{\beta_i}$. Since features can take the value of zero, Theorems~\ref{th:meta_cftp} and \ref{th:meta_coupling} cannot be applied. One solution to this problem would be to modify the state space of the Markov chain in such a way that we sample (uniformly) only those structures satisfying $|\phi_j(x,y)| \geq \gamma$, where $\gamma$ is a parameter, and then run the Metropolis process. It would be interesting to further analyse the approximation that is introduced by discarding all those structures satisfying $|\phi_j(x,y)| < \gamma$, but we do not focus on this aspect of the problem here.

A note on computational issues follows. At first glance, it may seem computationally inefficient to run the machinery for every feature, but note that it is possible to reuse the computations of individual ratios of the telescoping product by designing an appropriate cooling schedule. First, consider the following expression:
\begin{equation*}
\begin{aligned}
& \sum\limits_{y \in \Ycal} \phi_j(x,y) \exp(\ip{w}{\phi(x,y)}) \\
= & \sum\limits_{y \in \Ycal} \exp(\ip{w}{\phi(x,y)} + \ln \phi_j(x,y)) \\
= & \sum\limits_{y \in \Ycal} \exp(c_j \ip{w}{\phi(x,y)}) \ , \\
\end{aligned}
\end{equation*}
where $c_j = 1+\ln \phi_j(x,y)/\exp(\ip{w}{\phi(x,y)})$. Let $c=\max_{j} c_j$. The cooling schedule is then given as
\begin{equation*}
0, \frac{1}{q}, \frac{2}{q}, \frac{3}{q}, \ldots, \frac{ c p  \lfloor R \|w\| \rfloor}{q}, 1  \ ,
\end{equation*}
where $q = cpR\|w\|$.

\section{Mixing Time Analysis of $\mathrm{MC_{cube}}$ using Path Coupling}\label{sec:mc_cube}
The state space $\Omega$ is the vertices of the $d$-dimensional hypercube $\{0,1\}^d$. Consider the following Markov chain $\mathrm{MC_{cube}}$ on $\Omega$. Initialise at $\zero$ and repeat the following steps: (i) pick $(i,b) \in \numset{d} \times \{0,1\}$, and (ii) move to the next state, formed by changing $i$th bit to $b$, with probability $\text{min}\left(1, \frac{\pi(v)}{\pi(u)}\right)$. Let $d(\cdot,\cdot)$ denote the Hamming distance. The transition probabilities of this chain are given by
\begin{align*}
P(u,v) = &\begin{cases}
\frac{1}{2d}\text{min}\left(1, \frac{\pi(v)}{\pi(u)}\right),\ \text{if}\  d(u,v) = 1; \\
\frac{1}{2},\ \text{if}\ u = v; \\
0,\ \text{otherwise} \ .
\end{cases}
\end{align*}

We now analyse the mixing time of $\mathrm{MC_{cube}}$ using path coupling (cf. Section~\ref{sc:BasicsMCMC}). Recall that the first step in using path coupling is to define a coupling on pairs of states that are close to each other according to some distance metric $\delta: \Omega \times \Omega \to \R$ on the state space.
\begin{theorem}
The Markov chain $\mathrm{MC_{cube}}$ on the vertices of a $d$-dimensional hypercube has mixing time $\tau(\epsilon) = O(d \ln (d \epsilon^{-1}))$.
\end{theorem}
\begin{proof}
We will first prove the bound on the mixing time for uniform stationary distribution and later generalise it to arbitrary distributions.

We choose Hamming distance as the metric, and consider pairs of states $(u,v)$ that differ by a distance of $1$. To couple, we choose $(i,b) \in \numset{d} \times \{0,1\}$ uniformly at random, and then update to $(u',v')$ formed by changing $i$th bit to $b$ if possible. We now need to show that the expected change in distance due to this update never increases. More precisely, we need to prove that
\begin{equation*}
\E[\delta(u',v')] = \beta \delta(u,v), \quad \beta \leq 1 \ ,
\end{equation*}
and invoke the path coupling lemma (cf. Lemma~\ref{th:pathcoupling}) to obtain the final result. Let $u$ and $v$ differ at the $j$-th bit. We have the following two cases.
\begin{itemize}
\item Case 1: $i \neq j$. In this case, if $u_i = v_i = b$, then there is no update and therefore no change in distance. If $u_i = v_i \neq b$, then both $u$ and $v$ update their $i$-th bit and therefore, again, there is no change in distance.
\item Case 2: $i = j$. In this case, there is definitely a decrease in distance with probability $1/d$ as one of $u$ or $v$ (but not both) will update their $i$-th bit.
\end{itemize}
We therefore have
\begin{equation*}
\begin{aligned}
\E[\delta(u',v')] & = 1 - \frac{1}{d} \\
& = \left(1 - \frac{1}{d}\right) \delta(u,v) \quad (\text{since } \delta(u,v) = 1) \\
& = \beta \delta(u,v)
\end{aligned}
\end{equation*}
with $\beta \le 1$ as desired. Invoking the path coupling lemma gives us the following bound on the mixing time
\begin{equation*}
\tau(\epsilon) \leq  d \ln (d \epsilon^{-1}) \ .
\end{equation*}

Let $z$ be the new state formed in cases 1 and 2 above. For non-uniform distributions, we have the following:
\begin{equation*}
\begin{aligned}
\E[\delta(u',v')] & = 1 +\frac{d-1}{2d} [P(u,z)(1-P(v,z)) + P(v,z)(1-P(u,z))] \\ 
& - \frac{1}{2d}[P(u,w)+P(v,w)] \ .
\end{aligned}
\end{equation*}
Note that $z$ differs from $u$ and $v$ by a single bit, and therefore under the assumptions that
\begin{equation*}
\begin{aligned}
P(u,z) &= \min\left[1,\frac{\pi(z)}{\pi(u)}\right] \approx 1 \quad \text{and} \\
P(v,z) &= \min\left[1,\frac{\pi(z)}{\pi(v)}\right] \approx 1 \ ,
\end{aligned}
\end{equation*}
we have, once again,
\begin{equation*}
\E[\delta(u',v')]  = \beta \delta(u,v) \ ,
\end{equation*}
with $\beta \le 1$.
\end{proof}

\hide{
\section{Vertices of a hypercube / Path Coupling}
The state space $\Omega$ is the vertices of the $d$-dimensional hypercube $\{0,1\}^d$. Consider the following Markov chain $\mathrm{MC_{cube}}$ on $\Omega$. Initialise at $\zero$ and repeat the following steps: (i) Pick $(i,b) \in \numset{d} \times \{0,1\}$, (ii) Move to the next state, formed by changing $i$th bit to $b$, with probability $\text{min}(1, \frac{\pi(v)}{\pi(u)})$. Let $d(.,.)$ denote the Hamming distance. The transition probabilities of this chain are given by
\begin{align*}
P(u,v) = &\begin{cases}
\frac{1}{2d}\text{min}\left(1, \frac{\pi(v)}{\pi(u)}\right),\ \text{if}\  d(u,v) = 1; \\
\frac{1}{2},\ \text{if}\ u = v; \\
0,\ \text{otherwise} \ .
\end{cases}
\end{align*}
A bound on the mixing time for $\mathrm{MC_{cube}}$ with uniform stationary distribution was given by \citet{Randall/03} using canonical paths. 
\begin{lemma}\cite{Randall/03} \label{th:mccube}
The mixing time $\tau(\epsilon)$ of $\mathrm{MC_{cube}}$ is bounded from above as follows:
\begin{equation*}
\tau(\epsilon) \leq d^2(d \log 2 + \log \epsilon^{-1}) \ .
\end{equation*}
\end{lemma}
We now generalise this result for arbitrary distributions.
\begin{theorem}
The mixing time $\tau(\epsilon)$ of $\mathrm{MC_{cube}}$ is bounded from above as follows:
\begin{equation*}
\tau(\epsilon) \leq O(1)(d^2(d \log 2 + \log \epsilon^{-1})) \ .
\end{equation*}
\end{theorem}
\begin{proof}
Let $e=(u,v)$ be an arbitrary edge. The capacity of this edge
\begin{align*}
Q(e) & = \pi(u)P(u,v) \\
& = \text{min}(\pi(u), \pi(v)) \\
& \geq \min\limits_z \pi(z) \\
& \geq e^{-2}/2^d \ .
\end{align*}
Now consider $\sum\limits_{\gamma_{st} \ni e} \pi(s)\pi(t)$. The number of terms in this summation is equal to the number paths passing through $e$ which is $2^{d-1}$. Therefore,
\begin{align*}
\sum\limits_{\gamma_{st} \ni e} \pi(s)\pi(t) &  \leq \sum\limits_{\gamma_{st} \ni e} (e^2/2^d)^2 \\
& \leq 2^{d-1}e^4 2^{-2d} \ .
\end{align*}
By noting that the length of any path is at most $d$, we have $\rho(\Gamma) = O(d)$.  We now apply Theorem~\ref{th:maxflow} to get
\begin{equation*}
\tau(\epsilon) \leq O(1)(d^2(d \log 2 + \log \epsilon^{-1})) \ .
\end{equation*}
\end{proof}
}

\clearemptydoublepage

\addcontentsline{toc}{chapter}{References}
\bibliographystyle{plainnat}
\bibliography{src/biblio}

\begin{thebibliography}{142}
\providecommand{\natexlab}[1]{#1}
\providecommand{\url}[1]{\texttt{#1}}
\expandafter\ifx\csname urlstyle\endcsname\relax
  \providecommand{\doi}[1]{doi: #1}\else
  \providecommand{\doi}{doi: \begingroup \urlstyle{rm}\Url}\fi

\bibitem[Ailon(2007)]{Ailon07}
Nir Ailon.
\newblock Aggregation of partial rankings, p-ratings and top-m lists.
\newblock In \emph{Proceedings of the 18th Annual ACM-SIAM Symposium on
  Discrete Algorithms}, 2007.

\bibitem[Ailon et~al.(2008)Ailon, Charikar, and Newman]{Ailon/etal/05}
Nir Ailon, Moses Charikar, and Alantha Newman.
\newblock Aggregating inconsistent information: {R}anking and clustering.
\newblock \emph{Journal of the ACM}, 55\penalty0 (5), 2008.

\bibitem[Aldous(1983)]{Aldous/83}
David Aldous.
\newblock Random walks on finite groups and rapidly mixing markov chains.
\newblock \emph{S\'eminaire de probabilit\'es de Strasbourg}, 17:\penalty0
  243--297, 1983.

\bibitem[Altun et~al.(2002)Altun, Hofmann, and Johnson]{Altun/etal/02}
Yasemin Altun, Thomas Hofmann, and Mark Johnson.
\newblock Discriminative learning for label sequences via boosting.
\newblock In \emph{Advances in Neural Information Processing Systems 15}, 2002.

\bibitem[Altun et~al.(2003)Altun, Johnson, and Hofmann]{Altun/etal/03}
Yasemin Altun, Mark Johnson, and Thomas Hofmann.
\newblock Loss functions and optimization methods for discriminative learning
  of label sequences.
\newblock In \emph{Proceedings of the Conference on Empirical Methods in
  Natural Language Processing}, 2003.

\bibitem[Andrieu et~al.(2003)Andrieu, de~Freitas, Doucet, and
  Jordan]{Andrieu/etal/03}
Christophe Andrieu, Nando de~Freitas, Arnaud Doucet, and Michael~I. Jordan.
\newblock An introduction to {MCMC} for machine learning.
\newblock \emph{Machine Learning}, 50\penalty0 (1-2):\penalty0 5--43, 2003.

\bibitem[Anguelov et~al.(2005)Anguelov, Taskar, Chatalbashev, Koller, Gupta,
  Heitz, and Ng]{Anguelov/etal/05}
Dragomir Anguelov, Benjamin Taskar, Vassil Chatalbashev, Daphne Koller, Dinkar
  Gupta, Geremy Heitz, and Andrew~Y. Ng.
\newblock Discriminative learning of {Markov} random fields for segmentation of
  {3D} scan data.
\newblock In \emph{Proceedings of the IEEE Conference on Computer Vision and
  Pattern Recognition}, 2005.

\bibitem[Arriaga and Vempala(1999)]{Arriaga/Vempala/99}
Rosa~I. Arriaga and Santosh Vempala.
\newblock An algorithmic theory of learning: {R}obust concepts and random
  projection.
\newblock In \emph{Proceedings of the 40th Annual Symposium on Foundations of
  Computer Science}, 1999.

\bibitem[Ausiello and Marchetti-Spaccamela(1980)]{AusMar80}
G.~Ausiello and A.~Marchetti-Spaccamela.
\newblock Toward a unified approach for the classification of {NP}-complete
  optimization problems.
\newblock \emph{Theoretical Computer Science}, 12\penalty0 (1):\penalty0
  83--96, September 1980.

\bibitem[Bakir et~al.(2007)Bakir, Hofmann, Sch\"olkopf, Smola, Taskar, and
  Vishwanathan]{Bakir/etal/07}
G\"okhan~H. Bakir, Thomas Hofmann, Bernhard Sch\"olkopf, Alexander~J. Smola,
  Ben Taskar, and S.V.N. Vishwanathan.
\newblock \emph{Predicting Structured Data}.
\newblock MIT Press, Cambridge, Massachusetts, USA, 2007.

\bibitem[Balcan et~al.(2006)Balcan, Blum, and Vempala]{Balcan/etal/04}
Maria-Florina Balcan, Avrim Blum, and Santosh Vempala.
\newblock Kernels as features: {O}n kernels, margins, and low-dimensional
  mappings.
\newblock \emph{Machine Learning}, 65\penalty0 (1):\penalty0 79--94, 2006.

\bibitem[Balcan et~al.(2008)Balcan, Bansal, Beygelzimer, Coppersmith, Langford,
  and Sorkin]{Balcan/etal/08}
Maria-Florina Balcan, Nikhil Bansal, Alina Beygelzimer, Don Coppersmith, John
  Langford, and Greg Sorkin.
\newblock Robust reductions from ranking to classification.
\newblock \emph{Machine Learning Journal}, 72\penalty0 (1-2):\penalty0
  139--153, 2008.

\bibitem[{Bartholdi~III} et~al.(1989){Bartholdi~III}, Tovey, and
  Trick]{Bartholdi/etal/89}
John {Bartholdi~III}, Craig Tovey, and Michael Trick.
\newblock Voting schemes for which it can be difficult to tell who won the
  election.
\newblock \emph{Social Choice and Welfare}, 6\penalty0 (2):\penalty0 157--165,
  1989.

\bibitem[Bj{\"o}rklund et~al.(2004)Bj{\"o}rklund, Husfeldt, and
  Khanna]{Bjorklund/etal/04}
Andreas Bj{\"o}rklund, Thore Husfeldt, and Sanjeev Khanna.
\newblock Approximating longest directed paths and cycles.
\newblock In \emph{Proceedings of the 31st International Colloquium on
  Automata, Languages and Programming}, 2004.

\bibitem[Block(1962)]{Block62}
H.D. Block.
\newblock The perceptron: {A} model for brain functioning.
\newblock \emph{Reviews of Modern Physics}, 34:\penalty0 123--135, 1962.

\bibitem[Boser et~al.(1992)Boser, Guyon, and Vapnik]{Boser/etal/92}
Bernhard~E. Boser, Isabelle Guyon, and Vladimir Vapnik.
\newblock A training algorithm for optimal margin classifiers.
\newblock In \emph{Proceedings of the 5th Annual ACM Conference on
  Computational Learning Theory}, 1992.

\bibitem[Boyd and Vandenberghe(2004)]{Boyd/Thomas/04}
Stephen Boyd and Lieven Vandenberghe.
\newblock \emph{Convex Optimization}.
\newblock Cambridge Univeristy Press, 2004.

\bibitem[Bregman(1967)]{Bregman/97}
L.~M. Bregman.
\newblock The relaxation method of finding the common points of convex sets and
  its application to the solution of problems in convex programming.
\newblock \emph{USSR Computational Mathematics and Mathematical Physics},
  7:\penalty0 200--217, 1967.

\bibitem[Brightwell and Winkler(1992)]{Brightwell/Winkler/92}
Graham Brightwell and Peter Winkler.
\newblock Counting linear extensions.
\newblock \emph{Order}, 8\penalty0 (3):\penalty0 225--242, 1992.

\bibitem[Brinker and H{\"u}llermeier(2006)]{Brinker/Huellermeier/06}
Klaus Brinker and Eyke H{\"u}llermeier.
\newblock Case-based label ranking.
\newblock In \emph{Proceedings of the 17th European Conference on Machine
  Learning}, 2006.

\bibitem[Brinker and H{\"u}llermeier(2007)]{Brinker/Huellermeier/07}
Klaus Brinker and Eyke H{\"u}llermeier.
\newblock Case-based multilabel ranking.
\newblock In \emph{Proceedings of the 20th International Joint Conference on
  Artificial Intelligence}, 2007.

\bibitem[Bubley and Dyer(1997)]{Bubley/Dyer/97}
Russ Bubley and Martin~E. Dyer.
\newblock Path coupling: {A} technique for proving rapid mixing in {M}arkov
  chains.
\newblock In \emph{Proceedings of the 38th Annual Symposium on Foundations of
  Computer Science}, 1997.

\bibitem[Caetano et~al.(2009)Caetano, McAuley, Cheng, Le, and
  Smola]{Caetano/etal/09}
Tib{\'e}rio~S. Caetano, Julian~John McAuley, Li~Cheng, Quoc~V. Le, and
  Alexander~J. Smola.
\newblock Learning graph matching.
\newblock \emph{IEEE Transactions on Pattern Analysis and Machine
  Intelligence}, 31\penalty0 (6):\penalty0 1048--1058, 2009.

\bibitem[Cesa-Bianchi et~al.(2006)Cesa-Bianchi, Gentile, and
  Zaniboni]{Cesa-Bianchi/etal/06}
Nicol{\`o} Cesa-Bianchi, Claudio Gentile, and Luca Zaniboni.
\newblock Incremental algorithms for hierarchical classification.
\newblock \emph{Journal of Machine Learning Research}, 7:\penalty0 31--54,
  2006.

\bibitem[Cheng and H{\"u}llermeier(2008)]{Cheng/Huellermeier/08}
Weiwei Cheng and Eyke H{\"u}llermeier.
\newblock Instance-based label ranking using the {M}allows model.
\newblock In \emph{Proceedings of the Preference Learning Workshop at the
  European Conference on Machine Learning and Principles and Practice of
  Knowledge Discovery in Databases}, 2008.

\bibitem[Cheng et~al.(2009)Cheng, Huhn, and H{\"u}llermeier]{Cheng/etal/09}
Weiwei Cheng, Jens~C. Huhn, and Eyke H{\"u}llermeier.
\newblock Decision tree and instance-based learning for label ranking.
\newblock In \emph{Proceedings of the 26th Annual International Conference on
  Machine Learning}, 2009.

\bibitem[Collins(2002)]{Collins02}
Michael Collins.
\newblock Discriminative training methods for hidden {M}arkov models: Theory
  and experiments with perceptron algorithms.
\newblock In \emph{Proceedings of the Conference on Empirical Methods in
  Natural Language Processing}, 2002.

\bibitem[Collins et~al.(2002)Collins, Schapire, and Singer]{Collins/etal/02}
Michael Collins, Robert~E. Schapire, and Yoram Singer.
\newblock Logistic regression, {AdaBoost} and {Bregman} distances.
\newblock \emph{Machine Learning}, 48\penalty0 (1-3):\penalty0 253--285, 2002.

\bibitem[Cortes and Vapnik(1995)]{Cortes/Vapnik/95}
Corinna Cortes and Vladimir Vapnik.
\newblock Support-vector networks.
\newblock \emph{Machine Learning}, 20\penalty0 (3):\penalty0 273--297, 1995.

\bibitem[Crammer and Singer(2003)]{Crammer/Singer/03a}
Koby Crammer and Yoram Singer.
\newblock A family of additive online algorithms for category ranking.
\newblock \emph{Journal of Machine Learning Research}, 3:\penalty0 1025--1058,
  2003.

\bibitem[Crammer and Singer(2005)]{Crammer/Singer/05}
Koby Crammer and Yoram Singer.
\newblock Loss bounds for online category ranking.
\newblock In \emph{Proceedings of the 18th Annual Conference on Learning
  Theory}, 2005.

\bibitem[Crammer et~al.(2006)Crammer, Dekel, Keshet, Shalev-Shwartz, and
  Singer]{Crammer/etal/06}
Koby Crammer, Ofer Dekel, Joseph Keshet, Shai Shalev-Shwartz, and Yoram Singer.
\newblock Online passive-aggressive algorithms.
\newblock \emph{Journal of Machine Learning Research}, 7:\penalty0 551--585,
  2006.

\bibitem[Cybenko(1989)]{Cybenko89}
G.~Cybenko.
\newblock Approximation by superpositions of a sigmoidal function.
\newblock \emph{Mathematics of Control, Signals, and Systems}, 2\penalty0
  (4):\penalty0 303--314, 1989.

\bibitem[Dasgupta and Gupta(1999)]{Dasgupta/Gupta/99}
Sanjoy Dasgupta and Anupam Gupta.
\newblock An elementary proof of the {Johnson–Lindenstrauss} lemma.
\newblock Technical Report TR 99–006, University of California, Berkeley,
  1999.

\bibitem[{Daum\'e III}(2006)]{Daume06}
Hal {Daum\'e III}.
\newblock \emph{Practical Structured Learning Techniques for Natural Language
  Processing}.
\newblock PhD thesis, University of Southern California, Los Angeles, CA,
  August 2006.

\bibitem[Dekel et~al.(2003)Dekel, Manning, and Singer]{Dekel/etal/03}
Ofer Dekel, Christopher~D. Manning, and Yoram Singer.
\newblock Log-linear models for label ranking.
\newblock In \emph{Advances in Neural Information Processing Systems 16}, 2003.

\bibitem[Dekel et~al.(2008)Dekel, Shalev-Shwartz, and Singer]{Dekel/etal/05}
Ofer Dekel, Shai Shalev-Shwartz, and Yoram Singer.
\newblock The forgetron: {A} kernel-based perceptron on a budget.
\newblock \emph{SIAM Journal on Computing}, 37\penalty0 (5):\penalty0
  1342--1372, 2008.

\bibitem[Dempster et~al.(1977)Dempster, Laird, and Rubin]{Dempster/etal/77}
Arthur Dempster, Nan Laird, and Donald Rubin.
\newblock Maximum likelihood from incomplete data via the {EM} algorithm.
\newblock \emph{Journal of the Royal Statistical Society. Series B
  (Methodological)}, 39\penalty0 (1):\penalty0 1--38, 1977.

\bibitem[Do et~al.(2009)Do, Le, and Foo]{Do/etal/09}
Chuong~B. Do, Quoc~V. Le, and Chuan-Sheng Foo.
\newblock Proximal regularization for online and batch learning.
\newblock In \emph{Proceedings of the 26th International Conference on Machine
  Learning}, 2009.

\bibitem[Dwork et~al.(2001)Dwork, Kumar, Naor, and Sivakumar]{Dwork/etal/01}
Cynthia Dwork, Ravi Kumar, Moni Naor, and D.~Sivakumar.
\newblock Rank aggregation methods for the web.
\newblock In \emph{Proceedings of the 10th International World Wide Web
  Conference}, 2001.

\bibitem[Elisseeff and Weston(2001)]{Elisseeff/Weston/01}
Andr\'{e} Elisseeff and Jason Weston.
\newblock A kernel method for multi-labelled classification.
\newblock In \emph{Advances in Neural Information Processing Systems 14}, 2001.

\bibitem[Elman(1990)]{Elman90}
Jeffrey~L. Elman.
\newblock Finding structure in time.
\newblock \emph{Cognitive Science}, 14\penalty0 (2):\penalty0 179--211, 1990.

\bibitem[Erhan et~al.(2006)Erhan, L'heureux, Yue, and
  Bengio]{erhan_collaborative}
Dumitru Erhan, Pierre-Jean L'heureux, Shi~Yi Yue, and Yoshua Bengio.
\newblock Collaborative filtering on a family of biological targets.
\newblock \emph{Journal of Chemical Information and Modeling}, 46\penalty0
  (2):\penalty0 626--635, 2006.

\bibitem[Fagin et~al.(2004)Fagin, Kumar, Mahdian, Sivakumar, and
  Vee]{Fagin/etal/04}
Ronald Fagin, Ravi Kumar, Mohammad Mahdian, D.~Sivakumar, and Erik Vee.
\newblock Comparing and aggregating rankings with ties.
\newblock In \emph{Proceedings of the 23rd ACM SIGACT-SIGMOD-SIGART Symposium
  on Principles of Database Systems}, 2004.

\bibitem[Finley and Joachims(2008)]{Finley/Joachims/08}
Thomas Finley and Thorsten Joachims.
\newblock Training structural {SVMs} when exact inference is intractable.
\newblock In \emph{Proceedings of the 25th International Conference on Machine
  Learning}, 2008.

\bibitem[Fowlkes et~al.(2004)Fowlkes, Belongie, Chung, and
  Malik]{Fowlkes/etal/04}
Charless Fowlkes, Serge Belongie, Fan R.~K. Chung, and Jitendra Malik.
\newblock Spectral grouping using the {Nystr\"om} method.
\newblock \emph{IEEE Transactions on Pattern Analysis and Machine
  Intelligence}, 26\penalty0 (2):\penalty0 214--225, 2004.

\bibitem[Frasconi and Passerini(2008)]{Frasconi/Passerini/08}
Paolo Frasconi and Andrea Passerini.
\newblock Predicting the geometry of metal binding sites from protein sequence.
\newblock In \emph{Advances in Neural Information Processing Systems 21}, 2008.

\bibitem[Freund and Schapire(1997)]{Freund/Schapire/97}
Yoav Freund and Robert~E. Schapire.
\newblock A decision-theoretic generalization of on-line learning and an
  application to boosting.
\newblock \emph{Journal of Computer and System Sciences}, 55\penalty0
  (1):\penalty0 119--139, 1997.

\bibitem[Freund and Schapire(1999)]{Freund/Schapire/99}
Yoav Freund and Robert~E. Schapire.
\newblock Large margin classification using the perceptron algorithm.
\newblock \emph{Machine Learning}, 37\penalty0 (3):\penalty0 277--296, 1999.

\bibitem[Froehlich and Krumm(2008)]{froehlich_route_2008}
Jon Froehlich and John Krumm.
\newblock Route prediction from trip observations.
\newblock In \emph{Society of Automotive Engineers World Congress}, 2008.

\bibitem[F{\"u}rnkranz(2002)]{Fuernkranz/02}
Johannes F{\"u}rnkranz.
\newblock Round robin classification.
\newblock \emph{Journal of Machine Learning Research}, 2:\penalty0 721--747,
  2002.

\bibitem[F{\"u}rnkranz and H{\"u}llermeier(2003)]{Fuernkranz/Huellermeier/03}
Johannes F{\"u}rnkranz and Eyke H{\"u}llermeier.
\newblock Pairwise preference learning and ranking.
\newblock In \emph{Proceedings of the 14th European Conference on Machine
  Learning}, 2003.

\bibitem[F{\"u}rnkranz et~al.(2008)F{\"u}rnkranz, H{\"u}llermeier, Menc\'{\i}a,
  and Brinker]{Fuernkranz/etal/08}
Johannes F{\"u}rnkranz, Eyke H{\"u}llermeier, Eneldo~Loza Menc\'{\i}a, and
  Klaus Brinker.
\newblock Multilabel classification via calibrated label ranking.
\newblock \emph{Machine Learning}, 73\penalty0 (2):\penalty0 133--153, 2008.

\bibitem[G\"artner(2005)]{Gaertner/05}
Thomas G\"artner.
\newblock \emph{Kernels for Structured Data}.
\newblock PhD thesis, Universit\"at Bonn, 2005.

\bibitem[Guruswami et~al.(2008)Guruswami, Manokaran, and
  Raghavendra]{Guruswami/etal/08}
Venkatesan Guruswami, Rajsekar Manokaran, and Prasad Raghavendra.
\newblock Beating the random ordering is hard : {I}napproximability of maximum
  acyclic subgraph.
\newblock In \emph{Proceedings of the 49th Annual IEEE Symposium on Foundations
  of Computer Science.}, 2008.

\bibitem[Halld{\'o}rsson(2000)]{Halldorsson/99}
Magn{\'u}s~M. Halld{\'o}rsson.
\newblock Approximations of weighted independent set and hereditary subset
  problems.
\newblock \emph{Journal of Graph Algorithms and Applications}, 4\penalty0 (1),
  2000.

\bibitem[Har-Peled et~al.(2002{\natexlab{a}})Har-Peled, Roth, and
  Zimak]{HarPeled/etal/02}
Sariel Har-Peled, Dan Roth, and Dav Zimak.
\newblock Constraint classification for multiclass classification and ranking.
\newblock In \emph{Advances in Neural Information Processing Systems 15},
  2002{\natexlab{a}}.

\bibitem[Har-Peled et~al.(2002{\natexlab{b}})Har-Peled, Roth, and
  Zimak]{HarPeled/etal/02b}
Sariel Har-Peled, Dan Roth, and Dav Zimak.
\newblock Constraint classification: A new approach to multiclass
  classification.
\newblock In \emph{Proceedings of the 13th International Conference on
  Algorithmic Learning Theory}, 2002{\natexlab{b}}.

\bibitem[Hassin and Khuller(2001)]{Hassin/Khuller/01}
Refael Hassin and Samir Khuller.
\newblock $z$-approximations.
\newblock \emph{Journal of Algorithms}, 41\penalty0 (2):\penalty0 429--442,
  2001.

\bibitem[Hassin and Rubinstein(1994)]{Hassin/Rubinstein/94}
Refael Hassin and Shlomi Rubinstein.
\newblock Approximations for the maximum acyclic subgraph problem.
\newblock \emph{Information Processing Letters}, 51:\penalty0 133--140, 1994.

\bibitem[Hazan et~al.(2007)Hazan, Agarwal, and Kale]{Hazan/etal/07}
Elad Hazan, Amit Agarwal, and Satyen Kale.
\newblock Logarithmic regret algorithms for online convex optimization.
\newblock \emph{Machine Learning}, 69\penalty0 (2-3):\penalty0 169--192, 2007.

\bibitem[Hochreiter and Schmidhuber(1997)]{Hochreiter/Schmidhuber/97}
Sepp Hochreiter and {Ju\"ergen} Schmidhuber.
\newblock Long short-term memory.
\newblock \emph{Neural Computation}, 9\penalty0 (8):\penalty0 1735--1780, 1997.

\bibitem[Huber(2006)]{Huber06}
Mark Huber.
\newblock Fast perfect sampling from linear extensions.
\newblock \emph{Discrete Mathematics}, 306\penalty0 (4):\penalty0 420--428,
  2006.

\bibitem[Huber(1998)]{Huber98}
Mark Huber.
\newblock Exact sampling and approximate counting techniques.
\newblock In \emph{Proceedings of the 30th Annual ACM Symposium on Theory of
  Computing}, 1998.

\bibitem[H{\"u}llermeier et~al.(2008)H{\"u}llermeier, F{\"u}rnkranz, Cheng, and
  Brinker]{Huellermeier/etal/08}
Eyke H{\"u}llermeier, Johannes F{\"u}rnkranz, Weiwei Cheng, and Klaus Brinker.
\newblock Label ranking by learning pairwise preferences.
\newblock \emph{Artificial Intelligence}, 178:\penalty0 1897--1916, 2008.

\bibitem[Jacob and Vert(2008)]{jacob_protein-ligand_2008}
Laurent Jacob and Jean-Philippe Vert.
\newblock Protein-ligand interaction prediction: {An} improved chemogenomics
  approach.
\newblock \emph{Bioinformatics}, 24\penalty0 (19):\penalty0 2149--2156, October
  2008.

\bibitem[Jerrum(1998)]{Jerrum98}
Mark Jerrum.
\newblock Mathematical foundations of the {M}arkov chain {M}onte {C}arlo
  method.
\newblock In \emph{Probabilistic Methods for Algorithmic Discrete Mathematics},
  pages 116--165. Springer-Verlag, 1998.

\bibitem[Jerrum and Sinclair(1996)]{Sinclair/Jerrum/96}
Mark~R. Jerrum and Alistair~J. Sinclair.
\newblock The {Markov chain Monte Carlo} method: {A}n approach to approximate
  counting and integration.
\newblock In \emph{{Hochbaum DS(ed) Approximation Algorithms for NP–hard
  Problems}}, pages 482--520. PWS Publishing, Boston, Mass., 1996.

\bibitem[Jerrum et~al.(1986)Jerrum, Valiant, and Vazirani]{Jerrum/etal/86}
Mark~R. Jerrum, Leslie~G. Valiant, and Vijay~V. Vazirani.
\newblock Random generation of combinatorial structures from a uniform
  distribution.
\newblock \emph{Theoretical Computer Science}, 32:\penalty0 169--188, 1986.

\bibitem[Johnson and Lindenstrauss(1984)]{Johnson/Lindenstrauss/84}
W.~Johnson and J.~Lindenstrauss.
\newblock Extensions of {L}ipschitz maps into a {H}ilbert space.
\newblock \emph{Contemporary Mathematics}, 26:\penalty0 189–--206, 1984.

\bibitem[Jordan(1986)]{Jordan86}
Michael~I. Jordan.
\newblock Attractor dynamics and parallelism in a connectionist sequential
  machine.
\newblock In \emph{Proceedings of the 8th Annual Conference of the Cognitive
  Science Society}, 1986.

\bibitem[Kalai and Vempala(2006)]{Adam/Vempala/06}
Adam~Tauman Kalai and Santosh Vempala.
\newblock Simulated annealing for convex optimization.
\newblock \emph{Mathematics of Operations Research}, 31\penalty0 (2):\penalty0
  253--266, 2006.

\bibitem[Kendall(1938)]{Kendall/38}
Maurice Kendall.
\newblock A new measure of rank correlation.
\newblock \emph{Biometrika}, 30:\penalty0 81--89, 1938.

\bibitem[Kenyon-Mathieu and Schudy(2007)]{Mathieu/Schudy/07}
Claire Kenyon-Mathieu and Warren Schudy.
\newblock How to rank with few errors: {A PTAS} for weighted feedback arc set
  on tournaments.
\newblock In \emph{Proceedings of the 37th Annual ACM Symposium on Theory of
  Computing}, 2007.

\bibitem[Kirkpatrick et~al.(1983)Kirkpatrick, Gelatt, and
  Vecchi]{Kirkpatrick/etal/83}
S.~Kirkpatrick, C.D. Gelatt, and M.P. Vecchi.
\newblock Optimization by simulated annealing.
\newblock \emph{Science}, 220:\penalty0 671--680, 1983.

\bibitem[Kivinen and Warmuth(1997)]{Kivinen/Warmuth/97}
Jyrki Kivinen and Manfred~K. Warmuth.
\newblock Exponentiated gradient versus gradient descent for linear predictors.
\newblock \emph{Information and Computation}, 132\penalty0 (1):\penalty0 1--63,
  1997.

\bibitem[Kivinen et~al.(2001)Kivinen, Smola, and Williamson]{Kivinen/etal/01}
Jyrki Kivinen, Alexander~J. Smola, and Robert~C. Williamson.
\newblock Online learning with kernels.
\newblock In \emph{Advances in Neural Information Processing Systems 14}, 2001.

\bibitem[Klein et~al.(2008)Klein, Brefeld, and Scheffer]{Klein/etal/08}
Thoralf Klein, Ulf Brefeld, and Tobias Scheffer.
\newblock Exact and approximate inference for annotating graphs with structural
  svms.
\newblock In \emph{Proceedings of the European Conference on Machine Learning,
  and Principles and Practice of Knowledge Discovery in Databases}, 2008.

\bibitem[Kulesza and Pereira(2007)]{Kulesza/Pereira/07}
Alex Kulesza and Fernando Pereira.
\newblock Structured learning with approximate inference.
\newblock In \emph{Advances in Neural Information Processing Systems 20}, 2007.

\bibitem[Lafferty et~al.(2001)Lafferty, McCallum, and
  Pereira]{Lafferty/etal/01}
John Lafferty, Andrew McCallum, and Fernando C.~N. Pereira.
\newblock Conditional random fields: {P}robabilistic models for segmenting and
  labeling sequence data.
\newblock In \emph{Proceedings of the 18th International Conference on Machine
  Learning}, pages 282--289, 2001.

\bibitem[Langford and Zadrozny(2005)]{Langford/Zadrozny/05}
John Langford and Bianca Zadrozny.
\newblock Estimating class membership probabilities using classifier learners.
\newblock In \emph{Proceedings of the 10th International Workshop on Artificial
  Intelligence and Statistics}, 2005.

\bibitem[Le and Smola(2007)]{Le/Smola/07}
Quoc~V. Le and Alexander~J. Smola.
\newblock Direct optimization of ranking measures.
\newblock Technical report, NICTA, Canberra, Australia, 2007.

\bibitem[Lebanon and Lafferty(2001)]{Lebanon/Lafferty/01}
Guy Lebanon and John Lafferty.
\newblock Boosting and maximum likelihood for exponential models.
\newblock In \emph{Advances in Neural Information Processing Systems 14}, 2001.

\bibitem[Liang et~al.(2006)Liang, Bouchard-C{\^o}t{\'e}, Klein, and
  Taskar]{Liang/etal/06}
Percy Liang, Alexandre Bouchard-C{\^o}t{\'e}, Dan Klein, and Ben Taskar.
\newblock An end-to-end discriminative approach to machine translation.
\newblock In \emph{Proceedings of the 21st International Conference on
  Computational Linguistics and the 44th Annual Meeting of the Association for
  Computational Linguistics}, 2006.

\bibitem[Lov{\'a}sz and Vempala(2006)]{Lovasz/Vempala/04}
L{\'a}szl{\'o} Lov{\'a}sz and Santosh Vempala.
\newblock Hit-and-run from a corner.
\newblock \emph{SIAM Journal on Computing}, 35\penalty0 (4):\penalty0
  985--1005, 2006.

\bibitem[Martins et~al.(2009)Martins, Smith, and Xing]{Martins/etal/2009}
Andr{\'e} F.~T. Martins, Noah~A. Smith, and Eric~P. Xing.
\newblock Polyhedral outer approximations with application to natural language
  parsing.
\newblock In \emph{Proceedings of the 26th Annual International Conference on
  Machine Learning}, 2009.

\bibitem[Mauser and Stahl(2007)]{MauserH}
Harald Mauser and Martin Stahl.
\newblock Chemical fragment spaces for de novo design.
\newblock \emph{Journal of Chemical Information and Modeling}, 47\penalty0
  (2):\penalty0 318--324, 2007.

\bibitem[McCallum(1999)]{McCallum99}
Andrew McCallum.
\newblock Multi-label text classification with a mixture model trained by {EM}.
\newblock In \emph{In Proceedings of the Workshop on Text Learning at the 16th
  National Conference on Artificial Intelligence}, 1999.

\bibitem[McDonald et~al.(2005)McDonald, Pereira, Ribarov, and
  Haji\v{c}]{McDonald/etal/05}
Ryan McDonald, Fernando Pereira, Kiril Ribarov, and Jan Haji\v{c}.
\newblock Non-projective dependency parsing using spanning tree algorithms.
\newblock In \emph{Proceedings of the Human Language Technology Conference and
  the Conference on Empirical Methods in Natural Language Processing}, 2005.

\bibitem[Metropolis et~al.(1953)Metropolis, Rosenbluth, Rosenbluth, Teller, and
  Teller]{Metropolis/etal/53}
Nicholas Metropolis, Arianna~W. Rosenbluth, Marshall~N. Rosenbluth, Augusta~H.
  Teller, and Edward Teller.
\newblock Equation of state calculation by fast computing machines.
\newblock \emph{Journal of Chemical Physics}, 21:\penalty0 1087--1092, 1953.

\bibitem[Minsky and Papert(1969)]{Minsky/Papert/69}
Marvin Minsky and Seymour Papert.
\newblock \emph{Perceptrons: {A}n Introduction to Computational Geometry}.
\newblock MIT Press, Cambridge, Massachusetts, 1969.

\bibitem[Mitchell(2006)]{Mitchell06}
Tom~M. Mitchell.
\newblock The discipline of machine learning.
\newblock Technical Report CMU-ML-06-108, School of Computer Science, Carnegie
  Mellon University, 2006.

\bibitem[Ng and Jordan(2001)]{Ng/Jordan/01}
Andrew~Y. Ng and Michael~I. Jordan.
\newblock On discriminative vs. generative classifiers: A comparison of
  logistic regression and na\"{\i}ve {B}ayes.
\newblock In \emph{Advances in Neural Information Processing Systems 14}, pages
  841--848, 2001.

\bibitem[Nilsson(1965)]{Nilsson65}
Nils~J. Nilsson.
\newblock \emph{Learning Machines: {Foundations} of Trainable
  Pattern-Classifying Systems}.
\newblock McGraw-Hill, New York, NY, USA, 1965.

\bibitem[Novikoff(1962)]{Novikoff62}
Albert~B. Novikoff.
\newblock On convergence proofs on perceptrons.
\newblock In \emph{Proceedings of the Symposium on the Mathematical Theory of
  Automata}, volume~12, 1962.

\bibitem[Papadimitriou(1994)]{Papadimitriou94}
Christos~H. Papadimitriou.
\newblock \emph{Computational Complexity}.
\newblock Addison-Wesley, 1994.

\bibitem[Papadimitriou and Yannakakis(1984)]{Papadimitriou/Yannakakis/82}
Christos~H. Papadimitriou and Mihalis Yannakakis.
\newblock The complexity of facets (and some facets of complexity).
\newblock \emph{Journal of Computer and System Sciences}, 28\penalty0
  (2):\penalty0 244--259, 1984.

\bibitem[Platt(2005)]{Platt05}
John~C. Platt.
\newblock Fastmap, metricmap, and landmark {MDS} are all {Nystr\"om}
  algorithms.
\newblock In \emph{Proceedings of the 10th International Workshop on Artificial
  Intelligence and Statistics}, 2005.

\bibitem[Propp and Wilson(1996)]{Propp/Wilson/96}
James~Gary Propp and David~Bruce Wilson.
\newblock Exact sampling with coupled {M}arkov chains and applications to
  statistical mechanics.
\newblock \emph{Random Structures and Algorithms}, 9\penalty0 (1-2):\penalty0
  223--252, 1996.

\bibitem[Rabiner(1989)]{Rabiner89}
Lawrence~R. Rabiner.
\newblock A tutorial on hidden {M}arkov models and selected applications in
  speech recognition.
\newblock \emph{Proceedings of the IEEE}, 2\penalty0 (77):\penalty0 257--286,
  1989.

\bibitem[Randall(2003)]{Randall/03}
Dana Randall.
\newblock Mixing.
\newblock In \emph{Proceedings of the 44th Symposium on Foundations of Computer
  Science}, 2003.

\bibitem[Ratliff(2009)]{Ratliff09}
Nathan Ratliff.
\newblock \emph{Learning to Search: Structured Prediction Techniques for
  Imitation Learning}.
\newblock PhD thesis, Carnegie Mellon University, Robotics Institute, May 2009.

\bibitem[Ratliff et~al.(2006)Ratliff, Bagnell, and Zinkevich]{Ratliff/etal/06}
Nathan Ratliff, J.~Andrew~(Drew) Bagnell, and Martin Zinkevich.
\newblock Maximum margin planning.
\newblock In \emph{Proceedings of the 23rd International Conference on Machine
  Learning}, 2006.

\bibitem[Ratliff et~al.(2007)Ratliff, Bagnell, and
  Zinkevich]{Ratliff/etal/2007}
Nathan Ratliff, J.~Andrew~(Drew) Bagnell, and Martin Zinkevich.
\newblock (online) subgradient methods for structured prediction.
\newblock In \emph{Proceedings ot the 11th International Conference on
  Artificial Intelligence and Statistics}, 2007.

\bibitem[Rifkin(2002)]{Rifkin02}
Ryan~M. Rifkin.
\newblock \emph{Everything old is new again : {A} fresh look at historical
  approaches in machine learning}.
\newblock PhD thesis, Massachusetts Institute of Technology, Sloan School of
  Management, 2002.

\bibitem[Rifkin and Lippert(2007)]{Rifkin/Lippert/07}
Ryan~M. Rifkin and Ross~A. Lippert.
\newblock Notes on regularized least squares.
\newblock Technical Report MIT-CSAIL-TR-2007-025, CBCL Memo 268, MIT Computer
  Science and Artificial Laboratory, 2007.

\bibitem[Rosenblatt(1958)]{Rosenblatt58}
Frank Rosenblatt.
\newblock The perceptron: {A} probabilistic model for information storage and
  organization in the brain.
\newblock \emph{Psychological Review}, 65\penalty0 (6):\penalty0 386--408,
  1958.

\bibitem[Rousu et~al.(2005)Rousu, Saunders, Szedm{\'a}k, and
  Shawe-Taylor]{Rousu/etal/05}
Juho Rousu, Craig Saunders, S{\'a}ndor Szedm{\'a}k, and John Shawe-Taylor.
\newblock Learning hierarchical multi-category text classification models.
\newblock In \emph{Proceedings of the 22nd International Conference on Machine
  Learning}, 2005.

\bibitem[Rousu et~al.(2006)Rousu, Saunders, Szedm{\'a}k, and
  Shawe-Taylor]{Rousu/etal/06}
Juho Rousu, Craig Saunders, S{\'a}ndor Szedm{\'a}k, and John Shawe-Taylor.
\newblock Kernel-based learning of hierarchical multilabel classification
  models.
\newblock \emph{Journal of Machine Learning Research}, 7\penalty0
  (2-3):\penalty0 1601--1626, 2006.

\bibitem[Rumelhart et~al.(1987)Rumelhart, Hinton, and
  Williams]{Rumelhart/etal/87}
D.~E. Rumelhart, G.~E. Hinton, and R.~J. Williams.
\newblock Learning internal representations by error propagation.
\newblock In D.~E. Rumelhart, J.~L. McClelland, et~al., editors, \emph{Parallel
  Distributed Processing: Volume 1: Foundations}, pages 318--362. MIT Press,
  Cambridge, 1987.

\bibitem[Rumelhart et~al.(1986)Rumelhart, Hinton, and
  Williams]{Rumelhart/etal/86}
David~E. Rumelhart, Geoffrey~E. Hinton, and Ronald~J. Williams.
\newblock {Learning internal representations by error propagation}.
\newblock \emph{MIT Press Computational Models Of Cognition And Perception
  Series}, pages 318--362, 1986.

\bibitem[Sattolo(1986)]{Sattolo86}
Sandra Sattolo.
\newblock An algorithm to generate a random cyclic permutation.
\newblock \emph{Inf. Process. Lett.}, 22\penalty0 (6):\penalty0 315--317, 1986.

\bibitem[Schapire and Singer(2000)]{Schapire/Singer/00}
Robert~E. Schapire and Yoram Singer.
\newblock Boostexter: {A} boosting-based system for text categorization.
\newblock \emph{Machine Learning}, 39\penalty0 (2/3):\penalty0 135--168, 2000.

\bibitem[Schapire and Singer(1999)]{Schapire/Singer/98}
Robert~E. Schapire and Yoram Singer.
\newblock Improved boosting algorithms using confidence-rated predictions.
\newblock \emph{Machine Learning}, 37\penalty0 (3):\penalty0 297--336, 1999.

\bibitem[Sch\"{o}lkopf and Smola(2002)]{Schoelkopf/Smola/02}
Bernhard Sch\"{o}lkopf and Alexander~J. Smola.
\newblock \emph{Learning with Kernels: {S}upport Vector Machines,
  Regularization, Optimization, and Beyond}.
\newblock The MIT Press, 2002.

\bibitem[Sch{\"o}lkopf et~al.(1998)Sch{\"o}lkopf, Smola, and
  M{\"u}ller]{Scholkopf/etal/97}
Bernhard Sch{\"o}lkopf, Alexander~J. Smola, and Klaus-Robert M{\"u}ller.
\newblock Nonlinear component analysis as a kernel eigenvalue problem.
\newblock \emph{Neural Computation}, 10\penalty0 (5):\penalty0 1299--1319,
  1998.

\bibitem[Sch{\" o}lkopf et~al.(2001)Sch{\" o}lkopf, Herbrich, and
  Smola]{Schoelkopf/etal/01}
Bernhard Sch{\" o}lkopf, Ralf Herbrich, and Alexander~J. Smola.
\newblock A generalized representer theorem.
\newblock In \emph{Proceedings of the 14th Annual Conference on Computational
  Learning Theory}, 2001.

\bibitem[Schraudolph(1999)]{Schraudolph99}
Nicol~N. Schraudolph.
\newblock Local gain adaptation in stochastic gradient descent.
\newblock In \emph{Proceedings of the 9th International Conference on Neural
  Networks}, 1999.

\bibitem[Sha and Pereira(2003)]{Fei/Pereira/03}
Fei Sha and Fernando C.~N. Pereira.
\newblock Shallow parsing with conditional random fields.
\newblock In \emph{Proceedings of the Human Language Technology Conference of
  the North American Chapter of the Association for Computational Linguistics},
  2003.

\bibitem[Shalev-Shwartz and
  Singer(2007{\natexlab{a}})]{Shalev-Shwartz/Singer/07}
Shai Shalev-Shwartz and Yoram Singer.
\newblock A primal-dual perspective of online learning algorithms.
\newblock \emph{Machine Learning}, 69\penalty0 (2-3):\penalty0 115--142,
  2007{\natexlab{a}}.

\bibitem[Shalev-Shwartz and Singer(2007{\natexlab{b}})]{Shwartz/Singer/07}
Shai Shalev-Shwartz and Yoram Singer.
\newblock A unified algorithmic approach for efficient online label ranking.
\newblock In \emph{Proceedings of the 11th International Conference on
  Artificial Intelligence and Statistics}, 2007{\natexlab{b}}.

\bibitem[Shalev-Shwartz and Singer(2006)]{Shwartz/etal/06}
Shai Shalev-Shwartz and Yoram Singer.
\newblock Efficient learning of label ranking by soft projections onto
  polyhedra.
\newblock \emph{Journal of Machine Learning Research}, 7:\penalty0 1567--1599,
  2006.

\bibitem[Shalev-Shwartz et~al.(2007)Shalev-Shwartz, Singer, and
  Srebro]{Shwartz/etal/07}
Shai Shalev-Shwartz, Yoram Singer, and Nathan Srebro.
\newblock Pegasos: {P}rimal estimated sub-gradient solver for {SVM}.
\newblock In \emph{Proceedings of the 24th International Conference on Machine
  Learning}, 2007.

\bibitem[Sinclair(1992)]{Sinclair/92}
Alistair~J. Sinclair.
\newblock Improved bounds for mixing rates of {M}arkov chains and
  multicommodity flow.
\newblock \emph{Information and Computation}, 1:\penalty0 351--370, 1992.

\bibitem[Sonnenburg(2008)]{Sonnenburg08}
S\"oren Sonnenburg.
\newblock \emph{Machine Learning for Genomic Sequence Analysis}.
\newblock PhD thesis, Fraunhofer Institute FIRST, December 2008.

\bibitem[Spearman(1904)]{Spearman/04}
Charles Spearman.
\newblock The proof and measurement of association between two things.
\newblock \emph{American Journal of Psychology}, 15:\penalty0 72--101, 1904.

\bibitem[Stefankovic et~al.(2009)Stefankovic, Vempala, and
  Vigoda]{Stefankovic/etal/07}
Daniel Stefankovic, Santosh Vempala, and Eric Vigoda.
\newblock Adaptive simulated annealing: {A} near-optimal connection between
  sampling and counting.
\newblock \emph{Journal of the ACM}, 56\penalty0 (3):\penalty0 1--36, 2009.

\bibitem[Suykens(1999)]{Suykens99}
Johan~A.K. Suykens.
\newblock Multiclass least squares support vector machines.
\newblock In \emph{Proceedings of the International Joint Conference on Neural
  Networks}, 1999.

\bibitem[Suykens and Vandewalle(1999)]{Suykens/Vandewalle/99}
Johan~A.K. Suykens and Joos~P.L. Vandewalle.
\newblock Least squares support vector machine classifiers.
\newblock \emph{Neural Processing Letters}, 9\penalty0 (3):\penalty0 293--300,
  1999.

\bibitem[Taskar(2004)]{Taskar/04}
Ben Taskar.
\newblock \emph{Learning Structured Prediction Models: {A} Large Margin
  Approach}.
\newblock PhD thesis, Stanford University, 2004.

\bibitem[Taskar et~al.(2003)Taskar, Guestrin, and Koller]{Taskar/etal/03}
Ben Taskar, Carlos Guestrin, and Daphne Koller.
\newblock Max-margin {M}arkov networks.
\newblock In \emph{Advances in Neural Information Processing Systems 16}, 2003.

\bibitem[Taskar et~al.(2005)Taskar, Chatalbashev, Koller, and
  Guestrin]{Taskar/etal/05}
Ben Taskar, V.~Chatalbashev, Daphne Koller, and C.~Guestrin.
\newblock Learning structured prediction models: {A} large margin approach.
\newblock In \emph{Proceedings of the 22nd International Conference on Machine
  Learning}, 2005.

\bibitem[Tsochantaridis et~al.(2005)Tsochantaridis, Joachims, Hofmann, and
  Altun]{Tsochantaridis/etal/05}
Ioannis Tsochantaridis, Thorsten Joachims, Thomas Hofmann, and Yasemin Altun.
\newblock Large margin methods for structured and interdependent output
  variables.
\newblock \emph{Journal of Machine Learning Research}, 6:\penalty0 1453--1484,
  2005.

\bibitem[van Zuylen and Williamson(2007)]{Zuylen/Williamson/07}
Anke van Zuylen and David~P. Williamson.
\newblock Deterministic algorithms for rank aggregation and other ranking and
  clustering problems.
\newblock In \emph{Proceedings of the 5th International Workshop on
  Approximation and Online Algorithms}, 2007.

\bibitem[Vapnik(1995)]{Vapnik95}
Vladimir Vapnik.
\newblock \emph{The Nature of Statistical Learning Theory}.
\newblock Springer, 1995.

\bibitem[Vigoda(Fall 2006)]{Vigoda06}
Eric Vigoda.
\newblock Markov chain {M}onte carlo methods, lecture notes.
\newblock \url{http://www.cc.gatech.edu/~vigoda/teaching.html}, Fall 2006.

\bibitem[Vishwanathan et~al.(2006)Vishwanathan, Schraudolph, and
  Smola]{Vishy/etal/06}
S.V.N. Vishwanathan, Nicol~N. Schraudolph, and Alexander~J. Smola.
\newblock Step size adaptation in reproducing kernel {H}ilbert space.
\newblock \emph{Journal of Machine Learning Research}, 7:\penalty0 1107--1133,
  2006.

\bibitem[Waibel et~al.(1989)Waibel, Hanazawa, Hinton, Shikano, and
  Lang]{Waibal/etal/89}
Alexander Waibel, Toshiyuki Hanazawa, Geoffrey Hinton, Kiyohiro Shikano, and
  Kevin~J. Lang.
\newblock Phoneme recognition using time-delay neural networks.
\newblock \emph{IEEE Transactions on Acoustics, Speech, and Signal Processing},
  37\penalty0 (3):\penalty0 328--339, 1989.

\bibitem[Weston et~al.(2002)Weston, Chapelle, Elisseeff, Sch{\"o}lkopf, and
  Vapnik]{Weston/etal/02}
Jason Weston, Olivier Chapelle, Andr{\'e} Elisseeff, Bernhard Sch{\"o}lkopf,
  and Vladimir Vapnik.
\newblock Kernel dependency estimation.
\newblock In \emph{Advances in Neural Information Processing Systems 15}, 2002.

\bibitem[Yannakakis(1978)]{Yannakakis78}
Mihalis Yannakakis.
\newblock Node- and edge-deletion {NP}-complete problems.
\newblock In \emph{Proceedings of the Annual ACM Symposium on Theory of
  Computing}, 1978.

\bibitem[Zemel(1981)]{Zemel/81}
E.~Zemel.
\newblock Measuring the quality of approximate solutions to zero-one
  programming problems.
\newblock \emph{Mathematics of Operations Research}, 6\penalty0 (3):\penalty0
  319--332, 1981.

\bibitem[Zinkevich(2003)]{Zinkevich03}
Martin Zinkevich.
\newblock Online convex programming and generalized infinitesimal gradient
  ascent.
\newblock In \emph{Proceedings of the 20th International Conference on Machine
  Learning}, 2003.

\end{thebibliography}

%\clearpage
%\input{src/cv}
%\clearemptydoublepage

\end{document}